\documentclass[acmsmall,screen,nonacm]{acmart}
\usepackage[font=scriptsize]{caption}
\usepackage{array}
\usepackage{makecell}
\usepackage{graphicx}
\usepackage{float}
\usepackage{amsmath}

\usepackage{amssymb}
\usepackage{booktabs}
\usepackage{bbm}
\usepackage{listings}
\usepackage{rotating}
\usepackage{color}
\usepackage{makecell}
\usepackage{multirow}
\usepackage{booktabs}
\usepackage{xcolor}
\usepackage{algorithm}
\usepackage{algorithmicx}
\usepackage{algpseudocode}
\usepackage{subcaption}
\AtBeginEnvironment{bmatrix}{\setlength{\arraycolsep}{0.5pt}}

\newtheorem{proposition}{\textbf{Proposition}}

\newcolumntype{M}[1]{>{\centering\arraybackslash}m{#1}}
\usepackage{graphicx,booktabs,array}

\usepackage[capitalize]{cleveref}
\crefname{section}{Sec.}{Secs.}
\Crefname{section}{Section}{Sections}
\Crefname{table}{Table}{Tables}
\crefname{table}{Tab.}{Tabs.}

\AtBeginDocument{%
  }

\acmSubmissionID{123-A56-BU3}

\begin{document}

\title{Laplacian-based Cluster-Contractive t-SNE for High Dimensional Data Visualization}


\author{Yan Sun$^*$}
\author{Yi Han$^*$}
\author{Jicong Fan$^\heartsuit$}
\orcid{0001-9665-0355}
\affiliation{%
  \institution{School of Data Science, The Chinese University of Hong Kong, Shenzhen}
  \city{Shenzhen}
  \country{China}
}
\email{fanjicong@cuhk.edu.cn}

\makeatletter
\let\@authorsaddresses\@empty
\makeatother



\begin{abstract}
  Dimensionality reduction techniques aim at representing high-dimensional data in low-dimensional spaces to extract hidden and useful information or facilitate visual understanding and interpretation of the data.  However, few of them take into consideration the potential cluster information  contained implicitly in the high-dimensional data. In this paper, we propose \textsc{Lap}tSNE, a new graph-layout nonlinear dimensionality reduction method based on t-SNE, one of the best techniques for visualizing high-dimensional data as 2D scatter plots. Specifically, \textsc{Lap}tSNE leverages the eigenvalue information of the graph Laplacian to shrink the potential clusters in the low-dimensional embedding when learning to preserve  the local and global structure from high-dimensional space to low-dimensional space.  It is nontrivial to solve the proposed model because the eigenvalues of normalized symmetric Laplacian are functions of the decision variable. We provide a majorization-minimization algorithm with convergence guarantee to solve the optimization problem of \textsc{Lap}tSNE and show how to calculate the gradient analytically, which may be of broad interest when considering optimization with Laplacian-composited objective. We evaluate our method by a formal comparison with state-of-the-art methods on seven benchmark datasets, both visually and via established quantitative measurements. The results demonstrate the superiority of our method over baselines such as t-SNE and UMAP. We also provide out-of-sample extension, large-scale extension and mini-batch extension for our \textsc{Lap}tSNE to facilitate dimensionality reduction in various scenarios.
\end{abstract}



\maketitle
\def\thefootnote{*}\footnotetext{These authors contributed equally to this work.}
\def\thefootnote{$\heartsuit$}\footnotetext{Corresponding author.}

\section{Introduction}\label{intro}
Nowadays, data in science and engineering are usually high-dimensional. For instance, the number of genes (as features) in a gene dataset is often larger than ten thousands. In computer vision, the dimension of a vectorized high-resolution image can be higher than one million. In addition, the structures of these high-dimensional data are often complicated, especially when the numbers of samples are large.  Therefore, extracting potentially useful and understandable patterns from these intricate datasets becomes necessary and helpful. Among all the extant methods, dimensionality reduction (DR) is critically important for understanding the structures of large datasets. DR aims to extract or produce informative low-dimensional features from high-dimensional data. Such features can be easily visualized to identify the hidden patterns of the original data, provided that the reduced dimension is low enough, e.g. 3-D or 2-D. The low-dimensional features provide convenience for downstream tasks such as classification and clustering without the curse of dimensionality and often improve the corresponding performances.

DR has been an active and important research topic for more than fifty years. DR methods can be organized into two categories: linear methods and nonlinear methods. Principal Component Analysis (PCA) \cite{pearson1901liii,jolliffe2016principal}, Multidimensional Scaling (MDS) \cite{sammon1969nonlinear}, and Linear Discriminant Analysis (LDA) \cite{fisher1936use} are well-known linear DR methods. These methods are simple, effective, and well-understood, though they have difficulty in handling data with nonlinear low-dimensional latent structures. In the past decades, many nonlinear dimensionality reduction (NLDR) methods have been proposed to extract low-dimensional nonlinear features or visualize high-dimensional data in 2D or 3D spaces. Well-known NLDR methods include Self-Organized Map (SOM) \cite{kohonen1982self}, Kernel PCA (KPCA) \cite{scholkopf1998nonlinear}, Principal Curves \cite{hastie1989principal}, Locally Linear Embedding (LLE)\cite{roweis2000nonlinear}, Isomap \cite{tenenbaum2000global}, Laplacian Eigenmap (LE) \cite{baker1977numerical}, AutoEncoder \cite{demers1992non, hinton2006reducing}, t-distributed Stochastic Neighbor Embedding (t-SNE) \cite{van2008visualizing}, and Uniform Manifold Approximation and Projection (UMAP) \cite{mcinnes2018umap}. Note that LLE, Isomap, and LE are not roubust to noise and outliers and their performance on real data are not satisfactory enough. KPCA, AutoEncoder, and stacked AutoEncoders are actually nonlinear feature extraction methods and are not effective in data visualization. 

Among the aforementioned NLDR methods, t-SNE developed by van der Maaten and Hinton \cite{van2008visualizing} is arguably one of the most powerful and state-of-the-art methods in a wide range of applications. t-SNE maps the data points to a two- or three dimensional space, which exhibits the intrinsic data distribution of the original high-dimensional data. Therefore, the low-dimensional embedding always reveal trends, patterns and outliers. In many practices of scientific research, t-SNE has become an extraordinary tool of data visualization. It is worth mentioning that there have been a few variants of t-SNE \cite{yang2009heavy,carreira2010elastic,xie2011m,van2014accelerating,gisbrecht2015parametric,pezzotti2016approximated,linderman2019fast,chatzimparmpas2020t}. For instance, Yang et al. \cite{yang2009heavy} generalized t-SNE to accommodate various heavy-tailed embedding similarity functions and presented a fixed-point optimization algorithm that can be applied to all heavy-tailed functions. Van Der Maaten \cite{van2014accelerating} proposed to use tree-based algorithms to accelerate t-SNE. 

Data visualization is growing critically important nowadays for understanding the structure-complicated high-dimensional datasets, and has been recognized as one of the building blocks of data science \cite{donoho201750}. In most data visualization tasks, we are interested in discovering the potential clusters of the data and indeed, the data often contain multiple clusters naturally. On such data, the outline of clusters generated by t-SNE are often over\textsc{Lap}ped and obscure, because t-SNE does not explicitly explore and exploit the potential cluster structure of the data. To tackle with this problem, in this paper, we propose a new method, \textsc{Lap}tSNE, to construct low-dimensional embedding in a cluster-informative manner. Our contributions are three-fold.
\begin{itemize}
    \item We present a new NLDR algorithm \textsc{Lap}tSNE that explores and exploits the potential clusters of the data and produces cluster-informative low-dimensional visualization. 
    \item The objective function of the proposed method involves the eigenvalues of a graph Laplacian computed from the decision variables, which leads to difficulty in solving the optimization problem. We therefore develop an effective algorithm with convergence guarantee to solve the optimization of \textsc{Lap}tSNE.
    \item We provide out-of-sample extension, large-scale extension, and mini-batch extension for our \textsc{Lap}tSNE, which facilitate the implementation of \textsc{Lap}tSNE in various scenarios.
\end{itemize}
Experiments on many benchmark datasets (e.g. COIL20 \cite{nene1996columbia} and MNIST \cite{deng2012mnist}) further verify the effectiveness and superiority of our methods. For instance, compared to the vanilla t-SNE, in our method, the boundaries of clusters in the 2D embedding are clearer and the clusters are more compact. Quantitative evaluations such as the k-NN generalization error and clustering NMI also confirm the superiority of our methods. 
\section{Preliminary Knowledge}
t-SNE is actually a modification of SNE \cite{hinton2002stochastic}. They aim to preserve the pair-wise similarities from high-dimension space $\mathcal{P}$ to low-dimension $\mathcal{Q}$. Here the pair-wise similarities are quantified by neighborhood probability, i.e., the probability of two data points are neighbors mutually. Specifically, given $\left \{ \mathbf{x}_i \right \}_{i\in[N]} \in  \mathbb{R}^{D}$ (we define $[N]:=\{1,2,\ldots,N\}$ for convenience),  SNE and t-SNE find a low-dimensional embedding $\left \{\mathbf{y}_i \right \}_{i\in[N]} \in  \mathbb{R}^{d}$ where $d \ll D$, such that if $\mathbf{x}_i$ and $\mathbf{x}_j$ are close in the original data space, $\mathbf{y}_i$ and $\mathbf{y}_j$ are also close. 
t-SNE starts by computing the joint probability distribution over all input data, represented by a symmetric matrix $\mathbf{P_X}$. When $i=j$, $p_{ij} = 0$. Otherwise, 
\begin{equation*}
p_{i j}=\frac{p_{i \mid j}+p_{j \mid i}}{2 N},
\end{equation*}
where 
\begin{equation}
\quad p_{j \mid i}=\frac{\exp \left(-\left\|\mathbf{x}_{i}-\mathbf{x}_{j}\right\|_{2}^{2} / 2 \tau_{i}^{2}\right)}{\sum_{\ell \in[N] \backslash\{i\}} \exp \left(-\left\|\mathbf{x}_{i}-\mathbf{x}_{\ell}\right\|_{2}^{2} / 2 \tau_{i}^{2}\right)}.
\end{equation}
Here $\tau_i$ denotes the bandwidth of the Gaussian kernel based on user-specified perplexity $Perp$ \cite{van2008visualizing}. Similarly, an affinity matrix $\mathbf{Q_Y}$ in the low-dimensional space can be computed. Compared to SNE, t-SNE replaces the Gaussian kernel in the low dimension with T-Student kernel in one degree of freedom (same as Cauchy kernel)\cite{souza2010kernel}. The T-Student kernel can help the learning strongly repel dissimilar data points that are modeled by a small pair-wise distance in the low-dimensional representation, thus alleviating the crowding problem. Specifically, in t-SNE, for $i\neq j$,

\begin{equation}
    q_{i j}=\frac{\left(1+\left\|\mathbf{y}_{i}-\mathbf{y}_{j}\right\|_{2}^{2}\right)^{-1}}{\sum_{\ell, s \in[N], \ell \neq s}\left(1+\left\|\mathbf{y}_{\ell}-\mathbf{y}_{s}\right\|_{2}^{2}\right)^{-1}}.
\end{equation}
t-SNE thus searches for $\left \{\mathbf{y}_i \right \}_{i\in[N]}$ to minimize the Kullback-Leibler (KL) divergence between the joint distribution of points in the input data space $\mathcal{P}$ and embedding space $\mathcal{Q}$, i.e.,
\begin{equation}
\begin{aligned}
\left(\mathbf{y}_{1}, \ldots, \mathbf{y}_{N}\right) &=\underset{\mathbf{y}_{1}, \ldots, \mathbf{y}_{N}}{\arg \min } ~~D_{K L}(\mathbf{P_X}, \mathbf{Q_Y})\\ 
& = \underset{\mathbf{y}_{1}, \ldots, \mathbf{y}_{N}}{\arg \min } \sum_{i \neq j} q_{i j} \log \frac{p_{i j}}{q_{i j}}.
\end{aligned}
\label{f4}
\end{equation}
More details can be found in \cite{van2008visualizing}.

\section{Cluster-Contractive t-SNE}
\subsection{Motivation}
Although t-SNE is effective in preserving local and global structures of high-dimensional data \cite{pmlr-v75-arora18a}, it does not explore or utilize the potential cluster structures that are prevalent in real datasets \cite{kobak2019art,li2017application}. The potential cluster structures should be a useful prior and exploited if possible, though it is difficult to know the number of clusters of high-dimensional data in advance. If the high-dimensional data indeed consist of multiple clusters, the clusters should be preserved in the low-dimensional embedding. Thus, besides matching the similarity matrix $\mathbf{P_X}$ and $\mathbf{Q_Y}$, we also want to get a similarity graph (from the low-dimensional embedding) of which the edges between different groups have very low weights and the edges within a group have high weights. In other words, data within the same cluster are similar to each other while data in different clusters are dissimilar from each other.

However, the aforementioned objective is intractable because we do not know the clusters (or even the number of clusters) of the high-dimensional data. Note that the affinity matrix $\mathbf{P_X}$ in t-SNE is actually a connected graph owing to the use of Gaussian kernel. Therefore, solving \eqref{f4} does guarantee preserving clusters in the low-dimension space. Even if $\mathbf{P_X}$ is not connected but has multiple connected components, the solution of \eqref{f4} may not preserve the clusters.

To handle the problem, we consider using the graph Laplacian defined as
\begin{equation}\label{eq_Laplacian} 
\mathbf{L} = \mathbf{I} - \mathbf{D}^{-1/2}\mathbf{A}\mathbf{D}^{-1/2}.
\end{equation}
It is actually a symmetric normalized Laplacian matrix. In \eqref{eq_Laplacian}, $\mathbf{A}$ ($\mathbf{Q_Y}$ in this paper) is an adjacency (or similarity) matrix and $\mathbf{D}$ is the degree matrix (a diagonal matrix) of $\mathbf{A}$ defined as $D_{ii}=\sum_{j}a_{ij}$.
Laplacian matrix has the property of signifying the number of clusters (Proposition 1 in \cite{chung1997spectral}), i.e., the multiplicity $k$ of the eigenvalue 0 of $\mathbf{L}$ is equal to the number of connected components  $A_{1}, \ldots, A_{k}$ in the graph. When the eigenvalue 0 has a multiplicity k, we can observe k completely disconnected clusters. Therefore, we propose to maximize the number of zero eigenvalues of $\mathbf{L}$ such that we may identify more clusters from the data.

\subsection{Proposed Model}
In order to preserve both data structure and cluster information,  we may consider the following problem
\begin{equation}\label{eq_constrained_KL}
\begin{aligned}
\mathop{\text{minimize}}_{\mathbf{Y}}&~~ \text{KL}(\mathbf{P_X},\mathbf{Q_Y})\\
\text{subject to}&~~\sum_{i=1}^N\mathbbm{1}\big(\sigma_i(\mathbf{L}_{\mathbf{Y}})=0\big)=k,
\end{aligned}
\end{equation}
where $\mathbf{L}_Y$ is the symmetric normalized Laplacian matrix computed from $\mathbf{Q}_{\mathbf{Y}}$, $\sigma_i(\cdot)$ denotes the $i$-th eigenvalue of matrix, and $\mathbbm{1}(\cdot)$ is an indicator function with $\mathbbm{1}(\textit{True})=1$ and $\mathbbm{1}(\textit{False})=0$.  In \eqref{eq_constrained_KL}, we hope that there are exactly $k$ clusters in the low-dimensional embedding.
But it is difficult to know $k$ in advance.  We therefore relax \eqref{eq_constrained_KL} to the following regularized problem 
\begin{equation}\label{eq_reg_eig_l0}
\mathop{\text{minimize}}_{\mathbf{Y}} \text{KL}(\mathbf{P_X},\mathbf{Q_Y})+\lambda \Vert\mathbf{\sigma}(\mathbf{L}_{\mathbf{Y}})\Vert_0,
\end{equation}
where $\boldsymbol{\sigma}(\mathbf{L}_{\mathbf{Y}})=[\sigma_1(\mathbf{L}_{\mathbf{Y}}),\sigma_2(\mathbf{L}_{\mathbf{Y}}),\ldots,\sigma_N(\mathbf{L}_{\mathbf{Y}})]^\top$ and $\lambda$ is a tuning parameter.  $\Vert\cdot\Vert_0$ denotes the number of nonzero elements in a vector and $\Vert\boldsymbol{\sigma}(\mathbf{L}_{\mathbf{Y}})\Vert_0$ is actually the rank of $\mathbf{L}_{\mathbf{Y}}$.  In \eqref{eq_reg_eig_l0}, we want to increase the number of zero eigenvalues as large as possible such that the number of clusters in the data is sufficiently large.  Problem \eqref{eq_reg_eig_l0} is NP-hard due to the presence of the $\ell_0$ norm. It is known that the $\ell_p$ (quasi) norms, $\Vert\mathbf{x}\Vert_p:=(\sum_i \vert x_i\vert^p)^{1/p}$ ($0<p\leq 1$) are popular proxies of the $\ell_0$ norm. Particularly, $\Vert\mathbf{x}\Vert_1$ is a convex relaxation of $\Vert\mathbf{x}\Vert_0$. Therefore, we relax \eqref{eq_reg_eig_l0} to
\begin{equation}\label{eq_reg_eig_lp}
\mathop{\text{minimize}}_{\mathbf{Y}} \text{KL}(\mathbf{P_X},\mathbf{Q_Y})+\lambda \sum_{i=1}^N\sigma_i^p(\mathbf{L}_{\mathbf{Y}}).
\end{equation}
Note that $\sum_{i=1}^N\sigma_i^p(\mathbf{L}_{\mathbf{Y}})\rightarrow \Vert\mathbf{\sigma}(\mathbf{L}_{\mathbf{Y}})\Vert_0$ when $p\rightarrow 0$.
In this study, we only consider the case $p=1$ for simplicity. Then we arrive at
\begin{equation}\label{eq_reg_eig_trace}
\mathop{\text{minimize}}_{\mathbf{Y}} \text{KL}(\mathbf{P_X},\mathbf{Q_Y}) + \lambda \mathrm{Tr}(\mathbf{V}_{\mathbf{L_Y}}^{\top}\mathbf{L}_{\mathbf{Y}}\mathbf{V}_{\mathbf{L_Y}}),
\end{equation}
where $\mathbf{V}_{\mathbf{L_Y}}\in\mathbb{R}^{N\times N}$ denotes the eigenvectors of $\mathbf{L}_{\mathbf{Y}}$. Since $\mathbf{V}_{\mathbf{L_Y}}$ depends on $\mathbf{Y}$, we have to rewrite \eqref{eq_reg_eig_trace} as
\begin{equation}\label{eq_reg_eig_trace_V}
\mathop{\text{minimize}}_{\mathbf{Y}} \text{KL}(\mathbf{P_X},\mathbf{Q_Y}) + \lambda \min_{\mathbf{V}^\top\mathbf{V}=\mathbf{I}} \mathrm{Tr}(\mathbf{V}^{\top}\mathbf{L}_{\mathbf{Y}}\mathbf{V}).
\end{equation}
Note that in \eqref{eq_reg_eig_trace_V}, the matrix multiplications in the trace operator are costly because $\mathbf{V}_{\mathbf{L_Y}}$ is a square matrix. On the other hand, in \eqref{eq_reg_eig_trace}, we may just shrink the smallest few eigenvalues because in practice the number of clusters in the high-dimensional data is not large (e.g. 10 or 100) and much less than $n$. In view of the two reasons, we solve the following problem instead of \eqref{eq_reg_eig_trace_V}
\begin{equation}\label{eq_reg_eig_trace_Vk}
\mathop{\text{minimize}}_{\mathbf{Y}} \text{KL}(\mathbf{P_X},\mathbf{Q_Y}) + \lambda \min_{\mathbf{V}_{\hat{k}}^\top\mathbf{V}_{\hat{k}}=\mathbf{I}} \mathrm{Tr}(\mathbf{V}_{\hat{k}}^{\top}\mathbf{L}_{\mathbf{Y}}\mathbf{V}_{\hat{k}}),
\end{equation}
where $\mathbf{V}_{\hat{k}}\in\mathbb{R}^{N\times \hat{k}}$ and $\hat{k}$ is an estimated number of potential clusters in $\mathbf{X}$. This is exactly our proposed method \textsc{Lap}tSNE.

\subsection{Optimization}\label{sec_opt}
Note that it is non-trivial to solve \eqref{eq_reg_eig_trace_Vk} due to the presence of $\mathbf{V}_{\hat{k}}$. We present a majorization-minimization \cite{vaida2005parameter} algorithm for  \eqref{eq_reg_eig_trace_Vk}. Specifically, at iteration $t$, we solve
\begin{equation}\label{eq_reg_eig_trace_Vk_MM}
\mathop{\text{minimize}}_{\mathbf{Y}} \text{KL}(\mathbf{P_X},\mathbf{Q_Y}) + \lambda  \mathrm{Tr}(\mathbf{V}_{t-1}^{\top}\mathbf{L}_{\mathbf{Y}}\mathbf{V}_{t-1}),
\end{equation}
where $\mathbf{V}_{t-1}$ is the eigenvectors of $\mathbf{L}_{\mathbf{Y}_{t-1}}$ corresponding to the smallest $\hat{k}$ eigenvalues. 
There is no need to obtain the exact solution of \eqref{eq_reg_eig_trace_Vk_MM} and hence we propose to just update $\mathbf{Y}$ by gradient descent.
For convenience,  let 
\begin{align}
&\mathcal{L}_1(\mathbf{Y})=\text{KL}(\mathbf{P_X},\mathbf{Q_Y}),\\
&{\mathcal{L}}_2(\mathbf{Y})=\lambda  \min_{\mathbf{V}_{\hat{k}}^\top\mathbf{V}_{\hat{k}}=\mathbf{I}}\mathrm{Tr}(\mathbf{V}_{\hat{k}}^{\top}\mathbf{L}_{\mathbf{Y}}\mathbf{V}_{\hat{k}}),\\
&\hat{\mathcal{L}}_2(\mathbf{Y})=\lambda  \mathrm{Tr}(\mathbf{V}_{t-1}^{\top}\mathbf{L}_{\mathbf{Y}}\mathbf{V}_{t-1}),\\
&\mathcal{L}=\mathcal{L}_1(\mathbf{Y})+\mathcal{L}_2(\mathbf{Y}),\\
&\hat{\mathcal{L}}=\mathcal{L}_1(\mathbf{Y})+\hat{\mathcal{L}}_2(\mathbf{Y}).
\end{align}
 Then the one-step gradient descent is given as
\begin{equation}\label{eq_update_Y}
\mathbf{Y}_t\longleftarrow \mathbf{Y}_{t-1}-\alpha \nabla\hat{\mathcal{L}}(\mathbf{Y})
\end{equation}
where $\alpha$ is the step size and $\nabla\hat{\mathcal{L}}$ denotes the derivative of $\mathcal{L}$ with respect to $\mathbf{Y}$. The procedures are summarized into Algorithm \ref{alg_opt}.

\algrenewcommand\algorithmicrequire{\textbf{Input:}}
\algrenewcommand\algorithmicensure{\textbf{Output:}}
\algdef{SE}[DOWHILE]{Do}{doWhile}{\algorithmicdo}[1]{\algorithmicwhile\ #1}%

\begin{algorithm}[t]
\caption{Optimization for \textsc{Lap}tSNE} 
\label{alg_opt}
\begin{algorithmic}[1]
\Require
dataset $\mathbf{X} \in \mathbb{R}^{N\times D}$, target dimension $d$, estimate of cluster numbers $\hat{k}$, tuning parameter $\lambda$, perplexity $\delta_{\text{{Perp}}}$, step size $\alpha$, maximum iteration $T$
\State compute pairwise affinities $p_{j|i}$ with perplexity $\delta_{\text{{Perp}}}$ (refer to \cite{van2008visualizing}) 
\State set $\mathbf{P_X} = \left\{p_{ij}\right\}_{(i,j)\in[N]\times[N]}$ where $p_{ij}=\frac{p_{j|i}+p_{i|j}}{2N}$
\State compute the graph Laplacian $\mathbf{L}_{\mathbf{X}}$ of $\mathbf{P_X}$
\State Initialization: $\mathbf{Y}_0 \longleftarrow [\mathbf{V}_1(\mathbf{L}_{\mathbf{X}}), \cdots, \mathbf{V}_d(\mathbf{L}_{\mathbf{X}})]$
	  \For{$t=1,2,\ldots,T$}  
	  	 \State compute low-dimensional affinities $\{q_{ij}\}_{(i,j)\in[N]\times[N]}$
	  	 \State compute gradient of KL divergence $\nabla\mathcal{L}_1(\mathbf{Y})$
	  	 \State compute gradient of regularization  $\nabla\hat{\mathcal{L}}_2(\mathbf{Y})$
	  	 \State set $\mathbf{Y}_t\longleftarrow \mathbf{Y}_{t-1}-\alpha \left(\nabla\mathcal{L}_1(\mathbf{Y}) + \nabla\hat{\mathcal{L}}_2(\mathbf{Y})\right)$
      \EndFor
\Ensure cluster-contractive embedding  $\mathbf{Y}_T \in \mathbb{R}^{N\times d}$
\end{algorithmic}
\end{algorithm}

\begin{proposition}
Suppose there exists a positive constant $L$ such that $\Vert\nabla\mathcal{L}(\mathbf{Y})-\nabla\mathcal{L}(\mathbf{Y}')\Vert_F\leq L\Vert \mathbf{Y}-\mathbf{Y}'\Vert_F$.  Let $\mathbf{Y}_0,\mathbf{Y}_1,\ldots,\mathbf{Y}_T$ be the sequence generated by Algorithm \ref{alg_opt} with $\alpha\leq\frac{2}{L}$. Then:\\
(a) $\mathcal{L}(\mathbf{Y}_t)-\mathcal{L}(\mathbf{Y}_{t-1})\leq \frac{2-\alpha L}{2\alpha}\left\Vert \mathbf{Y}_t-\mathbf{Y}_{t-1}\right\Vert_F\leq 0$; \\
(b) $\mathcal{L}(\mathbf{Y}_T)\leq \mathcal{L}(\mathbf{Y}_0)-\frac{2-\alpha L}{2\alpha}\sum_{t=1}^{T}\left\Vert \mathbf{Y}_t-\mathbf{Y}_{t-1}\right\Vert_F$.
\end{proposition}

Proposition 1 shows that \textsc{Lap}tSNE Algorithm \ref{alg_opt} proves to be convergent.  We defer the proof to Appendix \ref{append_proof}.  Note that the algorithm can be accelerated by using momentum, i.e.,
$$\mathbf{Y}_t \longleftarrow \mathbf{Y}_{t-1} - \alpha\left(\nabla\mathcal{L}_1(\mathbf{Y}) + \nabla\hat{\mathcal{L}}_2(\mathbf{Y})\right) + \beta(t)\left(\mathbf{Y}_{(t-1)}-\mathbf{Y}_{t-2}\right),$$
where $\beta(t)$ is the momentum parameter.  Proving the convergence of Algorithm 1 with momentum is out of the scope of this paper and can be an future work.


\subsection{Gradient of Laplacian-Composited Objective}\label{sec_grad}
It is non-trivial to compute the gradient of $\nabla\mathcal{L}(\mathbf{Y})$ because $\mathbf{Y}$ involves the symmetric normalized Laplacian.  Here we elaborate the computation.

First,  the gradient of $\mathcal{L}_2(\mathbf{Y})$ ($\lambda$ is omitted for convenience) with respect to $\mathbf{Y}$ can be expressed as
\begin{equation}\label{eq_ded_g0}
\begin{aligned} 
     \frac{\partial \mathrm{Tr}(\mathbf{V^{\top} LV})}{\partial \mathbf{Y}}
     = \sum_{m,n}\left (\sum_{i,j}  (\mathbf{VV^{\top}})_{ij}\frac{\partial \mathbf{L}_{ij}}{\partial q_{mn}}\right) \frac{\partial q_{mn}}{\partial \mathbf{Y}}.
\end{aligned}
\end{equation}
The symmetric normalized Laplacian matrix $\mathbf{L}$ is constructed via $\mathbf{Q_Y}$ (note that $q_{ii} = 0, \forall i$), and the matrix representation of $\mathbf{L}$ could be further simplified to the equations in the right:
\begin{align*} 
    \mathbf{L} &=\begin{bmatrix}
          1 - \frac{q_{11}}{(\sum_b q_{1b} \sum_a q_{a1})^{\frac{1}{2}}}& \cdots &- \frac{q_{1n}}{(\sum_b q_{1b}\sum_a q_{a1})^{\frac{1}{2}}}\\
            & \cdots & \\
           - \frac{q_{nn}}{(\sum_b q_{nb}\sum_a q_{an})^{\frac{1}{2}}} & \cdots & 1 - \frac{q_{nn}}{(\sum_b q_{nb}\sum_a q_{an})^{\frac{1}{2}}}
        \end{bmatrix}; \qquad \mathbf{L}_{ij} =
\begin{cases}
  1 - (\sum_b q_{ib})^{-\frac{1}{2}}q_{ij}(\sum_a q_{aj})^{-\frac{1}{2}} & i = j,\\
  - (\sum_b q_{ib})^{-\frac{1}{2}}q_{ij}(\sum_a q_{aj})^{-\frac{1}{2}} & i \neq j.
\end{cases}
\end{align*}


These equations will help us discuss the gradient of $\mathbf{L}_{ij}$ with respect to $q_{mn}$ in different situations. For each similarity $q_{mn}$, there are actually three gradient components of the graph Laplacian matrix ($\mathbf{L}$). It follows that,

\begin{equation}\label{eq_ded_1}
\begin{aligned} 
& \sum_{i,j} (\mathbf{VV^{\top}})_{ij}\frac{\partial L_{ij}}{\partial q_{mn}} \\
= &\sum_{i\neq m} (\mathbf{VV^{\top}})_{in}\frac{\partial L_{in}}{\partial q_{mn}} + \sum_{j\neq n} (\mathbf{VV^{\top}})_{mj}\frac{\partial L_{mj}}{\partial q_{mn}} +(\mathbf{VV^{\top}})_{mn}\frac{\partial L_{mn}}{\partial q_{mn}}\\
= & \sum_i \frac{1}{2}(\mathbf{VV^{\top}})_{in}(\sum_b q_{ib})^{-\frac{3}{2}}q_{in}(\sum_a q_{an})^{-\frac{1}{2}} \\
& +\sum_j \frac{1}{2}(\mathbf{VV^{\top}})_{mj}(\sum_b q_{mb})^{-\frac{1}{2}}q_{mj}(\sum_a q_{aj})^{-\frac{3}{2}}\\
& -\frac{1}{2}(\mathbf{VV^{\top}})_{mn}(\sum_b q_{mb})^{-\frac{1}{2}}(\sum_a q_{an})^{-\frac{1}{2}}.
\end{aligned}
\end{equation}

Next, we introduce three auxiliary matrices $\mathbf{U}^0$, $\mathbf{U}^1$,  and $\mathbf{U}^2$. They represent the different components of the formula above. Specifically, let
\begin{equation}\label{eq_ded_01}
\mathbf{U}^0 := 
\begin{bmatrix}
\cdots & | & \cdots\\
\cdots & u^0_n
& \cdots \\
\cdots & | & \cdots
\end{bmatrix}_{N\times N},~~\mathbf{U}^1: = 
\begin{bmatrix}
\cdots \\
(u^1_m)^{\top}
\\
\cdots
\end{bmatrix}_{N\times N}   
\end{equation}
where
$$u^0_n =\sum_i \frac{1}{2}(\mathbf{VV^{\top}})_{in}\tfrac{q_{in}}{(\sum_b q_{ib})^{\frac{3}{2}}(\sum_a q_{an})^{\frac{1}{2}}}\mathbf{1},  ~~n = 1, \cdots, N$$ and $$u^1_m =\sum_j \frac{1}{2}(\mathbf{VV^{\top}})_{mj} \tfrac{q_{mj}}{(\sum_b q_{mb})^{\frac{1}{2}}(\sum_a q_{aj})^{\frac{3}{2}}}\mathbf{1},~~m = 1, \cdots, N$$
with all-one vector $\mathbf{1} \in \mathbb{R}^N$.
Further we let
\begin{equation}\label{eq_ded_2}
\mathbf{U}^2 := \{ -\frac{1}{2}(\mathbf{VV^{\top}})_{mn}(\sum_b q_{mb})^{-\frac{1}{2}}(\sum_a q_{an})^{-\frac{1}{2}} \}_{m,n = 1}^N.    
\end{equation}
Then, we can factorize the chain of gradients into point-wise multiplication. Invoking \eqref{eq_ded_01} and \eqref{eq_ded_2} into \eqref{eq_ded_g0}, we obtain
\begin{equation}\label{eq_all}
\begin{aligned}
 \frac{\partial \mathrm{Tr}(\mathbf{V^{T}LV})}{\partial \mathbf{Y}}
& = \sum_{m,n} \frac{\partial \mathrm{Tr}(\mathbf{V^{\top}LV})}{\partial q_{mn}} \frac{\partial q_{mn}}{\partial \mathbf{Y}}   
 = \sum_{m,n} (\mathbf{U}^0_{mn} + \mathbf{U}^1_{mn} + \mathbf{U}^2_{mn}) \odot  \frac{\partial q_{mn}}{\partial \mathbf{Y}}.
\end{aligned}
\end{equation}

Based on different similarity measurement,  we discuss $\frac{\partial q_{mn}}{\partial \mathbf{Y}}$ accordingly. 
In this study, we consider Gaussian kernel and T-Student kernel, though the later has better performance in our numerical studies. For the Gaussian kernel ($\mathbf{K} = \{k_{mn}\}_{m,n=1}^N$) with a constant sigma $\sigma$, the derivative of each row in $\mathbf{Y}$ is
\begin{equation}\label{eq_gaussion_q}
   \frac{\partial k_{mn}}{\partial \mathbf{y}_a} = 
   \begin{cases}
     k_{mn} \cdot (-\frac{1}{2\sigma^2}) \cdot 2(y_a - y_m) & n = a,\\
     k_{mn} \cdot (-\frac{1}{2\sigma^2}) \cdot 2(y_a - y_n) & m = a.
   \end{cases}
\end{equation}
Since $q_{mn} =k_{mn}$, substituting \eqref{eq_gaussion_q} into \eqref{eq_all}, we have
\begin{equation}\label{eq_ded_dy_gk}
  \frac{\partial \mathrm{Tr}(\mathbf{V^{\top}LV})}{\partial \mathbf{Y}} =  \frac{2}{\sigma^2}\left ( \mathbf{U}\mathbf{Y} - \mathbf{C}\odot \mathbf{Y}  \right), 
\end{equation}
where $\mathbf{U} := (\mathbf{U}^0 + \mathbf{U}^1 + \mathbf{U}^2)\odot \mathbf{K} \in \mathbb{R}^{N\times N}$ and $\mathbf{C} := \mathbf{U}\mathbf{1}_{N\times d} \in \mathbb{R}^{N\times d}$.

If we replace Gaussian kernel by T-Student kernel ($\mathbf{T} = \{t_{ij}\}_{i,j=1}^N$) and only take the numerator part as the similarity measure, then $q_{mn} = t_{mn} = (1+\left \| \mathbf{y}_m - \mathbf{y}_n \right \|^2)^{-1}$. The latent matrix $\mathbf{U} \in \mathbb{R}^{N\times N}$ takes the form of
\begin{equation}\label{eq_t}
\mathbf{U} = (\mathbf{U}^0 + \mathbf{U}^1 + \mathbf{U}^2) \odot \mathbf{T} \odot \mathbf{T}. 
\end{equation}
Then we arrive at
\begin{equation}
\frac{\partial \mathrm{Tr}(\mathbf{V^{\top}LV})}{\partial \mathbf{Y}} = 2(\mathbf{U}\mathbf{Y} - \mathbf{C} \odot \mathbf{Y})
\end{equation}
where $\mathbf{C} = \mathbf{U}\mathbf{1}_{N\times d}  \in \mathbb{R}^{N\times d}$.  

It is worth mentioning that the computation of the gradient we have introduced can be adapted to other problems where the eigenvalues of the symmetric normalized Laplacian are functions of decision variables. 

\subsection{Out-of-Sample and Large-Scale Extensions of \textsc{Lap}tSNE}\label{sec_ose}
The quadratic time and space complexities of \textsc{Lap}tSNE and t-SNE prevents their application to large-scale datasets.  In addition, it is difficult to use the learned models of them to reduce the dimension of new data.  Here we provide an out-of-sample extension for \textsc{Lap}tSNE. 

Suppose we have already reduced the dimension of $\mathbf{X}$ to 2 using our \textsc{Lap}tSNE and we want to reduce the dimension of some new samples $\mathbf{X}_{\text{new}}$ drawn from the same distribution as $\mathbf{X}$. There is no need to perform  \textsc{Lap}tSNE on $\{\mathbf{X}, \mathbf{X}_{\text{new}}\}$ again. We can just learn a nonlinear mapping $f$ from $\mathbf{x}$ to $\mathbf{y}$ using neural networks on $\{\mathbf{X},\mathbf{Y}\}$, where $\mathbf{Y}$ denotes the 2-D embedding given by \textsc{Lap}tSNE. Then we perform $f$ on $\mathbf{X}_{\text{new}}$ to get the low-dimensional embedding $\mathbf{Y}_{\text{new}}$. 

It is worth mentioning that the out-of-sample extension can also be applied to handle large-scale datasets.  Specifically, we perform k-means clustering (with a large enough $k$, e.g. 1000) on a large dataset $\mathbf{X}$ to get the cluster centers, which are regarded as some landmark data points. Then we perform \textsc{Lap}tSNE on the landmark points and use the out-of-sample extension to get the low-dimensional embedding of the large dataset. See Algorithm \ref{alg_large}.

\begin{algorithm}[t]
\caption{Large-scale extension of \textsc{Lap}tSNE} 
\label{alg_large}
\begin{algorithmic}[1]
\Require
dataset $\mathbf{X} \in \mathbb{R}^{N\times D}$, target dimension $d$, number of landmark points $S$
\State perform k-means to partition $\mathbf{X}$ into $S$ clusters
\State form a dataset $\mathbf{X}_S\in\mathbf{R}^{S\times D}$ using the data points in $\mathbf{X}$ closest to the $S$ cluster centers
\State perform \textsc{Lap}tSNE (Algorithm \ref{alg_opt}) on $\mathbf{X}_S$ to get $\mathbf{Y}_S\in\mathbf{R}^{S\times d}$
\State train a neural network (with parameters $\mathcal{W}$) $f_{\mathcal{W}}:\mathbb{R}^D\rightarrow\mathbb{R}^d$ using $\{\mathbf{X}_S,\mathbf{Y}_S\}$
\State obtain $\mathbf{Y}=f_{\mathcal{W}}(\mathbf{X})$
\Ensure cluster-contractive embedding  $\mathbf{Y} \in \mathbb{R}^{N\times d}$
\end{algorithmic}
\end{algorithm}

\subsection{Mini-Batch Extension of \textsc{Lap}tSNE}
Besides out-of-sample extension, mini-batch optimization could also serve as an alternative to reduce the computational complexity on especially large datasets. During the optimization procedure, instead of updating the entire $\mathbf{Y}$, we can just update a few rows of $\mathbf{Y}$ in each iteration. Specifically, at iteration $t$, let $\mathbf{Y}_{\Omega_t}\in\mathbb{R}^{m\times d}$ be a mini batch of $\mathbf{Y}$ indexed by $\Omega_t\subset [N]$ and let $\Omega_t^{\kappa}$ be the index set of the $\kappa$-nearest neighbors (according to $\mathbf{P_X}$) of $\mathbf{Y}_{\Omega_t}$ in $\mathbf{Y}_{[N]\slash\Omega_t}$. Denote $\bar{\Omega}_t=\Omega_i\cup\Omega_t^{\kappa}$. Then we update $\mathbf{Y}_{\bar{\Omega}_t}$ at iteration $t$
such that the local-connectivity in the original data space can be preserved:
\begin{align}
\mathbf{Y}_{\bar{\Omega}_t} \leftarrow \mathbf{Y}_{\bar{\Omega}_{t}} -\alpha \nabla\hat{\mathcal{L}}(\mathbf{Y}_{\bar{\Omega}_t}).
\end{align}
Note that the computation of the gradient $\nabla\hat{\mathcal{L}}(\mathbf{Y}_{\bar{\Omega}_t})$ involves evaluating $\mathbf{Q}_{\mathbf{Y}}$, which can be implemented via updating the rows and columns of $\mathbf{Q}_{\mathbf{Y}}$ with indices ${\bar{\Omega}_t}$. Since $\vert {\bar{\Omega}_t}\vert\leq \kappa m\ll N$, the computational complexity is reduced significantly. We call this approach \textsc{Lap}tSNE-Mini.

\subsection{Hyperparameters and Complexity Analysis}\label{sec_hyper_complex}
As described in Algorithm \ref{alg_opt},  compared with tSNE,  it seems that  our \textsc{Lap}tSNE algorithm has  two more hyperparameters $\hat{k}$ (estimated number of potential clusters)  and $\lambda$ (the regularization weight serving as contractive strength).  Actually,  LaptSNE at the same time eliminated a few hyperparameters required in t-SNE,  such as the exaggerate stage and the corresponding exaggeration rate.  Besides, the foregoing hyperparameters are hard to adjust and can lead to unconvergence in t-SNE.

Note that in \textsc{Lap}tSNE, $\hat{k}$ is not necessarily equal or very close to the true number of clusters $k$.  When $\hat{k}<k$,  \textsc{Lap}tSNE only shrinks partial clusters. When $\hat{k}>k$, the overestimated number of classes  makes \textsc{Lap}tSNE try to explore the nuanced subclass information from the original categories.  If the subclass structures are not significant,  the minimization for the eigenvalues $\sigma_{k+1},\ldots,\sigma_{\hat{k}}$ will has less impact on the result.  With a moderate $\hat{k} \approx k$, one may observe a clearly cluster-contractive structure in the low-dimensional embeddings. For the effect of contractive strength $\lambda$, suppose the data have been reduced to a well-clustered low-dimensional representation, the contractive strength is not necessarily large. Whereas the low-dimensional representations are over\textsc{Lap}ped, \textsc{Lap}tSNE requires a larger hyperparmeter $\lambda$ to shrink data into different groups. 

According to Section \ref{sec_opt} and Section \ref{sec_grad},  the time complexity (per iteration) and space complexity  of our \textsc{Lap}tSNE are $O(\hat{k} N^2)$ and $O(DN+N^2)$ respectively.  Note that we only need to compute the smallest $\hat{k}$ eigenvalues and eigenvectors of $\mathbf{L}$, which is efficient.  In  Table \ref{time},  we report the time costs of \textsc{Lap}tSNE and \textsc{Lap}tSNE-Mini in comparison to t-SNE on COIL20 and COIL100 datasets. Indeed, \textsc{Lap}tSNE and \textsc{Lap}tSNE-Mini are slower than t-SNE but they can provide better dimensionality reduction results, which will be shown in Section \ref{sec_exp}.

\begin{table}[h]
\centering
\caption{Running time comparison between \textsc{Lap}tSNE and t-SNE over 100 iterations
}
\label{time}
\begin{tabular}{l|ll}
\toprule
Dataset    & COIL20      & COIL100     \\ \hline
\textsc{Lap}tSNE & 60.4s  & 14876.7s \\
\textsc{Lap}tSNE-Mini & 20.1s & 887.0s \\
t-SNE   & 8.4s & 147.9s  \\
\bottomrule
\end{tabular}
\label{tcom}
\end{table}

\section{Experiments}\label{sec_exp}
\subsection{Datasets, Baselines, and Evaluation Metrics}
We test our \textsc{Lap}tSNE with T-Student kernel on seven real-world datasets detailed below.
\begin{itemize}
\item\textbf{Waveform} \cite{breiman2017classification} dataset contains 5,000 samples generated from 3 classes of waves with 21-dimensional attributes, all of which include noise.
\item \textbf{PenDigits} \cite{buitinck2013api} is a set of 1,797 grayscale images of digits. Each image is an 8x8 image which we directly flatten the pixels into a 64 dimensional vector.
\item\textbf{COIL 20} \cite{nene1996columbia} is a set of 1,440 grayscale images consisting of 20 objets under 72 different rotations. We flatten each image (128x128) into a 16,384 dimensional vector.
\item\textbf{COIL 100} \cite{nene1996columbia-2} is a set of 7,200 color images consisting of 100 objects under 72 different rations. Each image is a color image of size 128x128. We convert the images to grayscale and resize them to 32x32. Then we flatten each image into a 1,024 dimensional vector.
\item\textbf{HAR} \cite{reyes2013human} (Human Activity Recognition) is a dataset of 10,299 instances with 561 dimensional attributes. It was built from the recordings of 30 subjects performing activities of daily living.
\item\textbf{MNIST} \cite{deng2012mnist} is a dataset of 28x28 pixel grayscale images of handwritten digits. There are 10 digit classes (from 0 to 9) and 70,000 images in total. We treat these images as 784 dimensional vectors.
\item\textbf{Fashion-MNIST} \cite{xiao2017fashion} is a dataset of 28x28 pixel grayscale images of 10 kinds of fashion items, such as clothing and bags. There are 70,000 images. We treat them as 784 dimensional vectors.
\end{itemize}

\begin{figure}[t!]
\centering
    \centering
    \begin{subfigure}[t]{0.25\textwidth}
        \centering
        \includegraphics[height=1.2in]{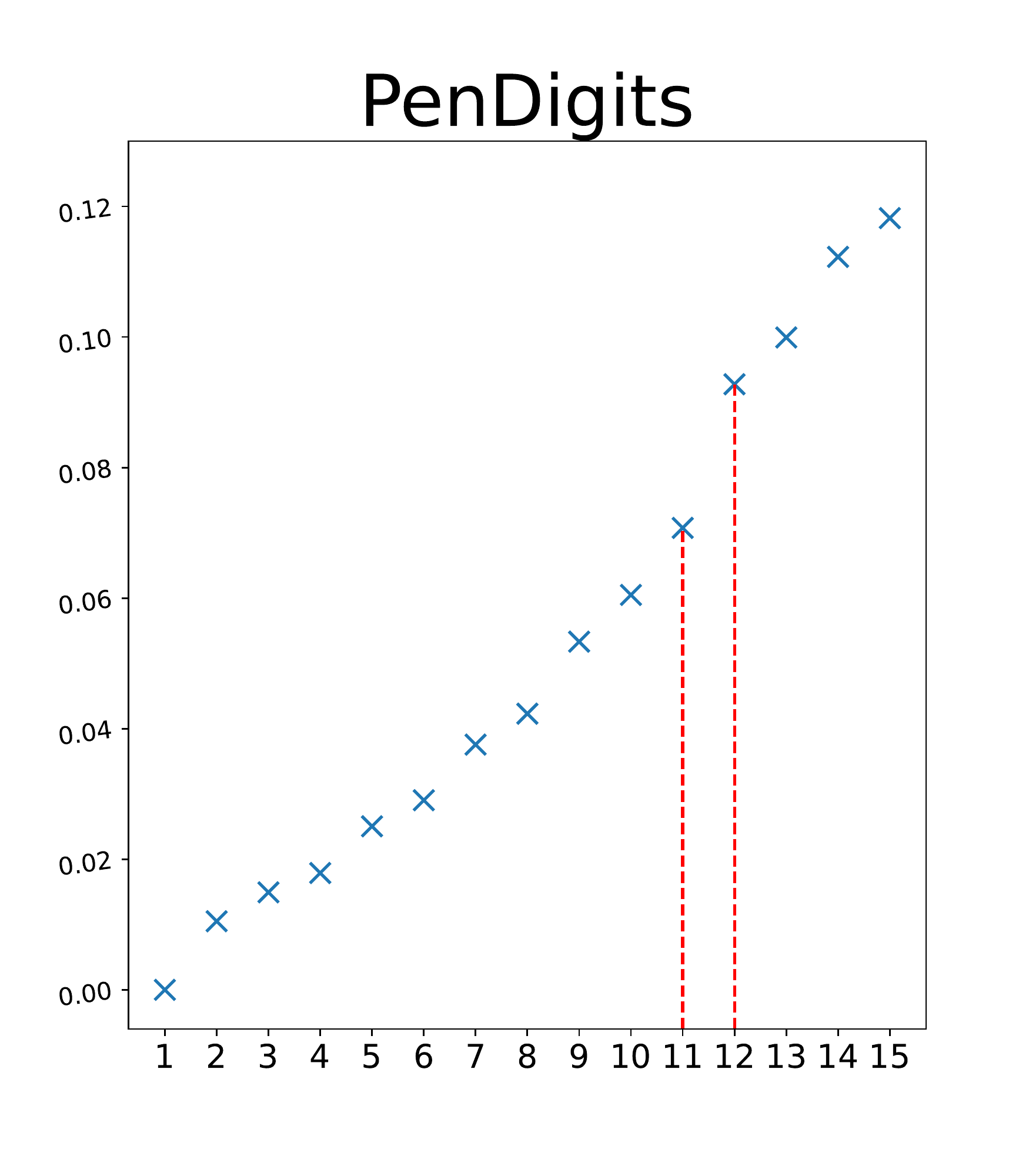}
    \end{subfigure}%
    ~ 
    \begin{subfigure}[t]{0.25\textwidth}
        \centering
        \includegraphics[height=1.2in]{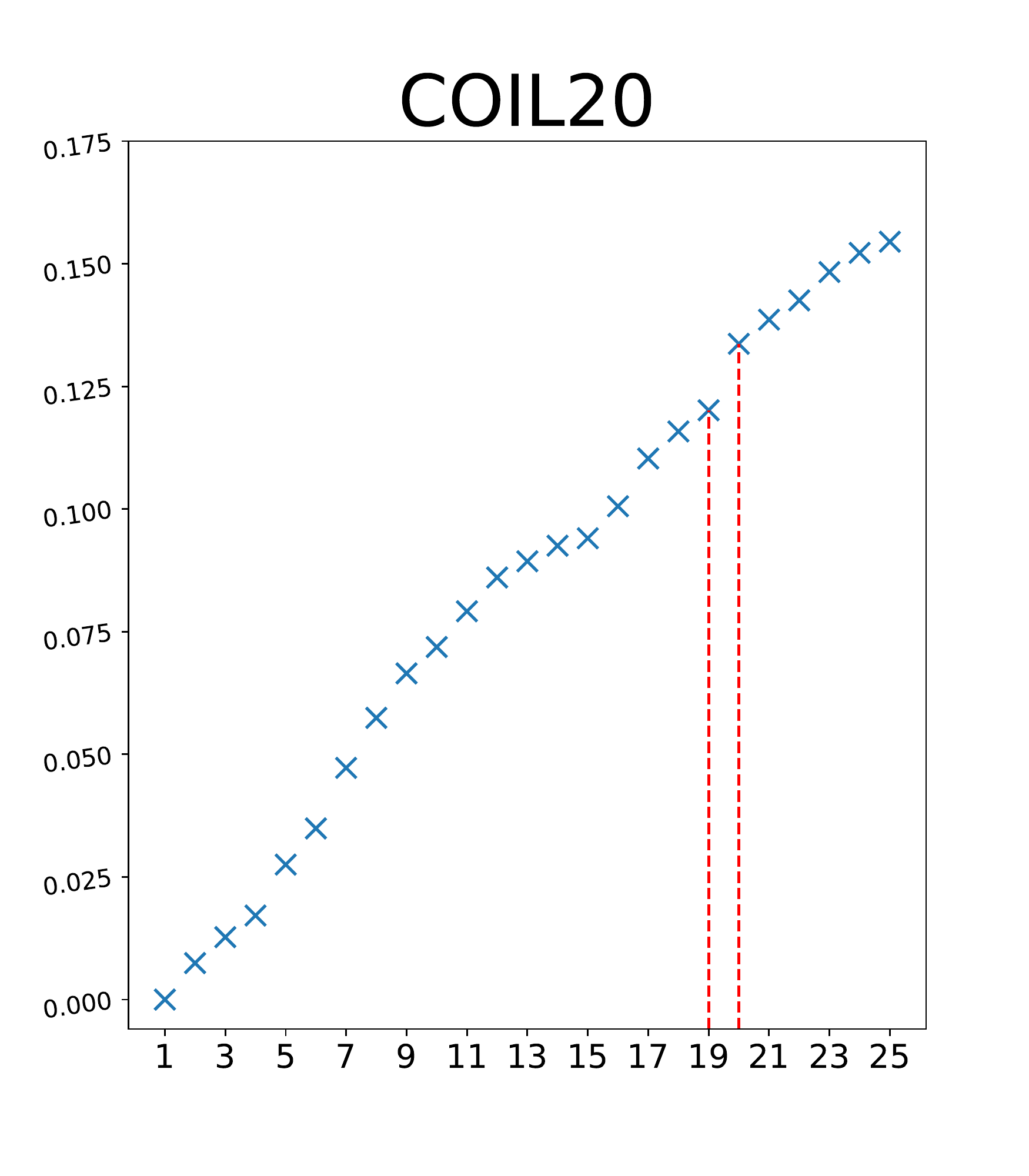}
    \end{subfigure}
    ~ 
    \begin{subfigure}[t]{0.25\textwidth}
        \centering
        \includegraphics[height=1.2in]{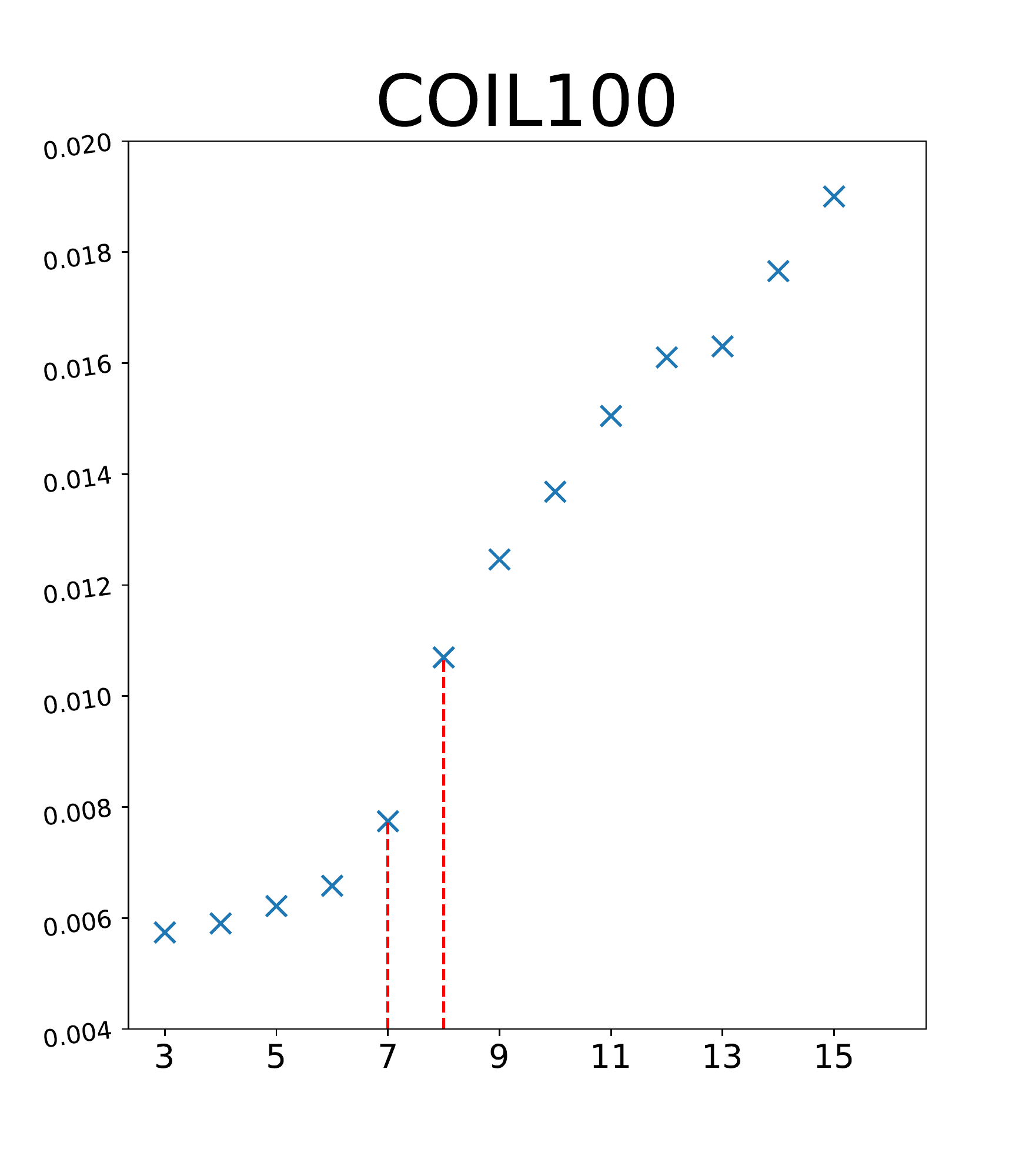}
    \end{subfigure}
    ~ 
    \begin{subfigure}[t]{0.25\textwidth}
        \centering
        \includegraphics[height=1.2in]{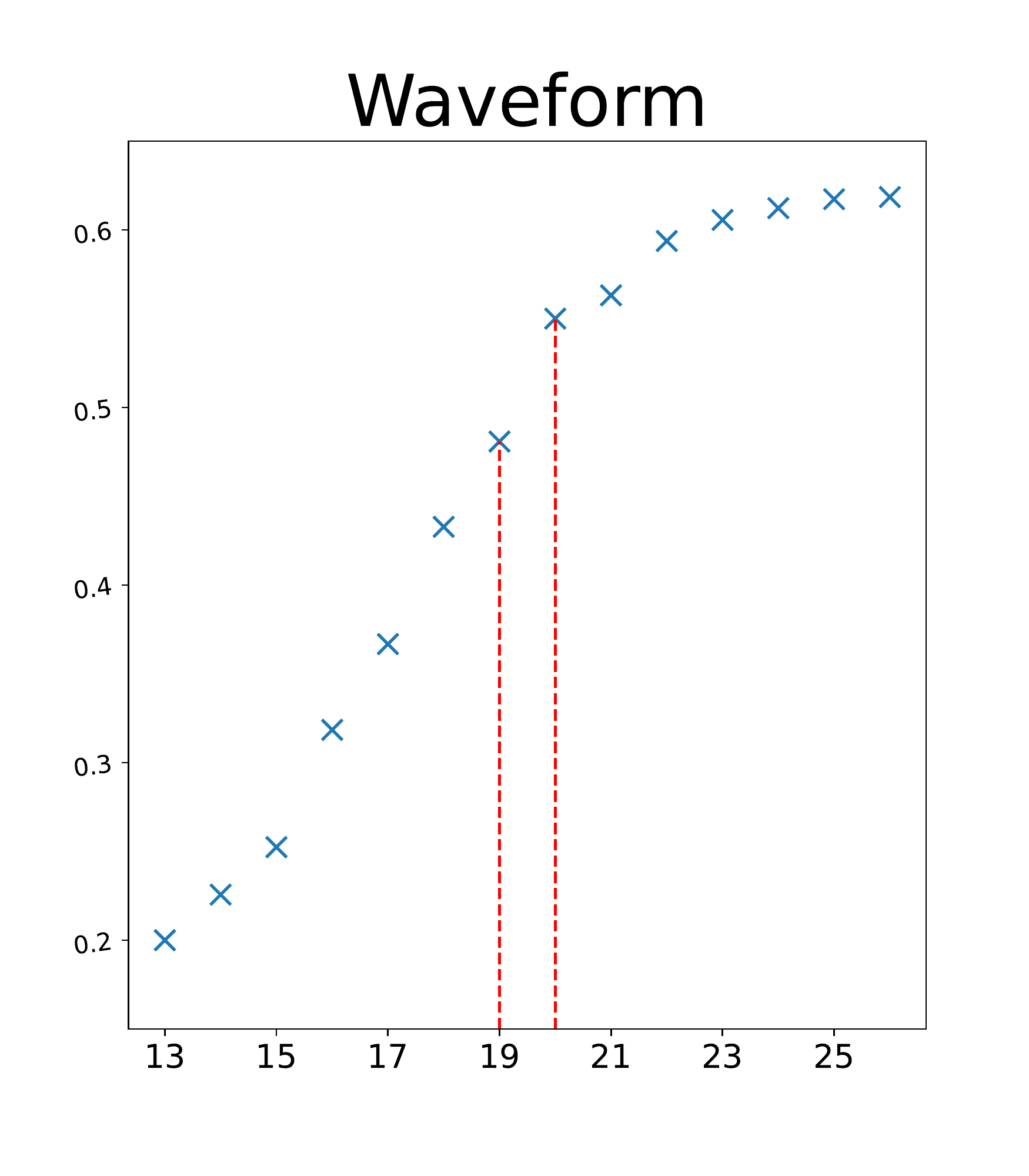} 
    \end{subfigure}
    \\
       \begin{subfigure}[t]{0.25\textwidth}
        \centering
        \includegraphics[height=0.9in]{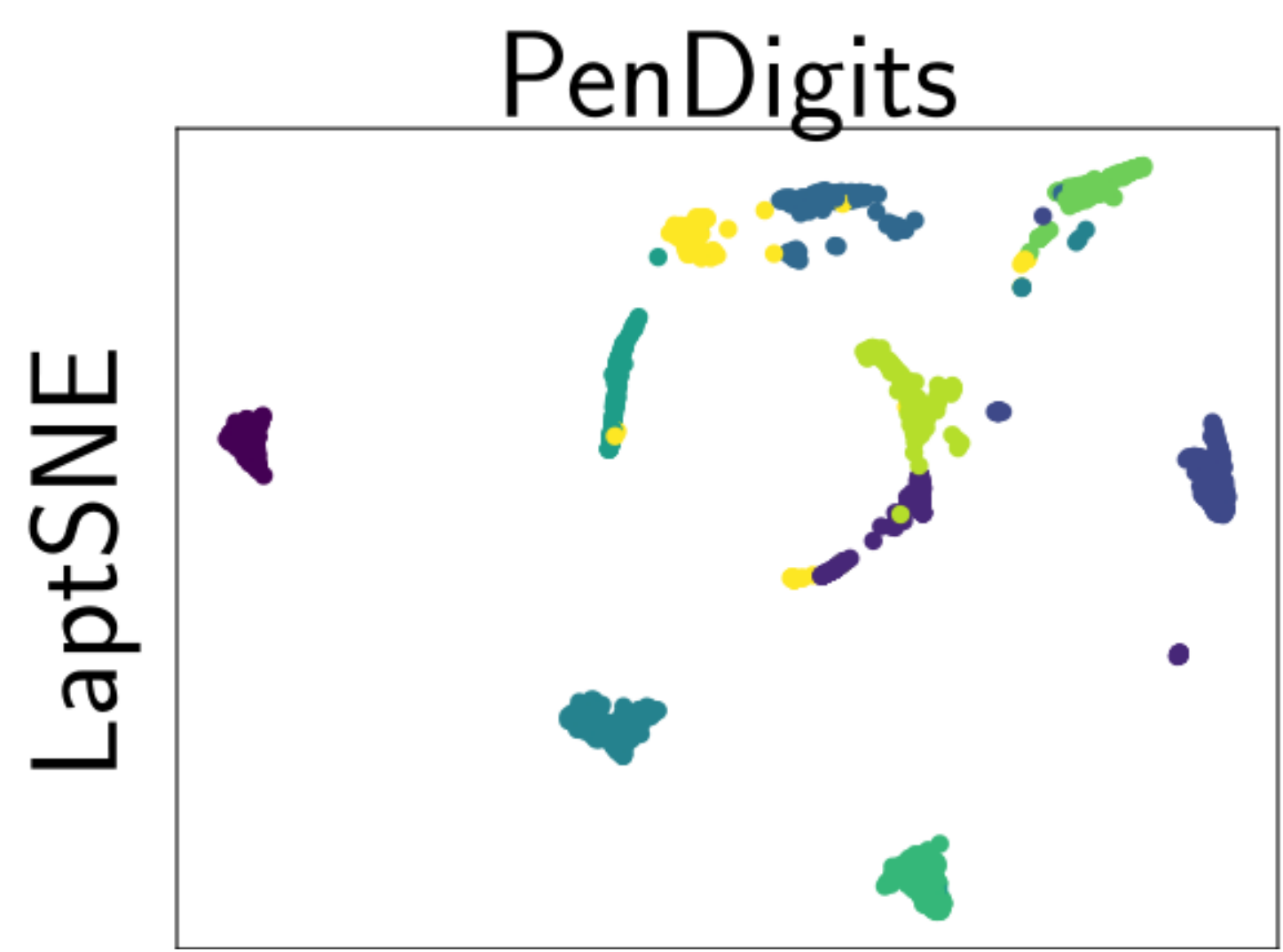}
    \end{subfigure}%
    ~ 
    \begin{subfigure}[t]{0.25\textwidth}
        \centering
        \includegraphics[height=0.9in]{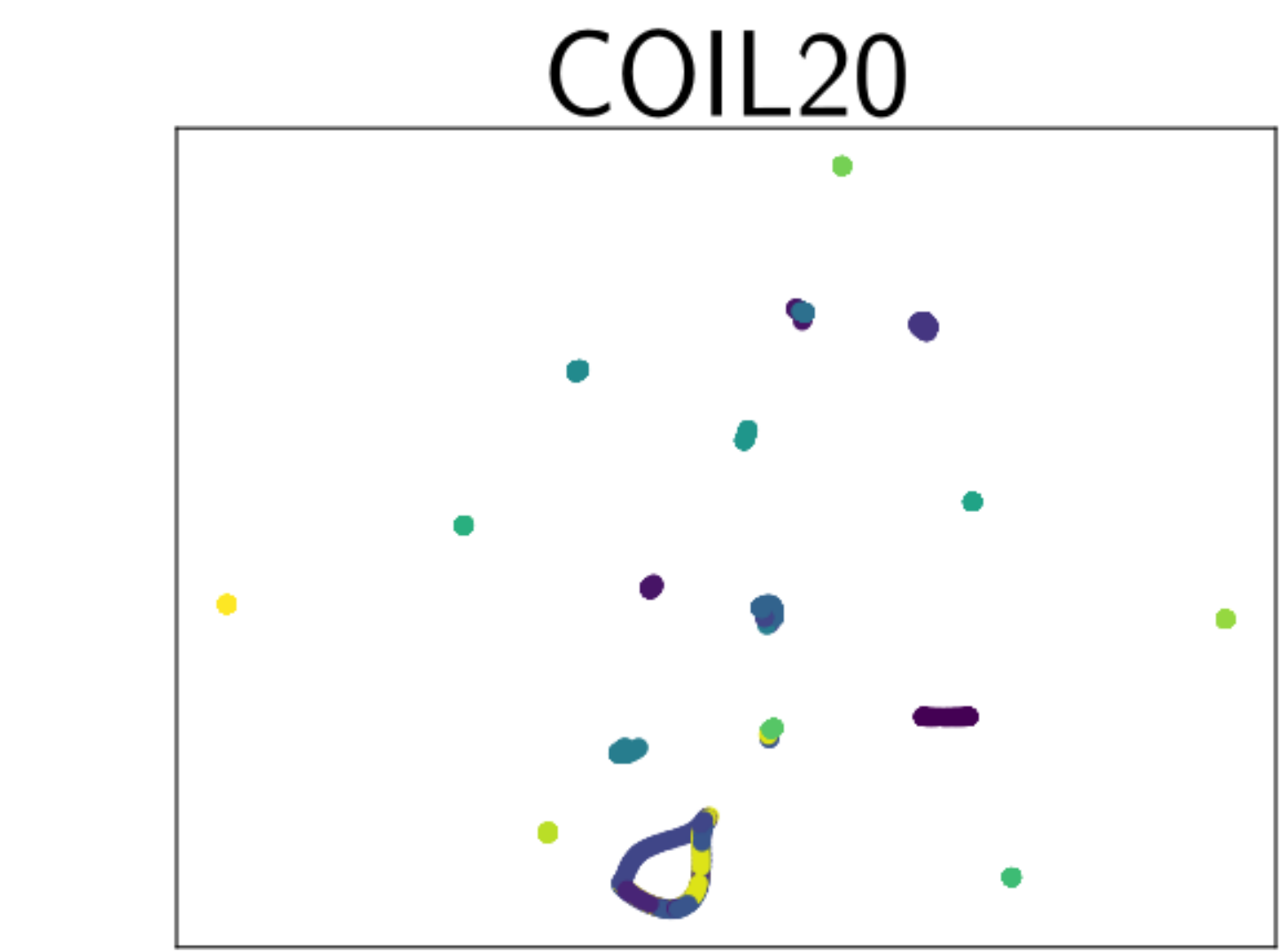}
    \end{subfigure}
    ~ 
    \begin{subfigure}[t]{0.25\textwidth}
        \centering
        \includegraphics[height=0.9in]{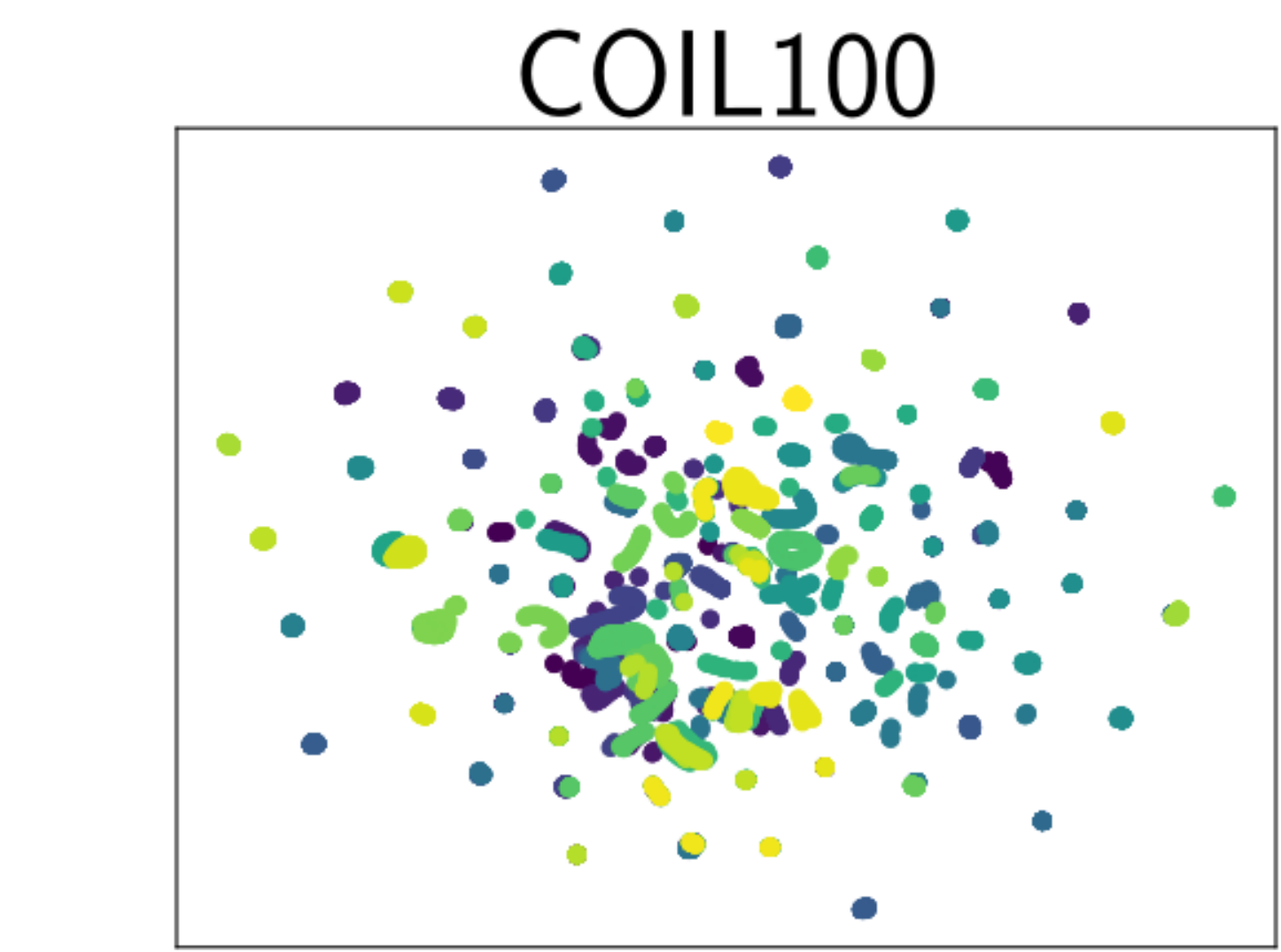}
    \end{subfigure}
    ~ 
    \begin{subfigure}[t]{0.25\textwidth}
        \centering
        \includegraphics[height=0.9in]{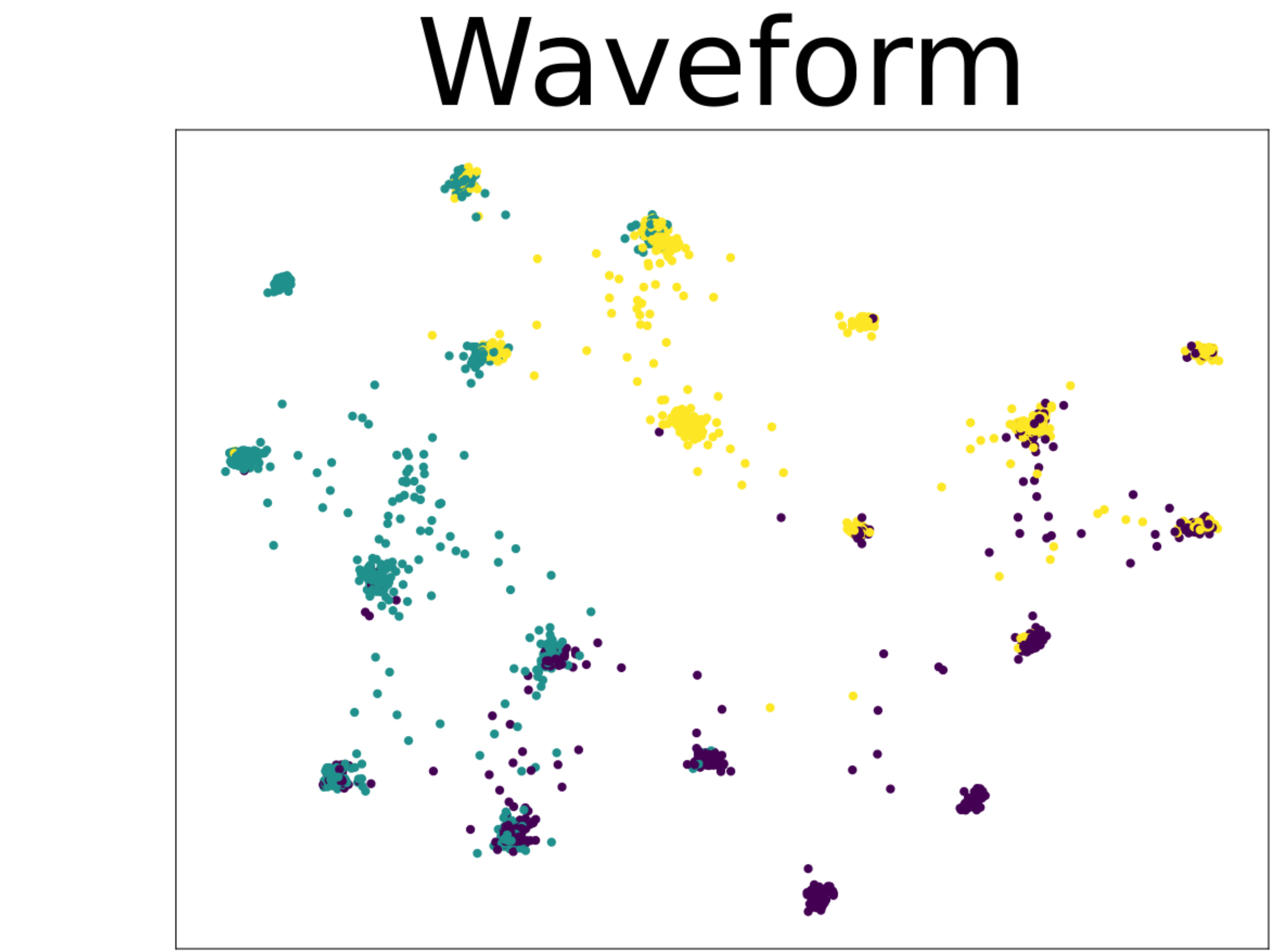}
    \end{subfigure}
    \\
       \begin{subfigure}[t]{0.25\textwidth}
        \centering
        \includegraphics[height=0.9in]{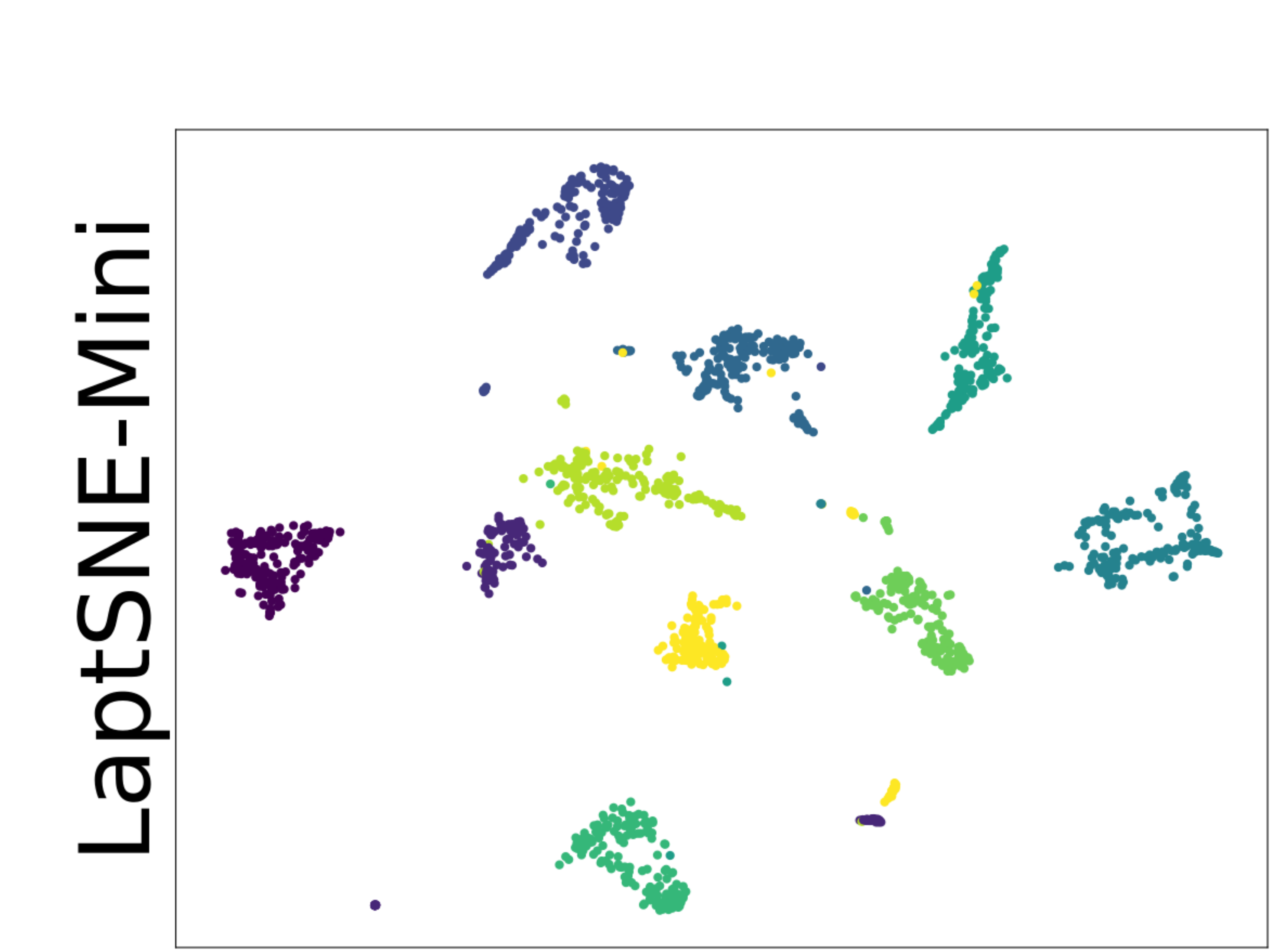}
    \end{subfigure}%
    ~ 
    \begin{subfigure}[t]{0.25\textwidth}
        \centering
        \includegraphics[height=0.9in]{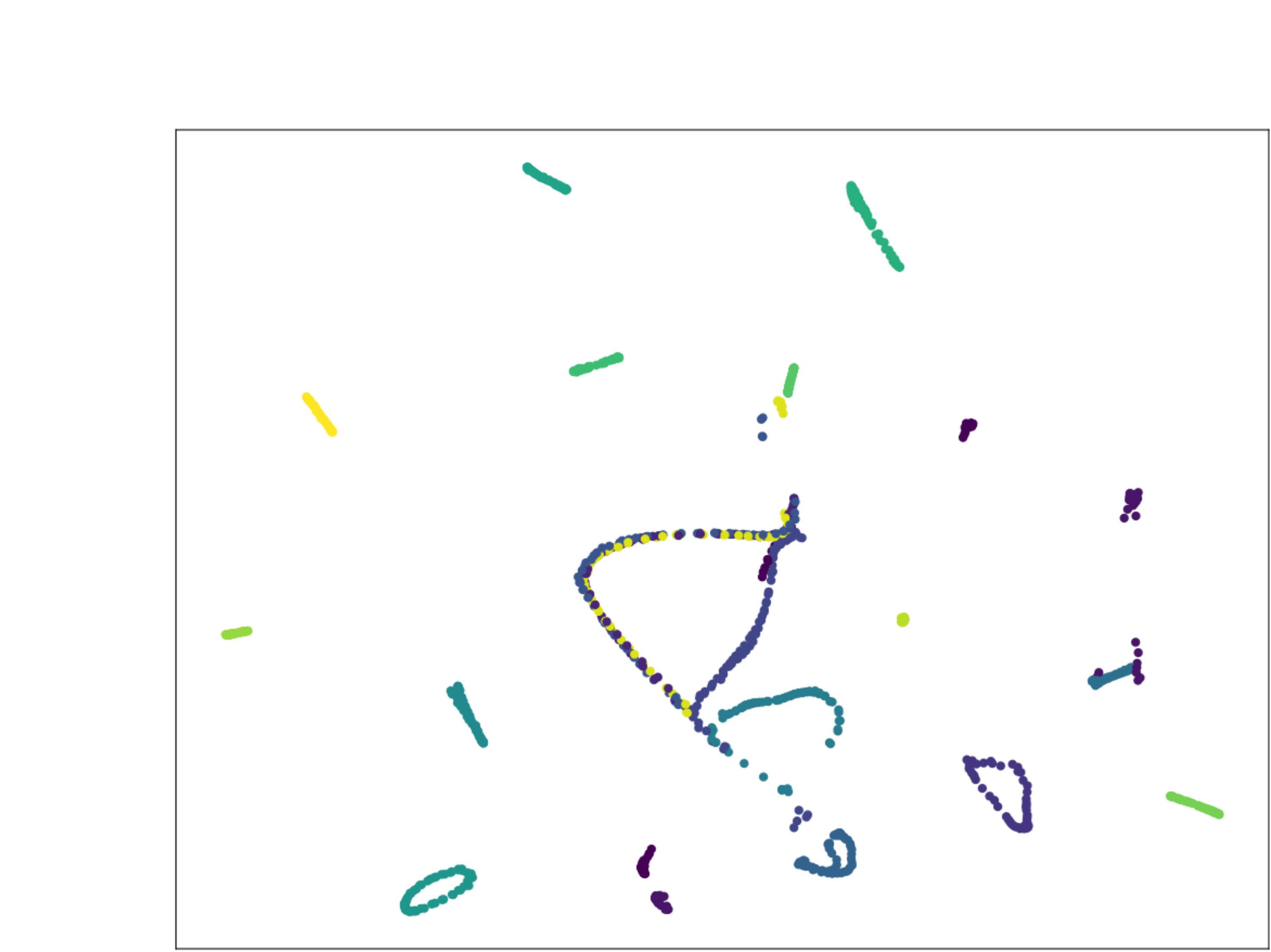}
    \end{subfigure}
    ~ 
    \begin{subfigure}[t]{0.25\textwidth}
        \centering
        \includegraphics[height=0.9in]{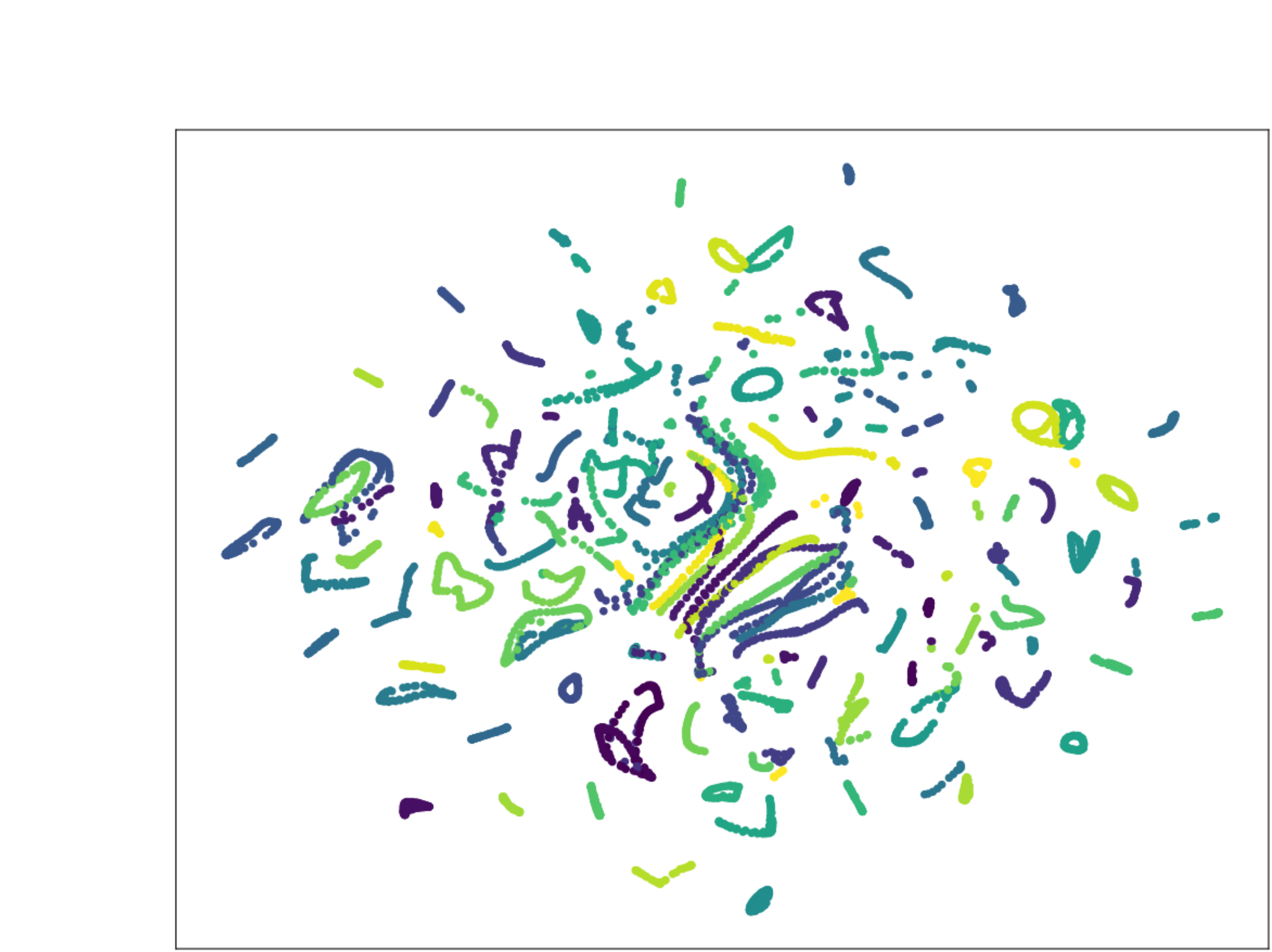}
    \end{subfigure}
    ~ 
    \begin{subfigure}[t]{0.25\textwidth}
        \centering
        \includegraphics[height=0.9in]{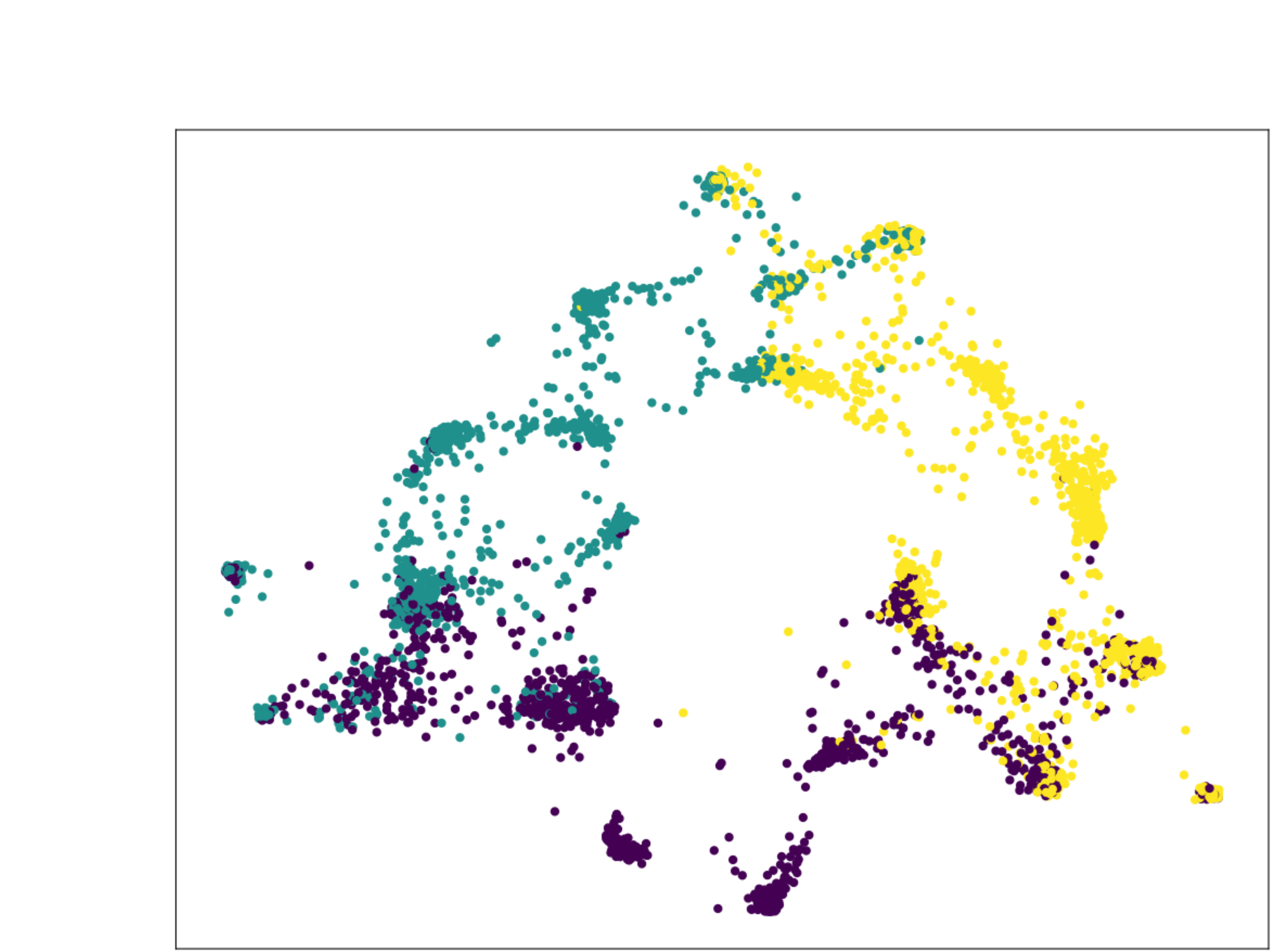}
    \end{subfigure}
    \\
       \begin{subfigure}[t]{0.25\textwidth}
        \centering
        \includegraphics[height=0.9in]{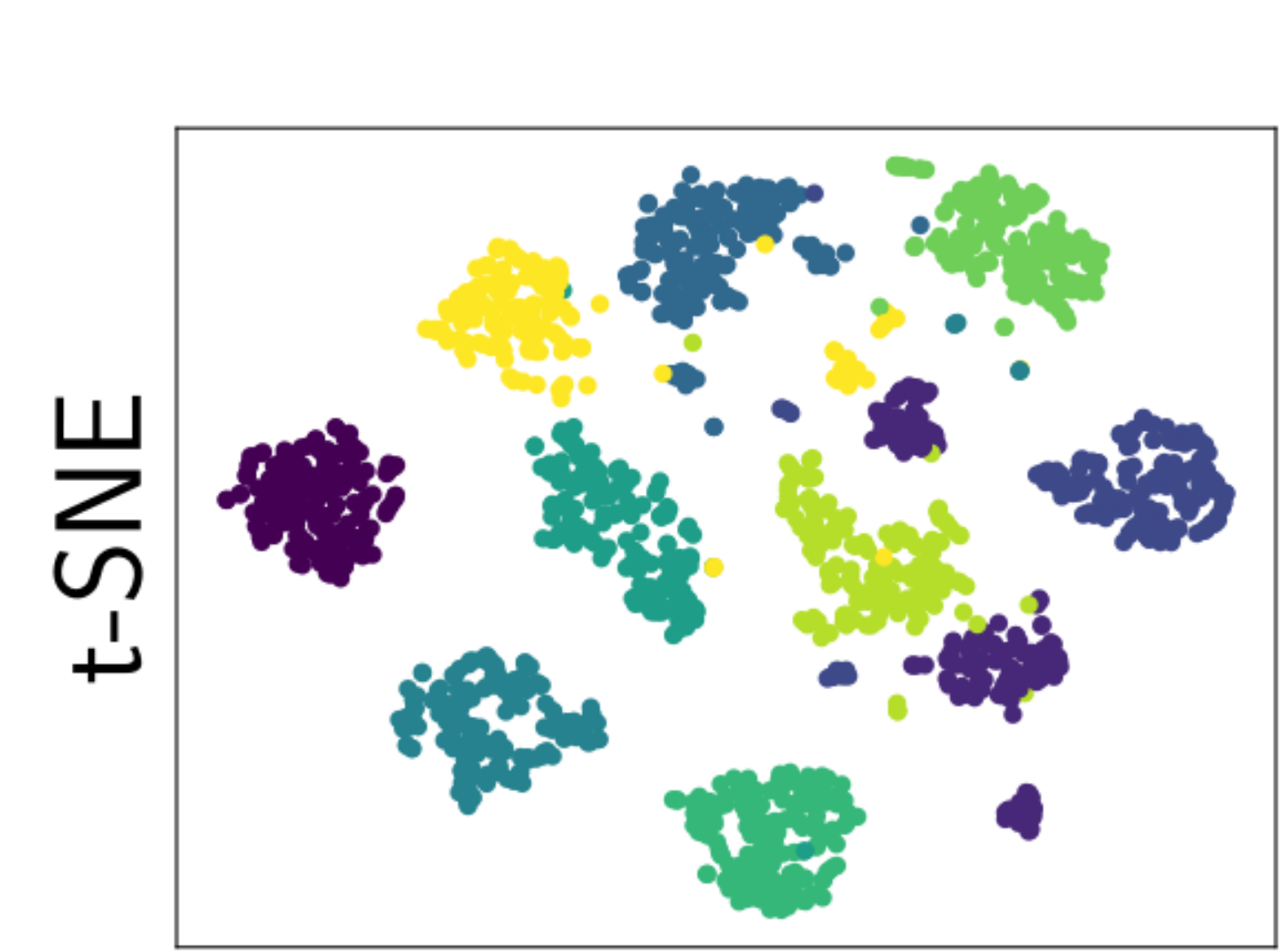}
    \end{subfigure}%
    ~ 
    \begin{subfigure}[t]{0.25\textwidth}
        \centering
        \includegraphics[height=0.9in]{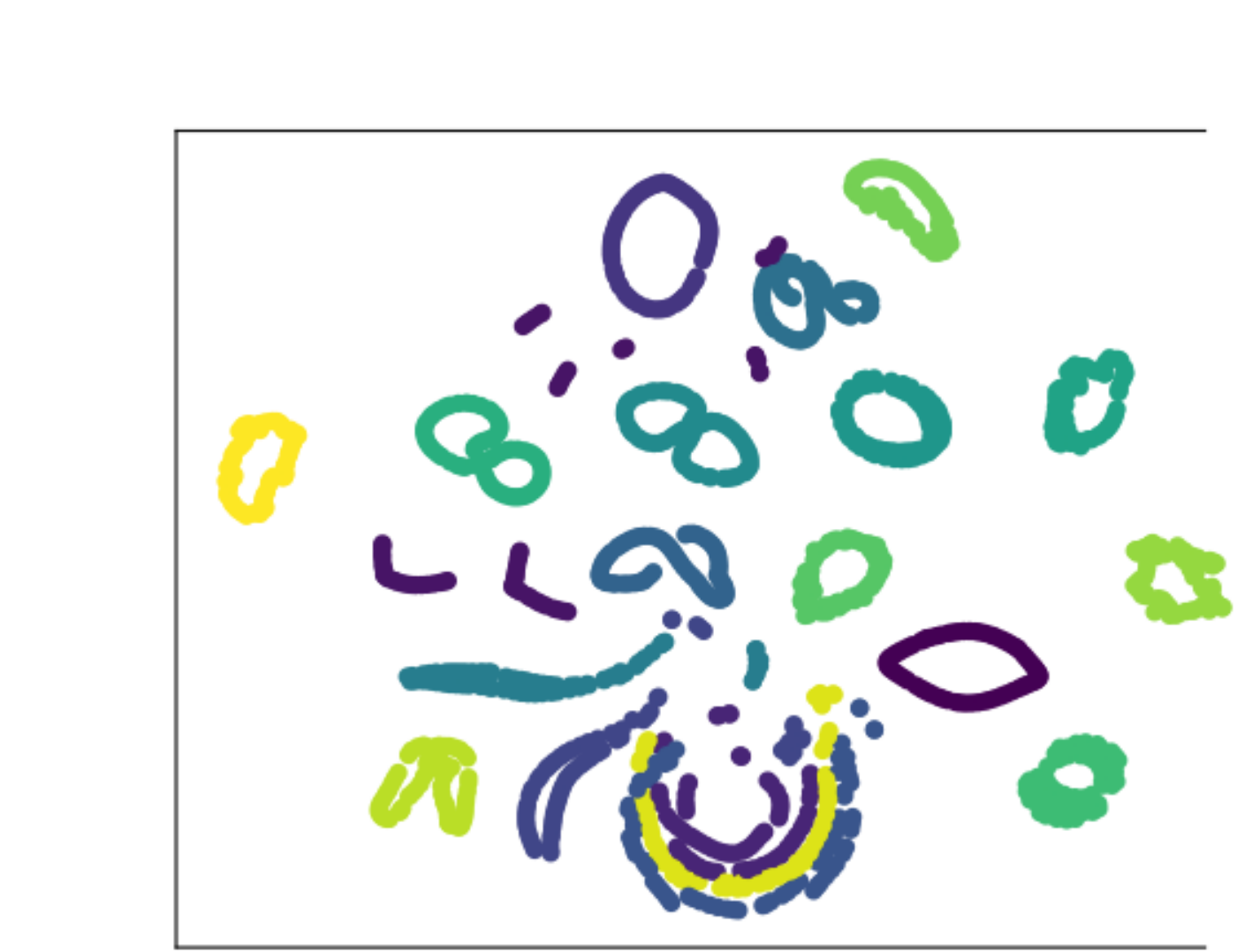}
    \end{subfigure}
    ~ 
    \begin{subfigure}[t]{0.25\textwidth}
        \centering
        \includegraphics[height=0.9in]{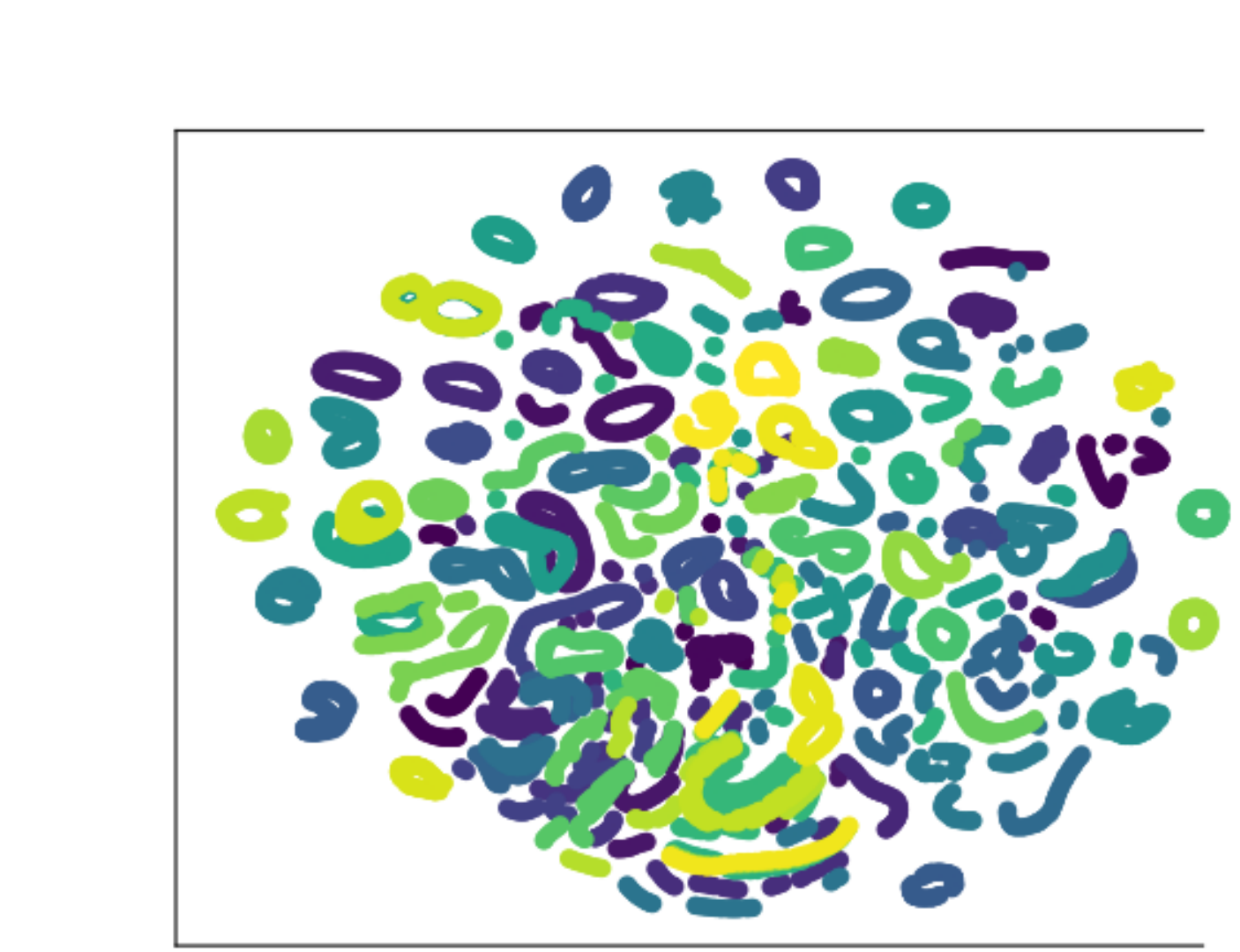}
    \end{subfigure}
    ~ 
    \begin{subfigure}[t]{0.25\textwidth}
        \centering
        \includegraphics[height=0.9in]{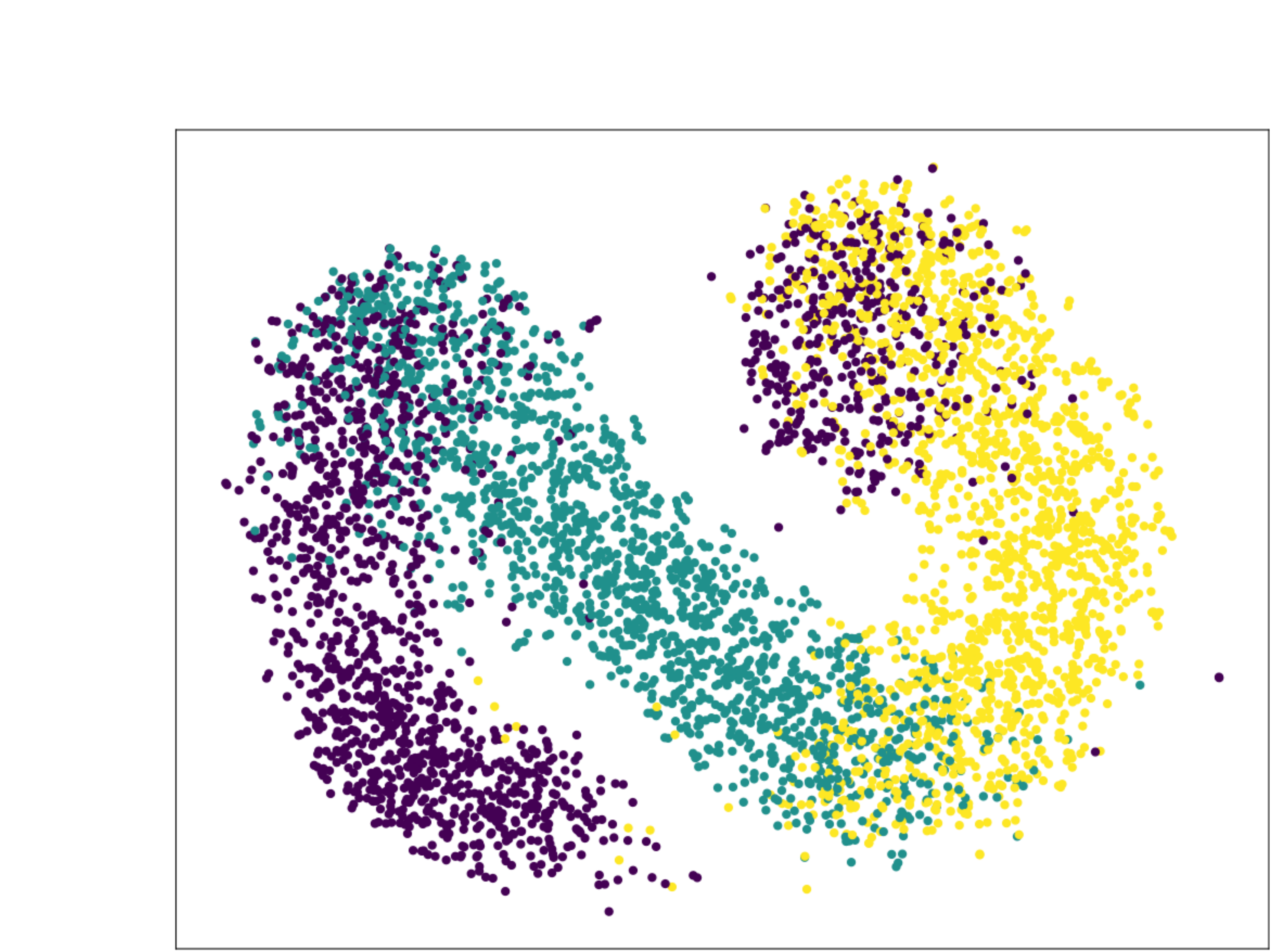}
    \end{subfigure}
  \caption{Row 1: First smallest eigenvalues of graph Laplacian for data $\mathbf{X}$ (From left to right: PenDigits, 11-th gap is large; COIL20, 19-th gap is large; COIL100, 7-th gap is large; Waveform, 19-th gap is large). Row 2-4: A qualitative comparison of \textsc{Lap}tSNE, \textsc{Lap}tSNE-Mini and t-SNE in visualizing the PenDigits, COIL20, COIL100 and Waveform datasets}
  \label{fig:small dataset}
\end{figure}

We see that Waveform, COIL20, COIL100 and PenDigits are small datasets with no more than 10,000 samples while HAR, MNIST and Fashion-MNIST are large datasets with more than ten thousand samples.  We will show the results on the small datasets and large datasets separately.
The proposed \textsc{Lap}tSNE and \textsc{Lap}tSNE-Mini are compared with the following baselines.
\begin{itemize}
    \item \textbf{Vanilla t-SNE} \cite{van2008visualizing}: We implement t-SNE using the TSNE module in scikit-learn package \cite{scikit-learn}. To ensure a fair comparison, we remove the trick of exaggeration stage and set desired perplexity and number of iterations the same as \textsc{Lap}tSNE.
    \item \textbf{UMAP} \cite{mcinnes2018umap}: We perform UMAP visualization using the latest umap-learn package \cite{mcinnes2018umap-software}. Specifically, we remain the parameters as default and set number of iterations the same as \textsc{Lap}tSNE.
    \item \textbf{Laplacian Eigenmaps} \cite{belkin2003laplacian}: We implement it as SpectralEmbedding module in the scikit-learn package \cite{scikit-learn}, where parameters are set as default.
    \item \textbf{PCA} \cite{jolliffe2016principal}: We use the LAPACK implementation of full SVD in the scikit-learn package \cite{scikit-learn}.
\end{itemize}
The implementation of our \textsc{Lap}tSNE and \textsc{Lap}tSNE-Mini are based on the scikit-learn package and gradient descent optimization.

\begin{table}[t]
\footnotesize
\centering
\caption{Comparison between \textsc{Lap}tSNE, \textsc{Lap}tSNE-Mini, t-SNE, UMAP, Eigenmaps and PCA on PenDigits, COIL20, COIL100 and Waveform: 1) $k$-NN classifier accuracy for different values of $k$ \cite{mcinnes2018umap}; 2) clustering performances in terms of NMI, SC and DBI based on the hyperparameter $\hat{k}$. \textit{Note: $\uparrow$ means the higher is better, whereas $\downarrow$ indicates the lower is better}. The best value in each case is highlighted in bold.}
\label{table:small dataset}
\begin{tabular}{m{1em}c|cccccc}
\toprule
   {\begin{sideways}\textbf{~~}\end{sideways}} & \textbf{score} & \textbf{\textsc{Lap}tSNE}  & \textbf{\textsc{Lap}tSNE-Mini} &\textbf{t-SNE}  & \textbf{UMAP} & \textbf{Eigenmaps} & \textbf{PCA}  \\ 
\cline{2-8}
\rotatebox{90}{PenDigits}& 
\makecell{10-nn $\uparrow$\\ 20-nn $\uparrow$\\ 40-nn $\uparrow$\\ 80-nn $\uparrow$\\NMI $\uparrow$\\SC $\uparrow$\\ DBI $\downarrow$} &
\makecell{ \textbf{0.990} \\ 0.987 \\ \textbf{0.975}\\ \textbf{0.970}\\ 
\textbf{0.9084} \\
\textbf{0.7535} \\ \textbf{0.4064}}   & 
\makecell{0.989\\\textbf{0.989}\\0.966\\0.952\\0.8793 \\ 0.6742 \\0.4531} &

\makecell{0.977\\0.973\\0.956\\0.948\\ 
0.7148\\
0.4754\\ 0.7121} &

\makecell{0.988 \\ 0.983 \\ 0.972 \\ 0.956\\
0.8981\\
0.6010 \\ 0.5688} &
\makecell{0.828 \\ 0.816 \\ 0.805 \\ 0.785\\
0.7869 \\
0.6976 \\ 0.4532} &
\makecell{ 0.710 \\ 0.682 \\ 0.671\\ 0.660 \\
0.5267 \\
0.3936 \\ 0.7992}   

\\
\cline{2-8}
\rotatebox{90}{COIL20}   & 
\makecell{10-nn $\uparrow$\\ 20-nn $\uparrow$\\ 40-nn $\uparrow$\\ 80-nn $\uparrow$\\NMI $\uparrow$\\SC $\uparrow$\\ DBI $\downarrow$} &
\makecell{\textbf{0.986} \\  \textbf{0.943}  \\ \textbf{0.909} \\ \textbf{0.860}\\\ 
\textbf{0.8723} \\ \textbf{0.7947} \\ \textbf{0.3251}} &

\makecell{0.897\\0.895\\0.874\\0.844\\0.8723 \\ 0.7947 \\ \textbf{0.3251}} & 

\makecell{0.934 \\ 0.901 \\0.857\\ 0.789\\ 0.8286 \\ 0.5105 \\0.6866} &

\makecell{0.901 \\ 0.885 \\ 0.877 \\ 0.822\\ 0.8489 \\ 0.5689 \\ 0.5997 } &
\makecell{0.773 \\ 0.744 \\ 0.694 \\ 0.638\\ 0.5527 \\ 0.5760 \\0.6394} &
\makecell{0.754 \\  0.732  \\ 0.677 \\ 0.578  \\ 0.6391 \\ 0.4750 \\ 0.7287 }  
\\
\cline{2-8}
\rotatebox{90}{COIL100}   & 
\makecell{10-nn $\uparrow$\\ 20-nn $\uparrow$\\ 40-nn $\uparrow$\\ 80-nn $\uparrow$\\NMI $\uparrow$\\SC $\uparrow$\\ DBI $\downarrow$} &
\makecell{\textbf{0.937} \\  0.900   \\ 0.860 \\ \textbf{0.798}\\ 0.8601 \\ \textbf{0.6026} \\ \textbf{0.5497}} & 

\makecell{0.936\\\textbf{0.907}\\\textbf{0.861}\\0.792\\ 0.8778 \\ 0.6001 \\ 0.5827} &

\makecell{0.894  \\  0.856 \\0.808 \\ 0.718\\ 0.8694 \\ 0.5682 \\ 0.5841} &

\makecell{0.832 \\ 0.810 \\ 0.779 \\ 0.735\\ \textbf{0.8785} \\ 0.5126 \\0.6507} &

\makecell{0.633 \\ 0.582 \\ 0.524 \\ 0.468\\ 0.6727 \\ 0.4280 \\ 0.7500} &

\makecell{0.626 \\  0.59   \\ 0.538 \\ 0.483\\ 0.4440 \\ 0.4253 \\ 0.7473}  
\\ 
\cline{2-8}
\rotatebox{90}{Waveform}   & 
\makecell{10-nn $\uparrow$\\ 20-nn $\uparrow$\\ 40-nn $\uparrow$\\ 80-nn $\uparrow$\\NMI $\uparrow$\\SC $\uparrow$\\ DBI $\downarrow$} &
\makecell{\textbf{0.859} \\  0.850   \\ \textbf{0.845} \\ 0.838\\ \textbf{0.3863} \\ 0.4658 \\ 0.7692 } & 

\makecell{\textbf{0.859}\\\textbf{0.853}\\0.844\\\textbf{0.841}\\  0.3686 \\ 0.4811 \\ 0.6980} &

\makecell{0.849  \\  0.842 \\0.836 \\ 0.838\\ 0.3506 \\ 0.4624 \\ 0.7349} &

\makecell{0.850 \\ 0.849 \\ 0.843 \\ \textbf{0.841}\\ 0.3709 \\ 0.5121 \\ 0.6549} &

\makecell{0.822\\ 0.834\\ 0.820\\ 0.812 \\ 0.3679 \\ \textbf{0.6821} \\ \textbf{0.4113}} &
\makecell{0.846 \\0.838 \\ 0.833\\ 0.832 \\ 0.3605 \\ 0.4997 \\ 0.6755}   
\\
\bottomrule
\end{tabular}
\end{table}

We will present both qualitative results and quantitative results in the comparison studies.
We consider the following evaluation metrics.
\begin{itemize}
\item \textbf{$k$-nearest neighbor classifier accuracy} It measures the quantitative performance of the preservation of cluster information in the original space. With the hyper-parameter $k$ varying, we can also consider how structure preservation changes from purely local to more global. When computing the errors, the labels collected beforehand are assumed to contain the inherent cluster information. The metric has been used in many previous works of dimensionality reduction such as \cite{mcinnes2018umap}.
\item \textbf{Normalized Mutual Information (NMI)}\cite{estevez2009normalized} NMI is a widely-used metric for evaluating the performance of clustering algorithm. It scales between 0 (no mutual information) and 1 (perfect correlation). In this study, we compute NMI based on the ground-truth label and the clustering result of k-means on the low-dimensional embedding $\mathbf{Y}$, where k is set as the number of actual classes.
\item \textbf{Silhouette Coefficient (SC)} \cite{rousseeuw1987silhouettes} SC ranges from -1 to 1, measuring how distinct or well-separated a cluster is from other clusters. The score of SC is higher when clusters are dense and well separated. Similar to NMI, we compute SC using the result of k-means on $\mathbf{Y}$.
\item \textbf{Davies-Bouldin Index (DBI)} \cite{davies1979cluster} DBI calculates the ratio of within-cluster and between-cluster distances, and therefore, lower the score the better separation there is between clusters. Similar to NMI, we compute DBI using the result of k-means on $\mathbf{Y}$.
\end{itemize}

 
\subsection{Results on Small Datasets}

The eigenvalues of graph Laplacian in the original space implies the potential number of clusters. Therefore, we first investigate into the eigenvalue sequence of the Laplacian matrix corresponding to $\mathbf{P_X}$ in the original data space. For Waveform, PenDigits, COIL20 and COIL100 datasets, the Gaussian kernel with perplexity of 25 was used to construct $\mathbf{P_X}$. 
As shown in the first row of Figure \ref{fig:small dataset}, for COIL20, the gap between eigenvalues 19 and 20 are more distinguishable than others,
although the gaps between eigenvalues 6 and 7 and between eigenvalues 8 and 9 are also large. As analyzed in Section \ref{sec_hyper_complex}, we prefer an overestimated cluster number and hence we set $\hat{k}=19$ for COIL20, even though the true cluster number is 20. Similarly, the estimated number of potential clusters in Waveform, PenDigits and COIL100 datasets are 19, 11 and 7 respectively. We do not require $\hat{k}=k$ because it is unrealistic in practice and our \textsc{Lap}tSNE is not sensitive to $\hat{k}$. We will illustrate this point via using a wide range of $\hat{k}$ later.

As for the results, we first plot the two dimensional embedding colored with ground-truth label in Figure \ref{fig:small dataset}.  We claim that the quality of embedding produced by \textsc{Lap}tSNE and \textsc{Lap}tSNE-Mini are better than t-SNE for the four small datasets,  although the three algorithms are all powerful in preserving the local and global structures from the original space.  \textsc{Lap}tSNE and \textsc{Lap}tSNE-Mini can shrink the clusters into compact clusters in a powerful cluster-contractive manner. 

In Table \ref{table:small dataset}, we quantitatively compare \textsc{Lap}tSNE, \textsc{Lap}tSNE-Mini, t-SNE, UMAP, Eigenmaps and PCA embedding with respect to the four evaluation metrics. We see that \textsc{Lap}tSNE performs remarkably better on PenDigits and COIL20 among all the dimensional reduction methods. It performs at least as well as t-SNE on COIL100. Note that on Waveform, in terms of the SC and DBI metrics, Eigenmaps outperforms other methods, but in terms of NMI, Eigenmaps is outperformed by our \textsc{Lap}tSNE. One possible reason is that SC and DBI are not effective enough to quantify the with-class and between-class differences of this dataset. In addition, NMI is more reliable than SC and DBI because it utilizes the true labels.

Evidently, the embedding quality of \textsc{Lap}tSNE is better compared with t-SNE and UMAP at both local and non-local scales in terms of PenDigits and COIL20. As for COIL100, it provides largely comparable performance in embedding at local scales, but performs superior at non-local scales.
Besides, as shown in Table \ref{tcom} and Table \ref{table:small dataset}, the proposed mini-batch based \textsc{Lap}tSNE could remarkably save the computational time while maintaining good performance. In this way, \textsc{Lap}tSNE-Mini could be considered as an efficient alternative when dealing with large datasets.

\subsection{Results on Large Datasets}

\begin{figure}[t!]
\centering
    \centering
    \begin{subfigure}[t]{0.2\textwidth}
        \centering
        \includegraphics[height=1.3in]{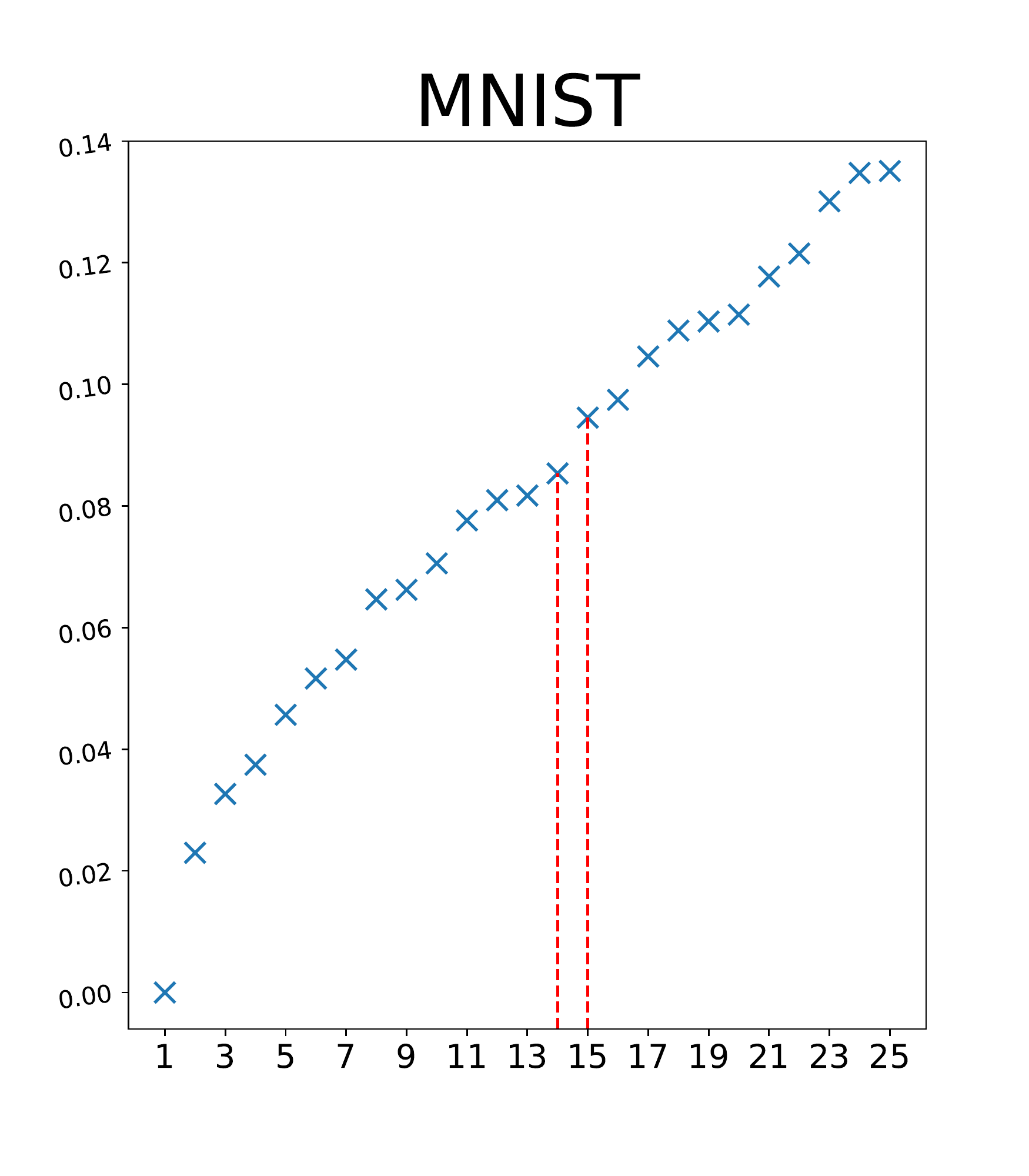}
    \end{subfigure}%
    ~ 
    \begin{subfigure}[t]{0.2\textwidth}
        \centering
        \includegraphics[height=1.3in]{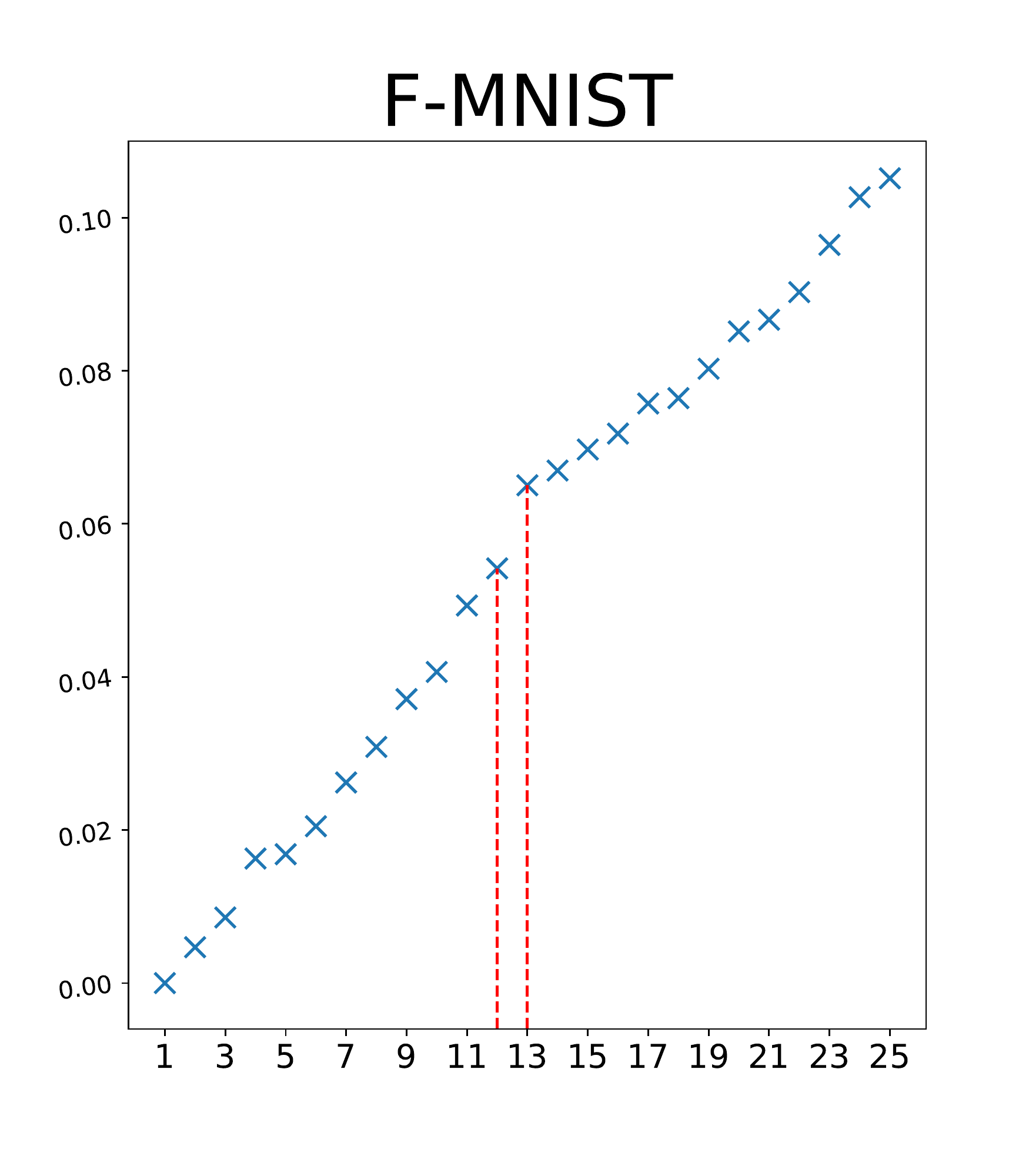}
    \end{subfigure}
    ~ 
    \begin{subfigure}[t]{0.2\textwidth}
        \centering
        \includegraphics[height=1.3in]{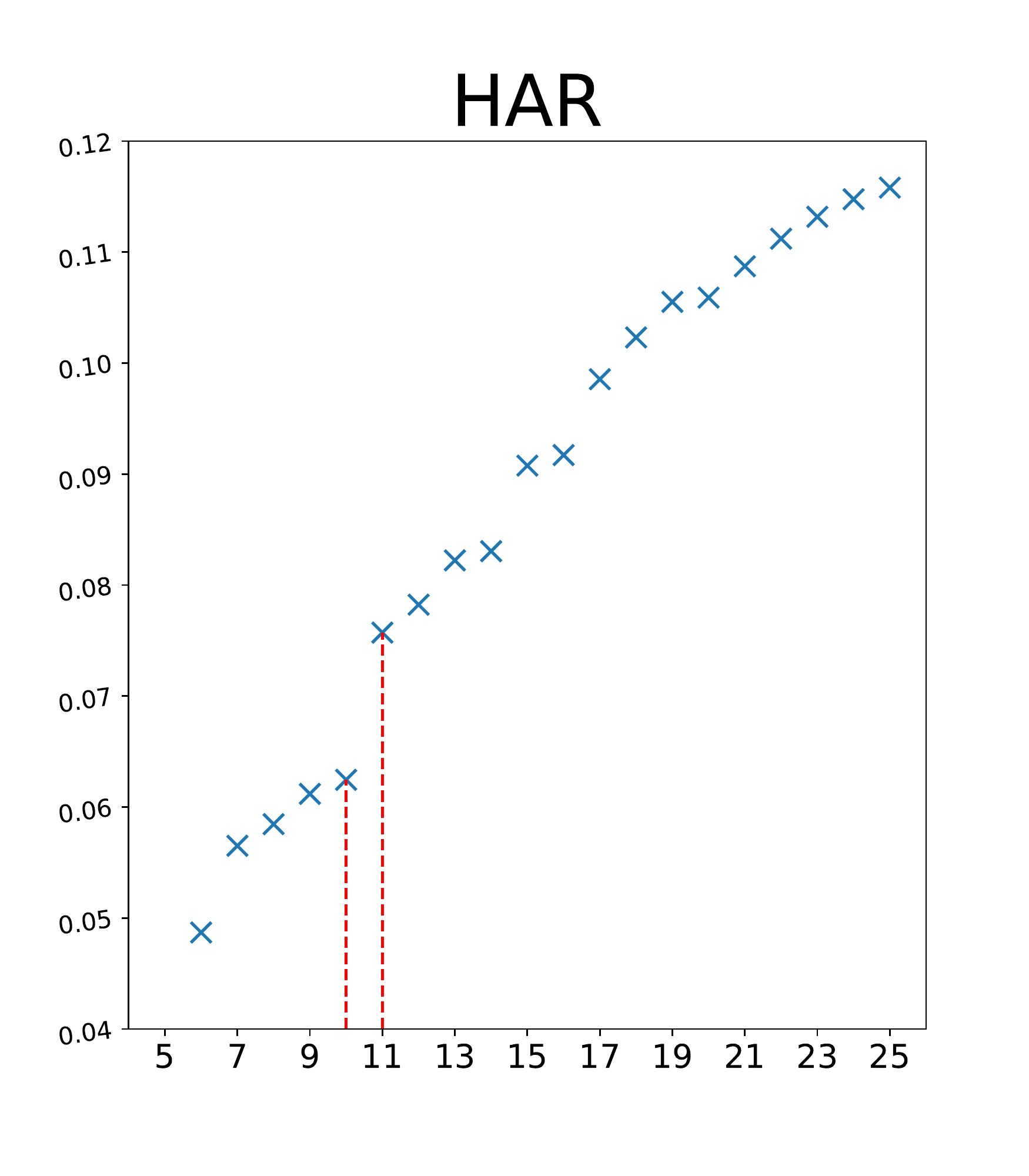}
    \end{subfigure}
    \\
       \begin{subfigure}[t]{0.2\textwidth}
        \centering
        \includegraphics[height=0.9in]{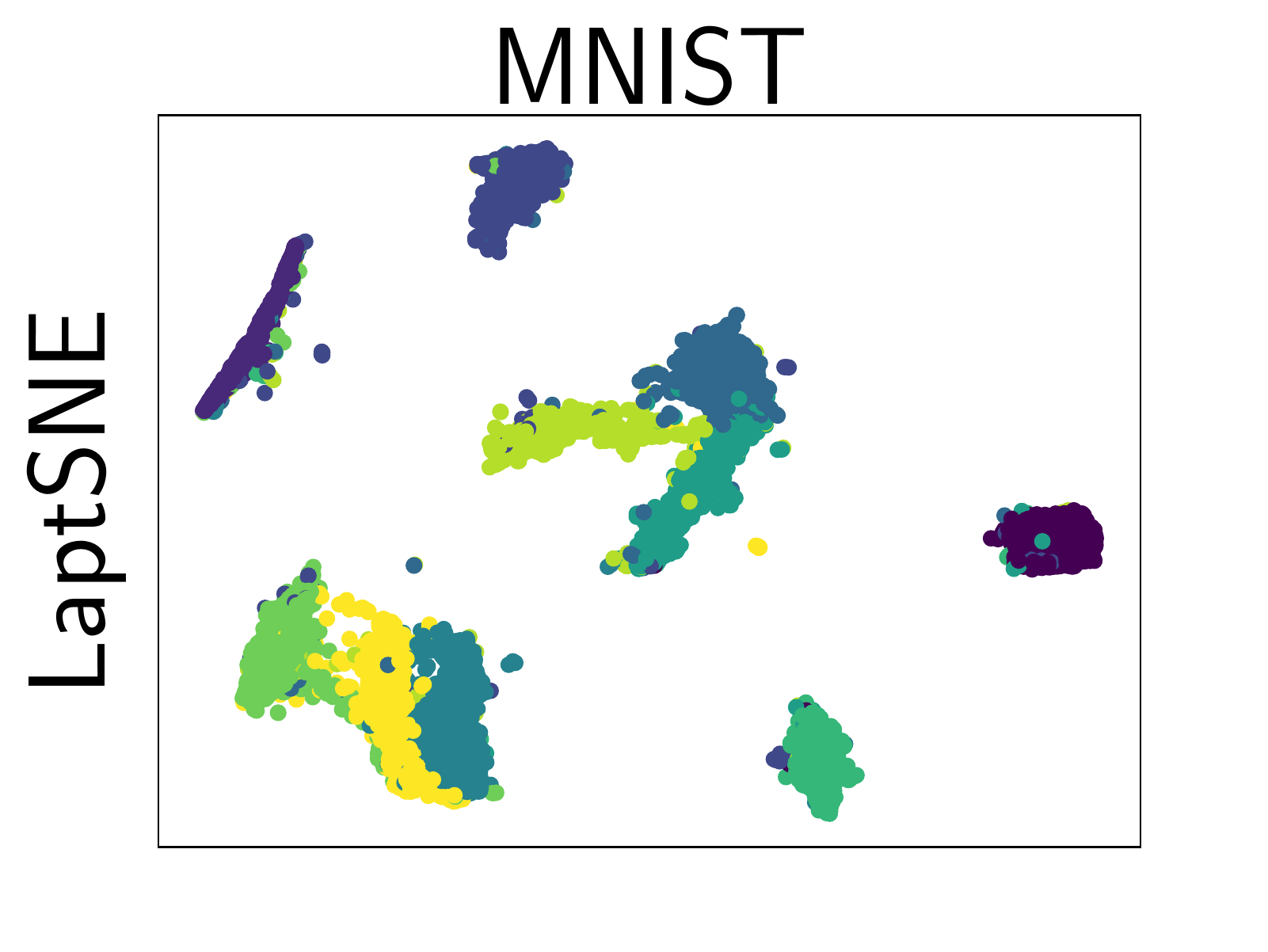}
    \end{subfigure}%
    ~ 
    \begin{subfigure}[t]{0.2\textwidth}
        \centering
        \includegraphics[height=0.9in]{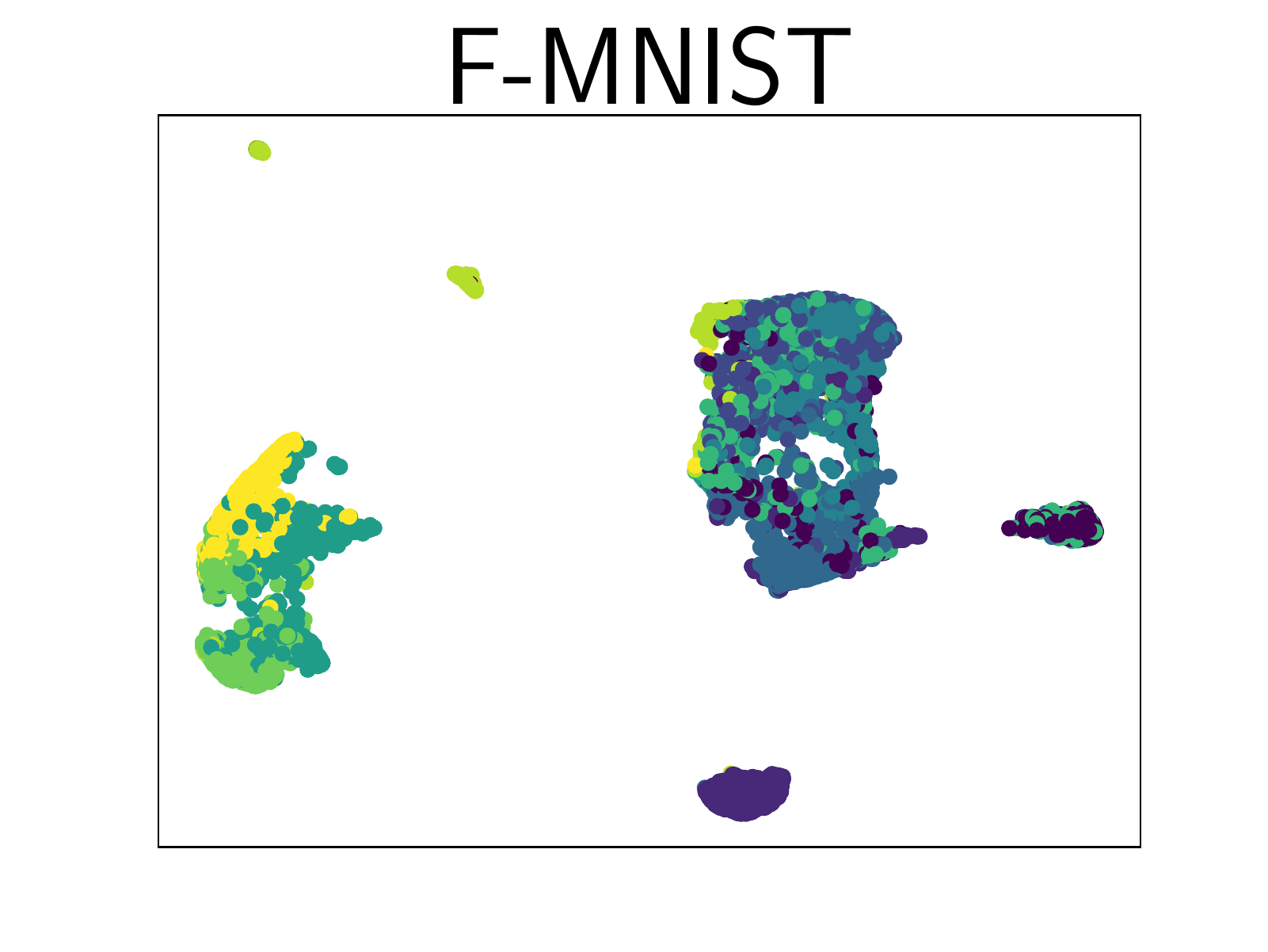}
    \end{subfigure}
    ~ 
    \begin{subfigure}[t]{0.2\textwidth}
        \centering
        \includegraphics[height=0.9in]{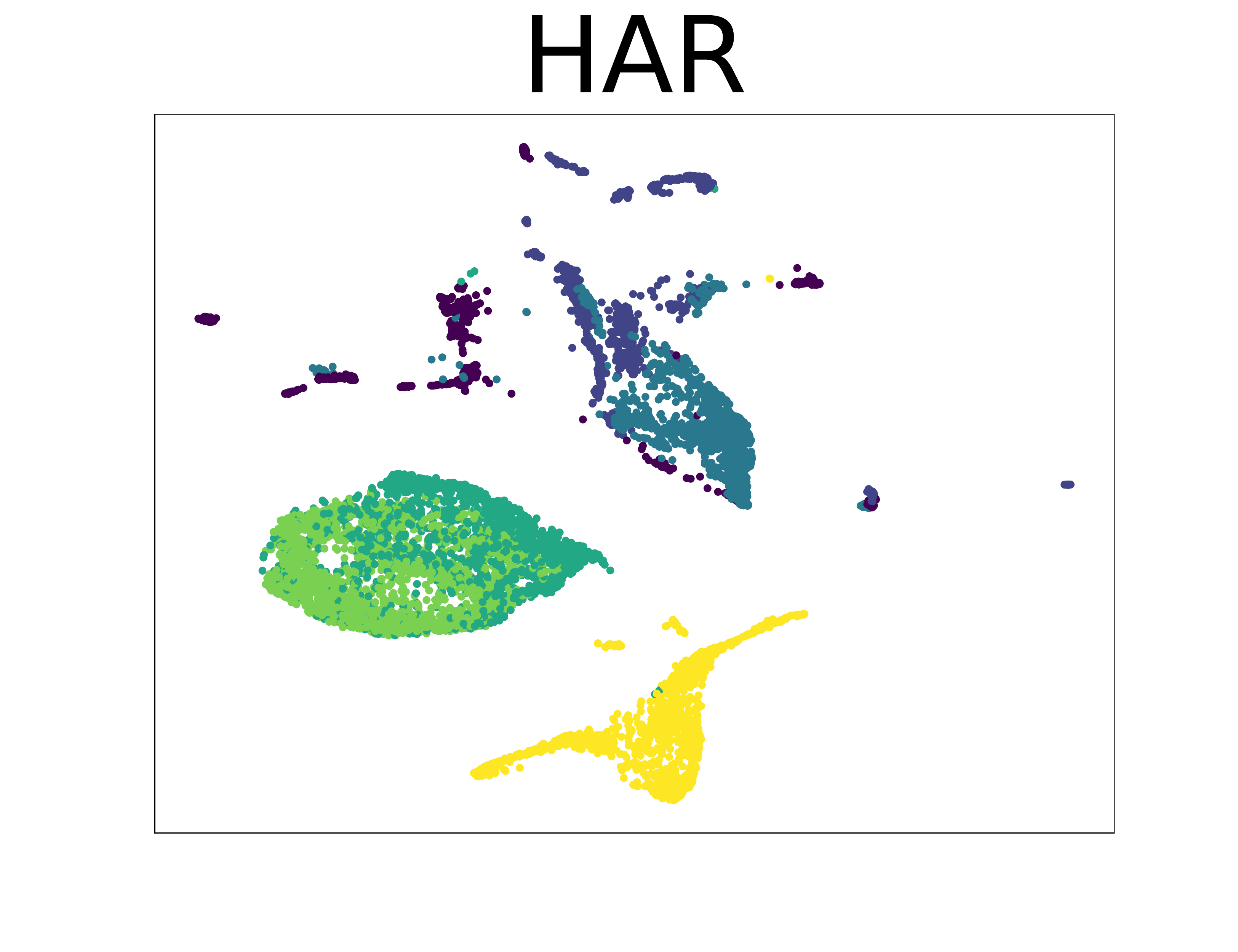}
    \end{subfigure}
    \\
       \begin{subfigure}[t]{0.2\textwidth}
        \centering
        \includegraphics[height=0.9in]{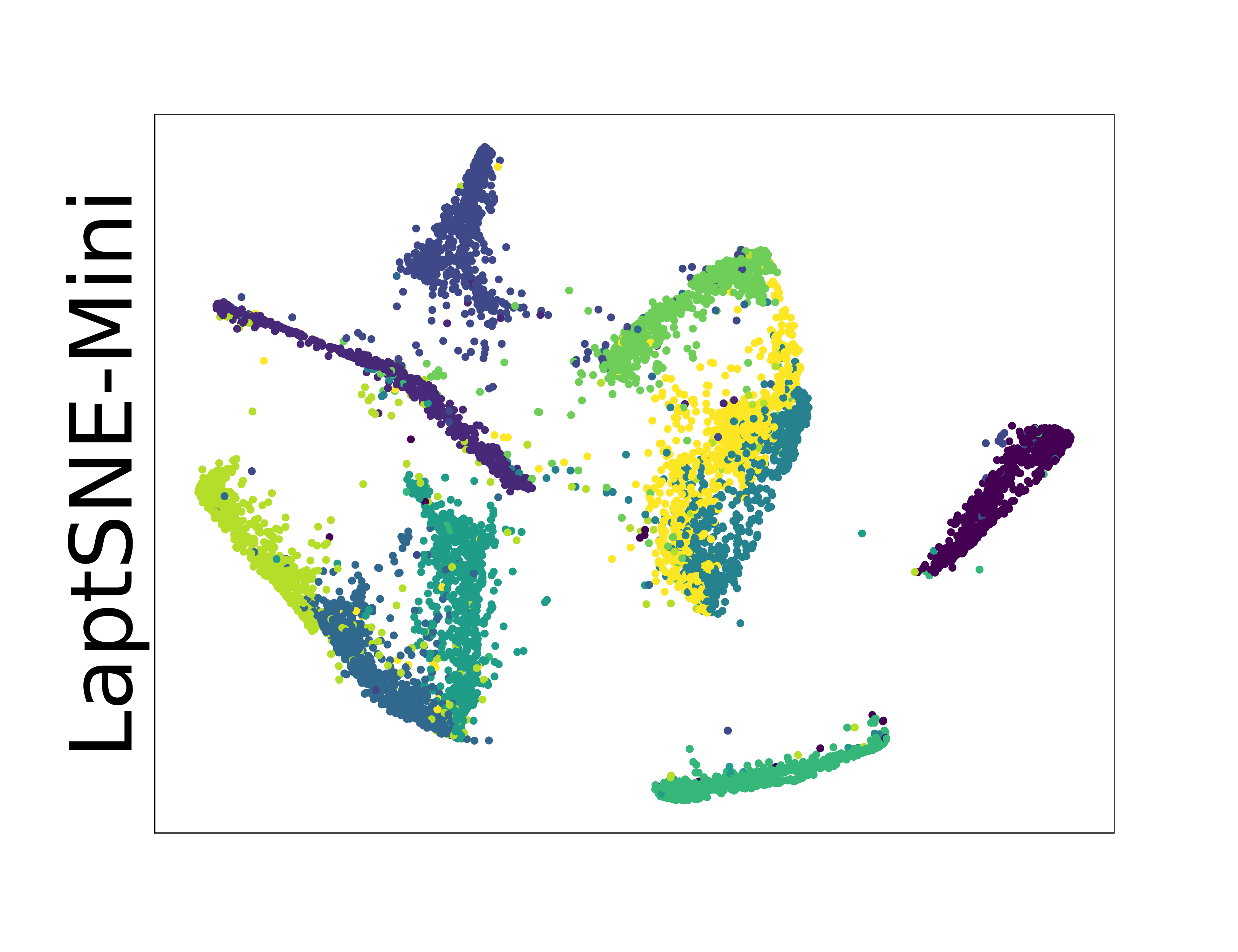}
    \end{subfigure}%
    ~ 
    \begin{subfigure}[t]{0.2\textwidth}
        \centering
        \includegraphics[height=0.9in]{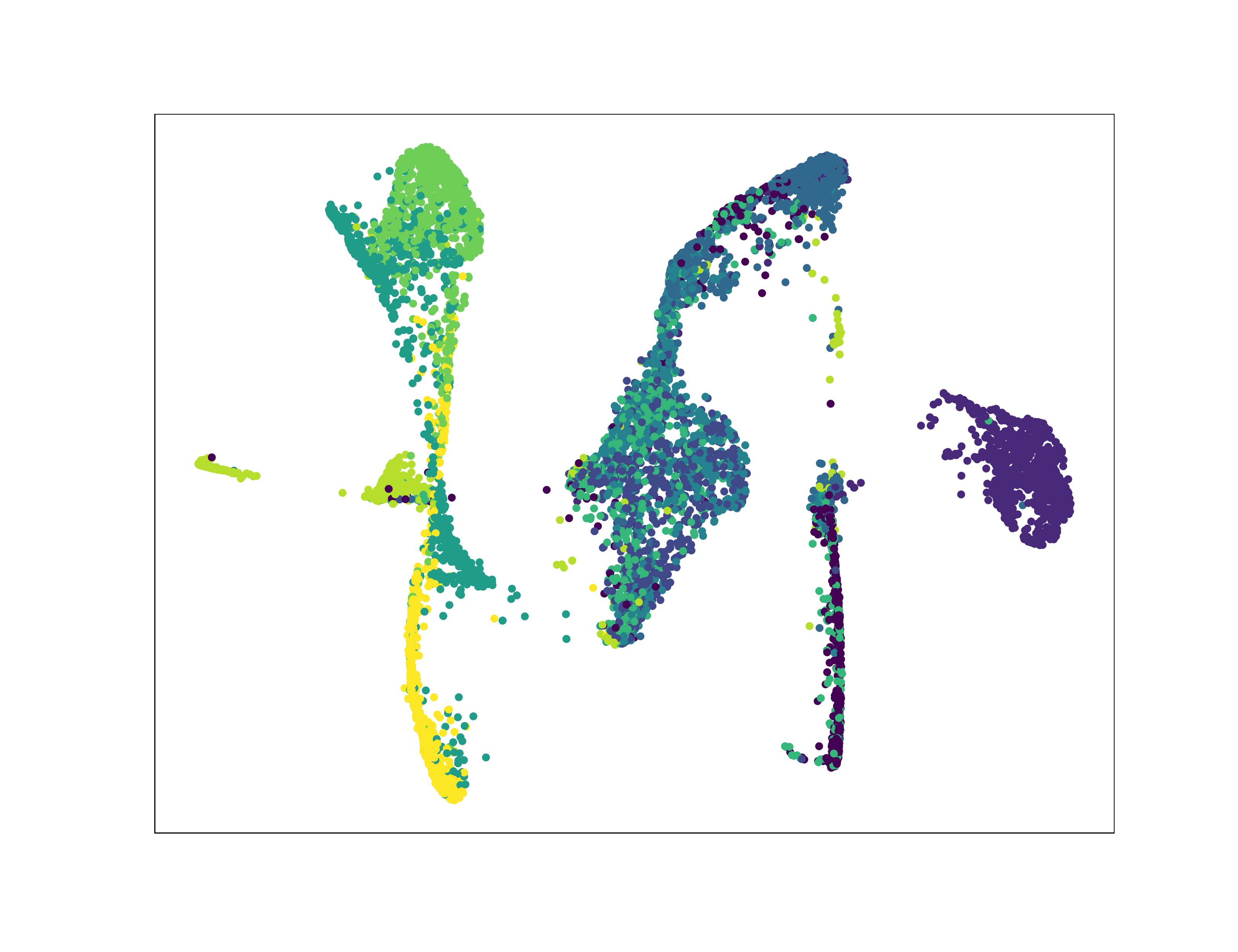}
    \end{subfigure}
    ~ 
    \begin{subfigure}[t]{0.2\textwidth}
        \centering
        \includegraphics[height=0.9in]{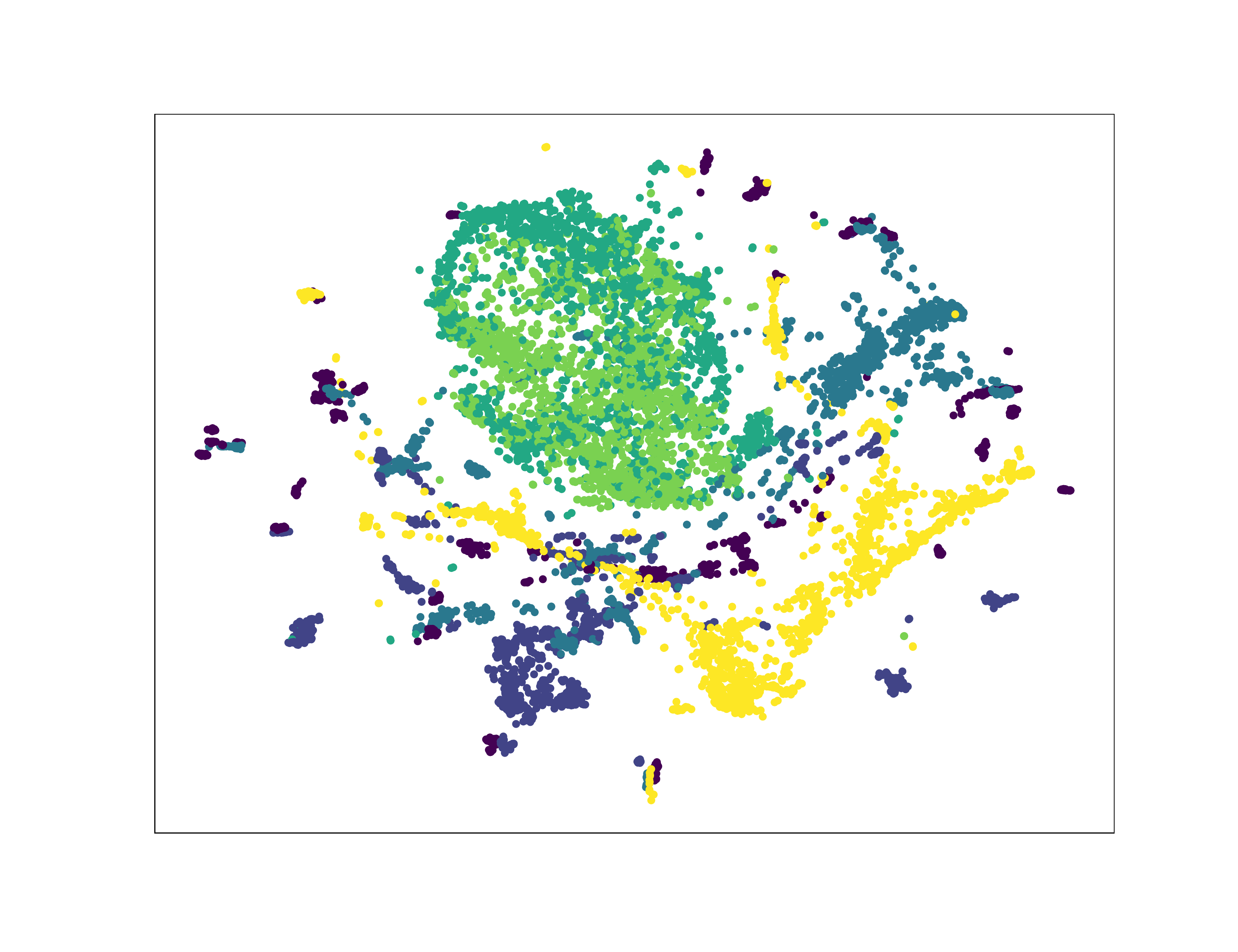}
    \end{subfigure}
    \\
        \begin{subfigure}[t]{0.2\textwidth}
        \centering
        \includegraphics[height=0.9in]{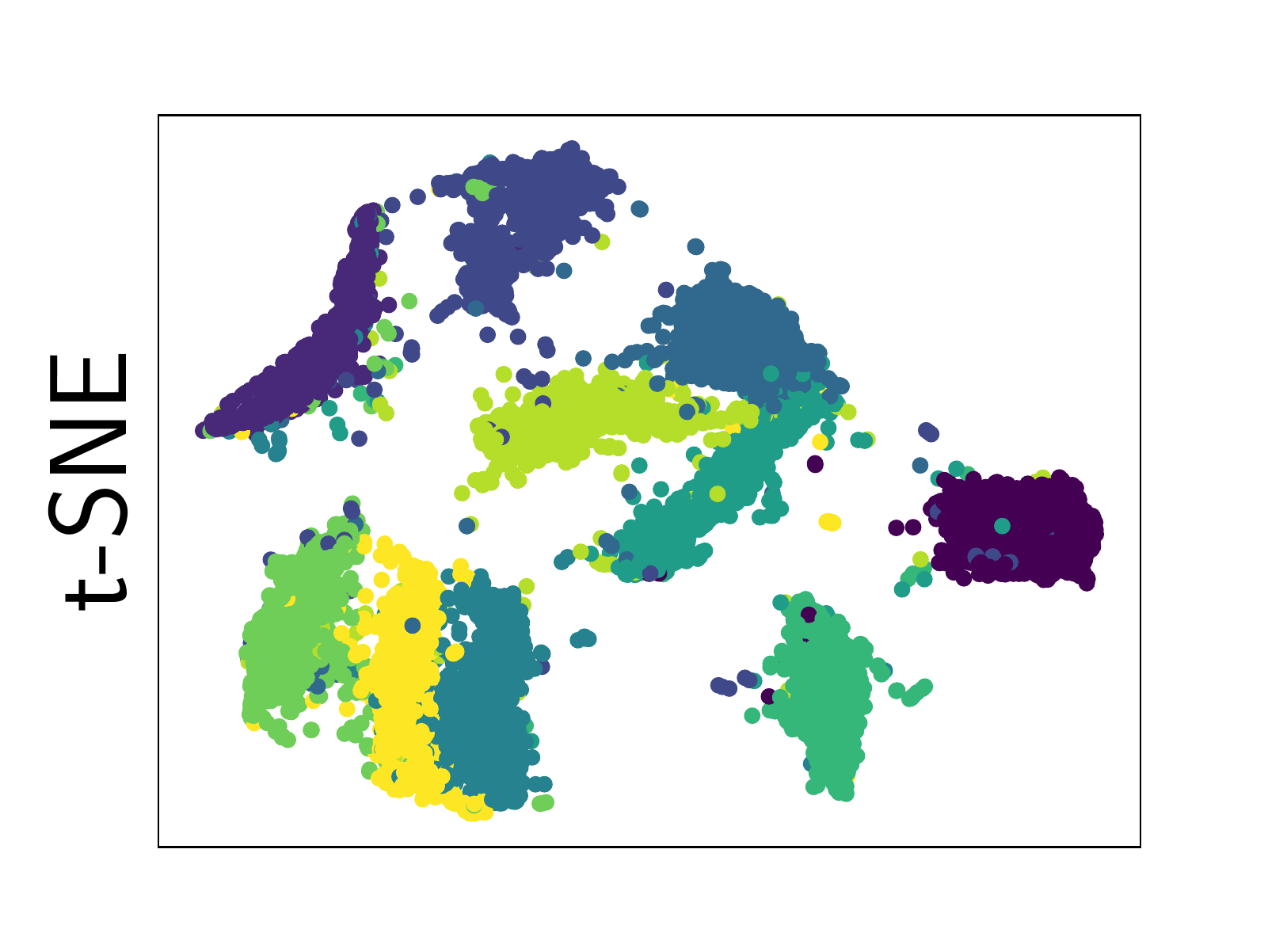}
    \end{subfigure}%
    ~ 
    \begin{subfigure}[t]{0.2\textwidth}
        \centering
        \includegraphics[height=0.9in]{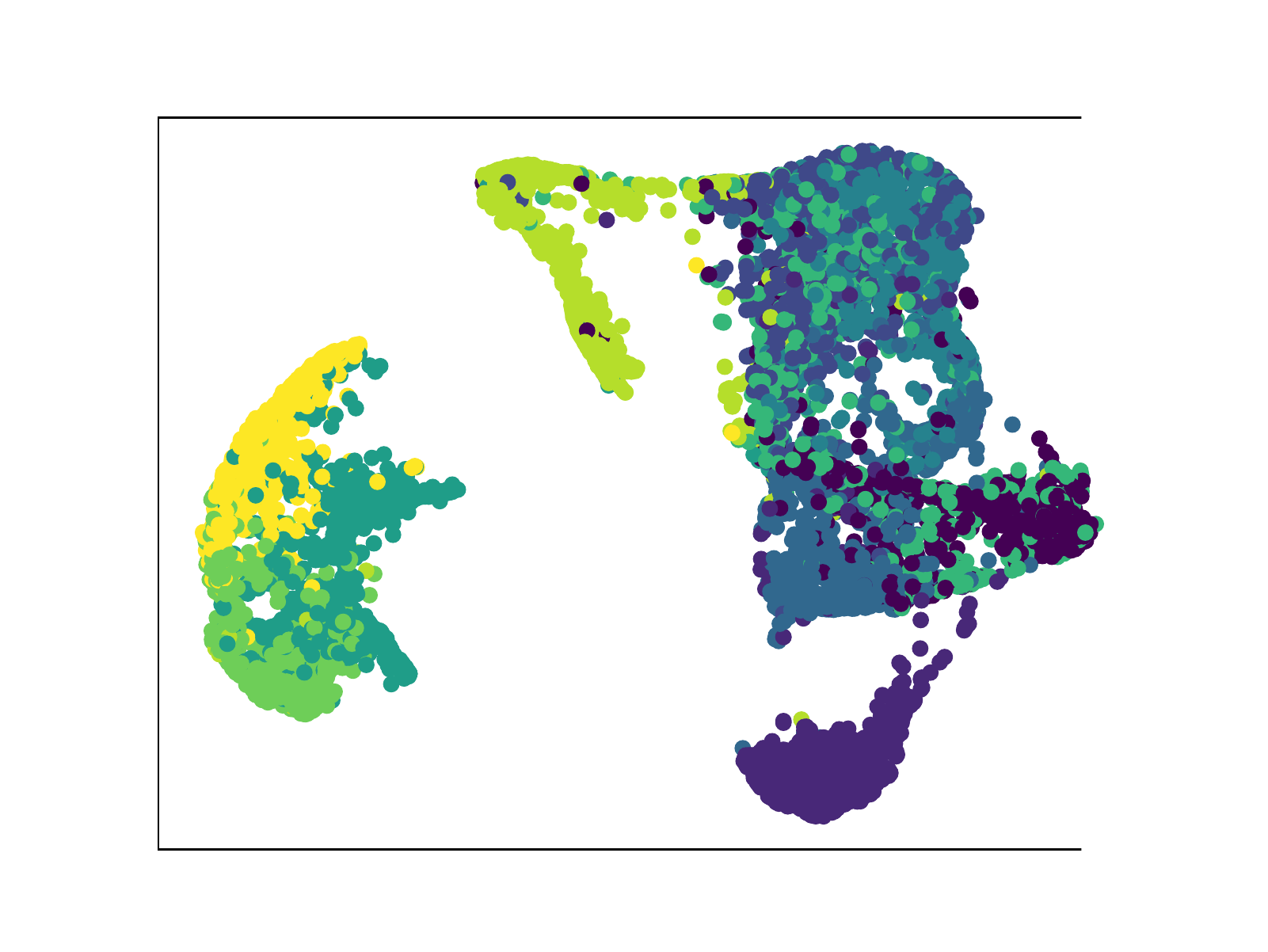}
    \end{subfigure}
    ~ 
    \begin{subfigure}[t]{0.2\textwidth}
        \centering
        \includegraphics[height=0.9in]{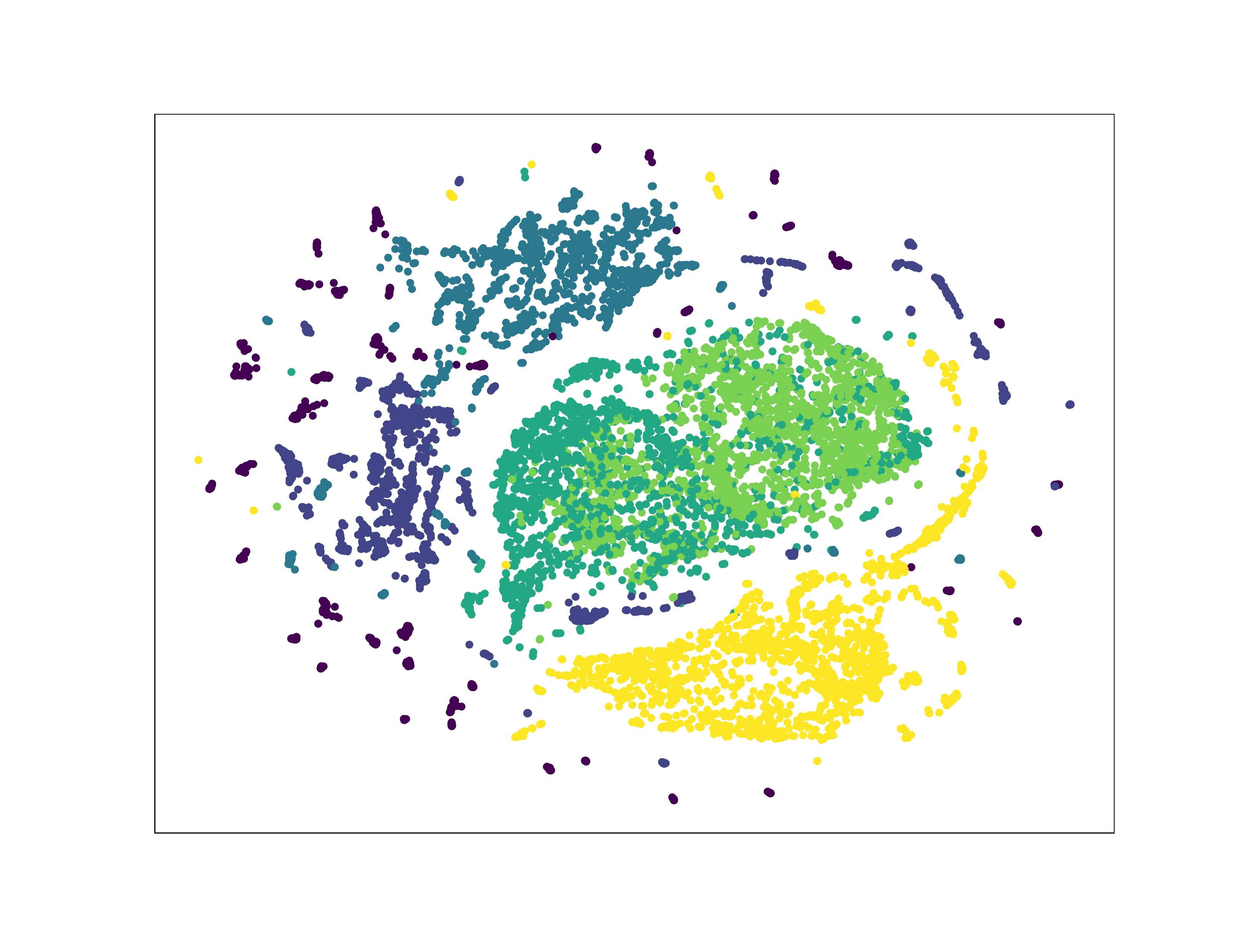}
    \end{subfigure}
  \caption{Row 1: Smallest few eigenvalues of graph Laplacian of each dataset (From left to right: MNIST, 14-th gap is large; Fashion-MNIST, 12-th gap is large; HAR, 10-th gap is large). Rows 2-4: Qualitative comparison of \textsc{Lap}tSNE, \textsc{Lap}tSNE-Mini and t-SNE in visualizing MNIST, Fashion-MNIST and HAR.}
  \label{fig:large dataset}
\end{figure}

In the same way mentioned above, the largest eigengap of HAR, MNIST and FMNIST are estimated, which turns out to be 10, 14 and 12 respectively, shown by the first row of Figure \ref{fig:large dataset}.  Since these are quite large datasets, we use the out-of-sample extension of \textsc{Lap}tSNE proposed in Section \ref{sec_ose}. The neural network consists of 4 hidden layers of size $W = \left \{  \left \lfloor\frac{2}{3}  \right \rfloor D, \left \lfloor\frac{1}{2}  \right \rfloor D, \left \lfloor\frac{1}{3}  \right \rfloor D, \left \lfloor\frac{1}{5}  \right \rfloor D\right \}$ where $D$ is the dimension of $\mathbf{X}$. The activation functions in the hidden layers are ReLU. The number of landmark points is 10000.

As shown in Figure \ref{fig:large dataset},  \textsc{Lap}tSNE is more effective in capturing the latent cluster information than t-SNE. For instance, in the result of \textsc{Lap}tSNE, the global relationships among different clusters of the digits in MNIST are more clearly identified and the clusters themselves are more compact.

\begin{table}[t!]
\footnotesize
\centering
\caption{Comparison between \textsc{Lap}tSNE, \textsc{Lap}tSNE-Mini, t-SNE, UMAP, Eigenmaps and PCA on PenDigits, MNIST, Fashion-MNIST and HAR dataset: 1) $k$-NN classifier accuracy for different values of $k$ \cite{mcinnes2018umap}; 2) clustering performances in terms of NMI, SC and DBI based on the hyperparameter $\hat{k}$. \textit{Note: $\uparrow$ means the higher is better, whereas $\downarrow$ indicates the lower is better}. The best value in each case is highlighted in bold.}
\label{table:large dataset}
\begin{tabular}{m{1em}c|cccccc}
\toprule
   {\begin{sideways}\textbf{~~}\end{sideways}} & \textbf{score} & \textbf{\textsc{Lap}tSNE}  & \textbf{\textsc{Lap}tSNE-Mini}  & \textbf{t-SNE}  & \textbf{UMAP} & \textbf{Eigenmaps} &\textbf{PCA}
   \\ 
\cline{2-8}
\rotatebox{90}{MNIST}& 
\makecell{100-nn $\uparrow$\\ 200-nn $\uparrow$\\ 400-nn $\uparrow$\\ 800-nn $\uparrow$\\NMI $\uparrow$\\SC $\uparrow$\\ DBI $\downarrow$} &

\makecell{ \textbf{0.943} \\ 0.940 \\ \textbf{0.937}\\ \textbf{0.935} \\ \textbf{0.7498} \\ \textbf{0.5886} \\ \textbf{0.5592}}   & 

\makecell{0.940\\0.939\\0.934\\0.933\\ 0.7340 \\ 0.5809 \\ 0.6038}   & 

\makecell{0.941\\0.937\\0.933\\0.932\\ 0.6221 \\ 0.4178 \\ 0.7268}   & 

\makecell{\textbf{0.943} \\ \textbf{0.941} \\ 0.935 \\ 0.930 \\ 0.6892 \\0.4949 \\ 0.7476}   & 
\makecell{0.695 \\ 0.680 \\ 0.659 \\ 0.635\\ 0.3101 \\ 0.3464 \\ 0.8447}  &
\makecell{0.548 \\  0.524   \\ 0.503 \\ 0.497  \\ 0.3595 \\  0.3519 \\ 0.8487} 
\\
\cline{2-8}

\rotatebox{90}{{\small F-MNIST}}   & 
\makecell{100-nn $\uparrow$\\ 200-nn $\uparrow$\\ 400-nn $\uparrow$\\ 800-nn $\uparrow$\\NMI $\uparrow$\\SC $\uparrow$\\ DBI $\downarrow$} &
\makecell{\textbf{0.787} \\  0.768  \\ \textbf{0.762} \\ 0.753\\ 0.6009 \\\textbf{0.6299} \\ \textbf{0.4679}}   & 

\makecell{0.783\\0.760\\0.758\\0.753\\ 0.6004 \\0.6298 \\ 0.4680}   &

\makecell{0.782 \\ \textbf{0.771} \\\textbf{0.762}\\ \textbf{0.754}\\ 0.5905 \\ 0.6113 \\ 0.5035}   & 

\makecell{0.754 \\ 0.738 \\ 0.718 \\ 0.697\\ \textbf{0.6051} \\ 0.5325 \\ 0.6557}   & 

\makecell{0.695 \\ 0.675 \\ 0.652 \\ 0.636 \\ 0.4595\\ 0.4789 \\ 0.7410} &

\makecell{0.606 \\  0.583   \\ 0.574 \\ 0.563  \\0.4311 \\  0.4100 \\ 0.7709}
\\
\cline{2-8}
\rotatebox{90}{{\small HAR}}   & 
\makecell{100-nn $\uparrow$\\ 200-nn $\uparrow$\\ 400-nn $\uparrow$\\ 800-nn $\uparrow$\\NMI $\uparrow$\\SC $\uparrow$\\ DBI $\downarrow$} &
\makecell{\textbf{0.931} \\  \textbf{0.924}  \\ \textbf{0.910} \\ \textbf{0.884}\\ \textbf{0.734} \\ 0.4667 \\ 0.7973}   & 

\makecell{0.918 \\  0.906  \\ 0.893 \\ 0.870\\ 0.6015 \\ 0.4059 \\ 0.8046}   & 

\makecell{0.915 \\  0.909  \\ 0.897 \\ 0.840\\ 0.4601\\ 0.3893 \\ 0.8446}   & 

\makecell{0.893 \\  0.881  \\ 0.860 \\ 0.852\\ 0.6878 \\  0.5275  \\ 0.7376}   & 

\makecell{0.769 \\  0.746  \\ 0.739 \\ 0.722\\ 0.6289 \\ \textbf{0.7657} \\ \textbf{0.4756}} &

\makecell{0.650 \\  0.639  \\ 0.632 \\ 0.619 \\ 0.4590 \\ 0.4455 \\ 0.7371} 
\\
\bottomrule
\end{tabular}
\end{table}

In Table \ref{table:large dataset} we present the NMI, DC, DBI and k-NN scores on MNIST, Fashion-MNIST and HAR. \textsc{Lap}tSNE established comparable performance with t-SNE on Fashion-MNIST at both local and non-local scales, but notably better than UMAP. For MNIST, \textsc{Lap}tSNE is slightly better than other algorithms when k varies from 100 to 400, but has significantly higher accuracy for k values of 800. 

As evidenced by this comparison, \textsc{Lap}tSNE provides largely comparabe performance across large datasets both qualitatively and quantitatively. 
The success of \textsc{Lap}tSNE may be owing to the theory proved in \cite{cai2021theoretical} that the low-dimensional map in t-SNE algorithm converges cluster-wise towards some limit points in $\mathbb{R}^2$, only depending on the initialization, and each associated with a connected component of the underlying graph. According to our experiments, the Laplacian regularization term helps strengthen this process without undermining the fine properties of t-SNE.

\subsection{Sensitivity Analysis}

\begin{figure}[h]
\centering
\begin{minipage}{.45\textwidth}
  \centering
  \includegraphics[width=0.9\linewidth]{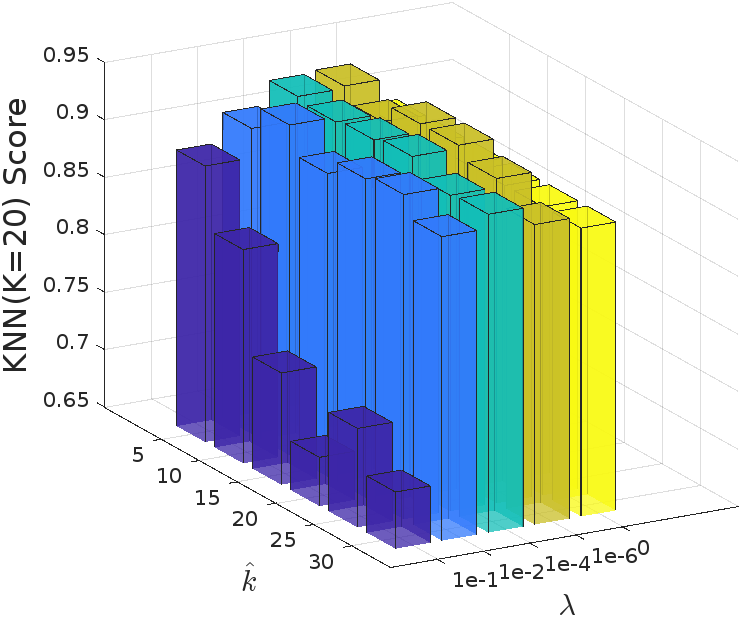}
  \captionof{figure}{K-NN scores (k=20) on the COIL20 dataset: \textsc{Lap}tSNE result with different $\lambda$ (ranging from 0 to 1e-1) and estimated number of potential cluster $\hat{k}$ (ranging from 5 to 30).}
  \label{fig:sensitivity analysis COIL20}
\end{minipage}%
\hspace{0.10cm}
\begin{minipage}{.45\textwidth}
  \centering
\includegraphics[width=0.5\linewidth]{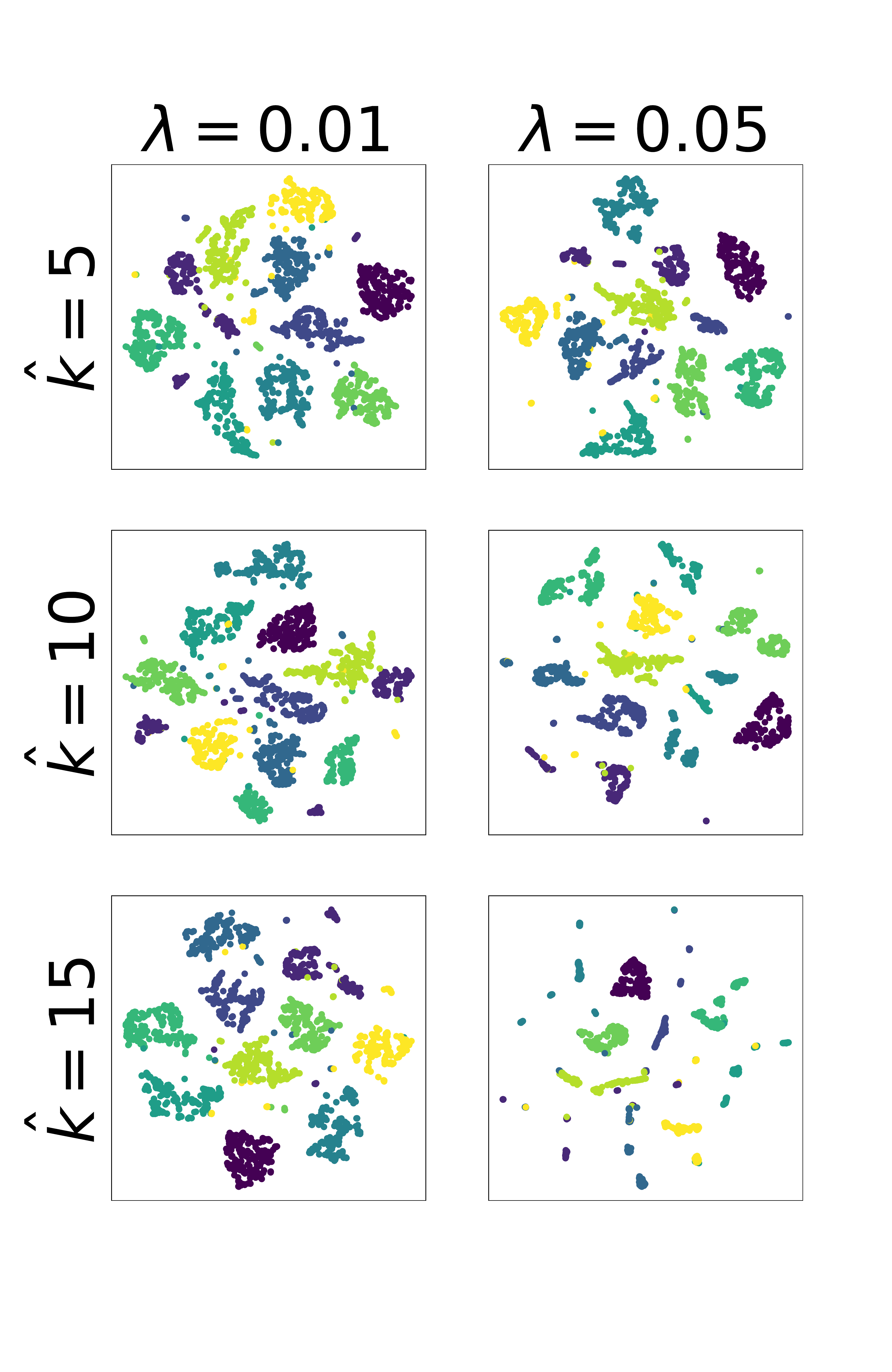}
  \captionof{figure}{Variation of \textsc{Lap}tSNE embedding  with different hyperparameters $\hat{k}$ and $\lambda$ on the PenDigits dataset.}
  \label{fig: pendigits}
\end{minipage}
\end{figure}

We analyze the sensitivity of \textsc{Lap}tSNE to the hyperparameters $\lambda$ and $\hat{k}$ qualitatively and quantitatively. For example, Figure \ref{fig:sensitivity analysis COIL20} presents the k-NN score (k=20) of \textsc{Lap}tSNE with different $\hat{k}$ and $\lambda$ on COIL20. 
We notice that large scale of $\lambda$ (e.g. 1e-1) would crush the performance but smaller ones from 1e-2 to 1e-6 could yield results even better than that of the vanilla t-SNE ($\lambda=0$). The hyperparameter $\hat{k}$ may fluctuate the performance in a non-monotonic manner. For the order of $\hat{k}$ where a large eigengap exists, the k-NN scores are the best. This indicates the importance of tuning the estimated number of potential cluster $\hat{k}$ according to the eigenvalues of graph Laplacian in the original space.
When tuning $\hat{k}=19$ and $\lambda=$1e-4, \textsc{Lap}tSNE has the highest k-NN score (k=20) as Table \ref{table:small dataset} shows,  and the embedding layout is cluster-contractive, which consists of many circles and lines in Figure \ref{fig:small dataset}.

In Figure \ref{fig: pendigits}, it is shown obviously that the clusters shrink with the increase of $\hat{k}$. Besides, the experiment further supports the interaction effect between the estimated number of potential clusters $\hat{k}$ and the tuning parameter $\lambda$: relatively larger $\lambda$ and $\hat{k}$ can produce more compact clusters.

\subsection{Trajectory of \textsc{Lap}tSNE}

\begin{figure}[t]
\centering
\begin{minipage}{.45\textwidth}
  \centering
   {\renewcommand{\arraystretch}{2}
   \begin{tabular}{c@{\hspace{0.1mm}}M{15mm}M{15mm}}
     &\includegraphics[width=0.3\textwidth]{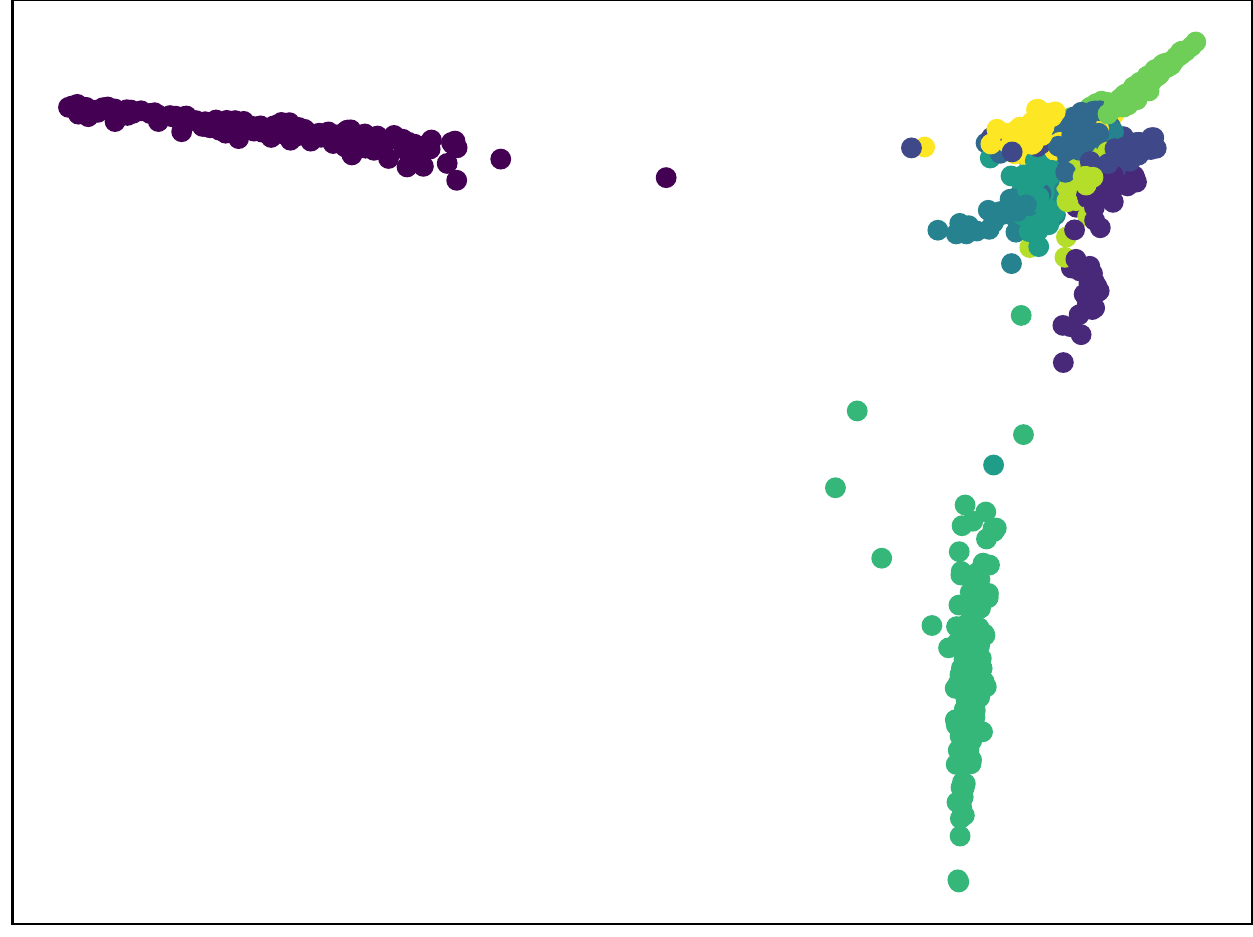} &
     \includegraphics[width=0.3\textwidth]{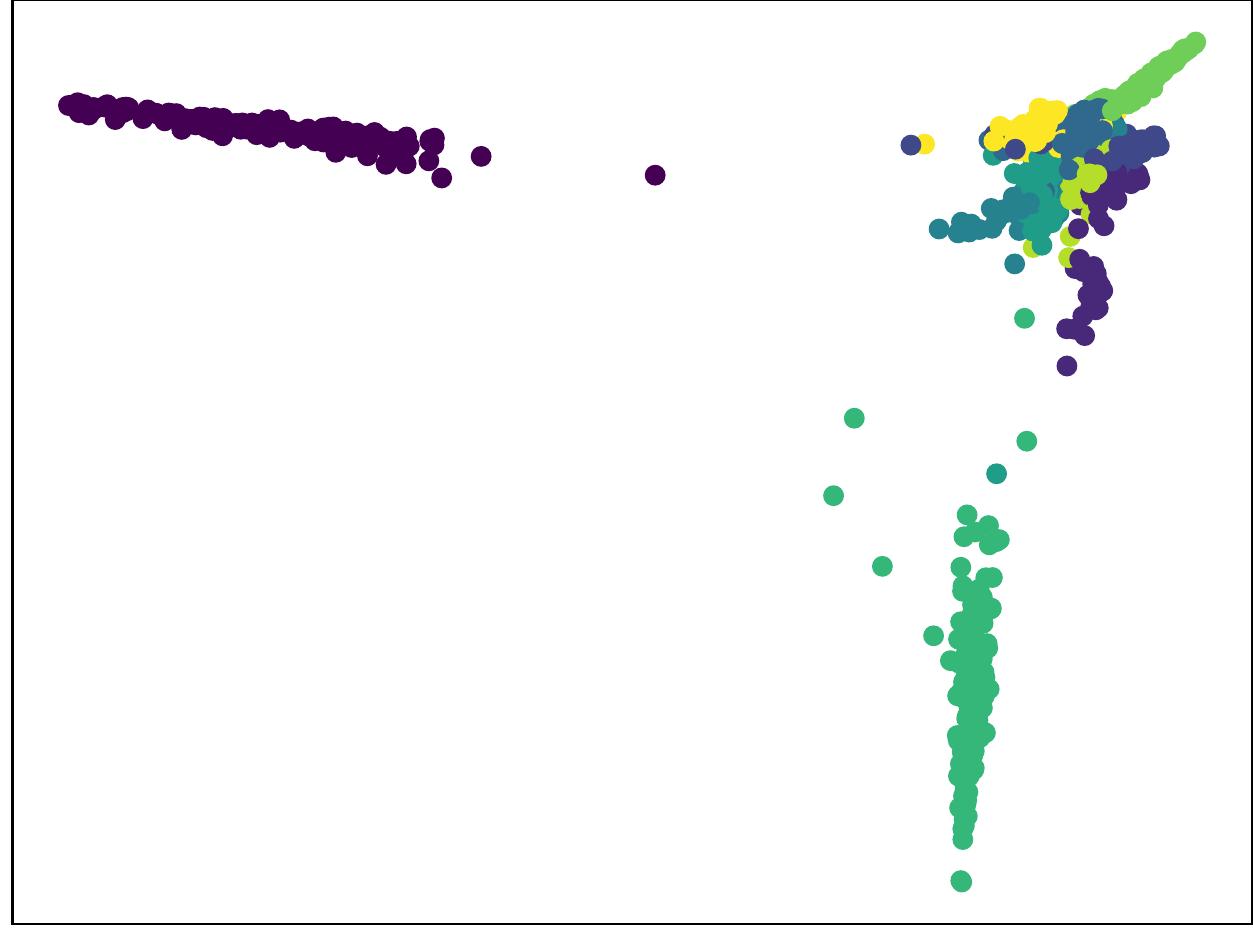} \\
     &\includegraphics[width=0.3\textwidth]{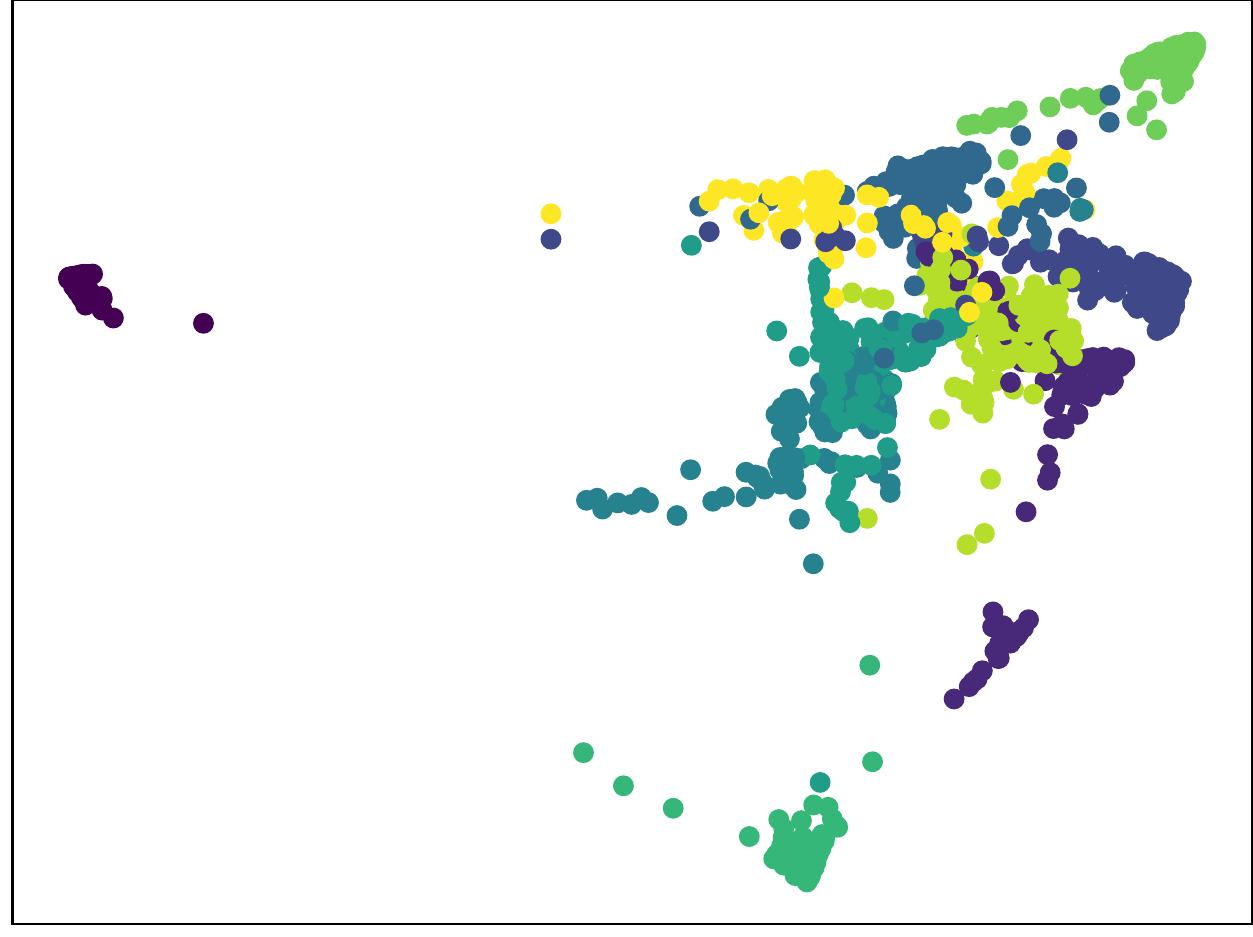} &
     \includegraphics[width=0.3\textwidth]{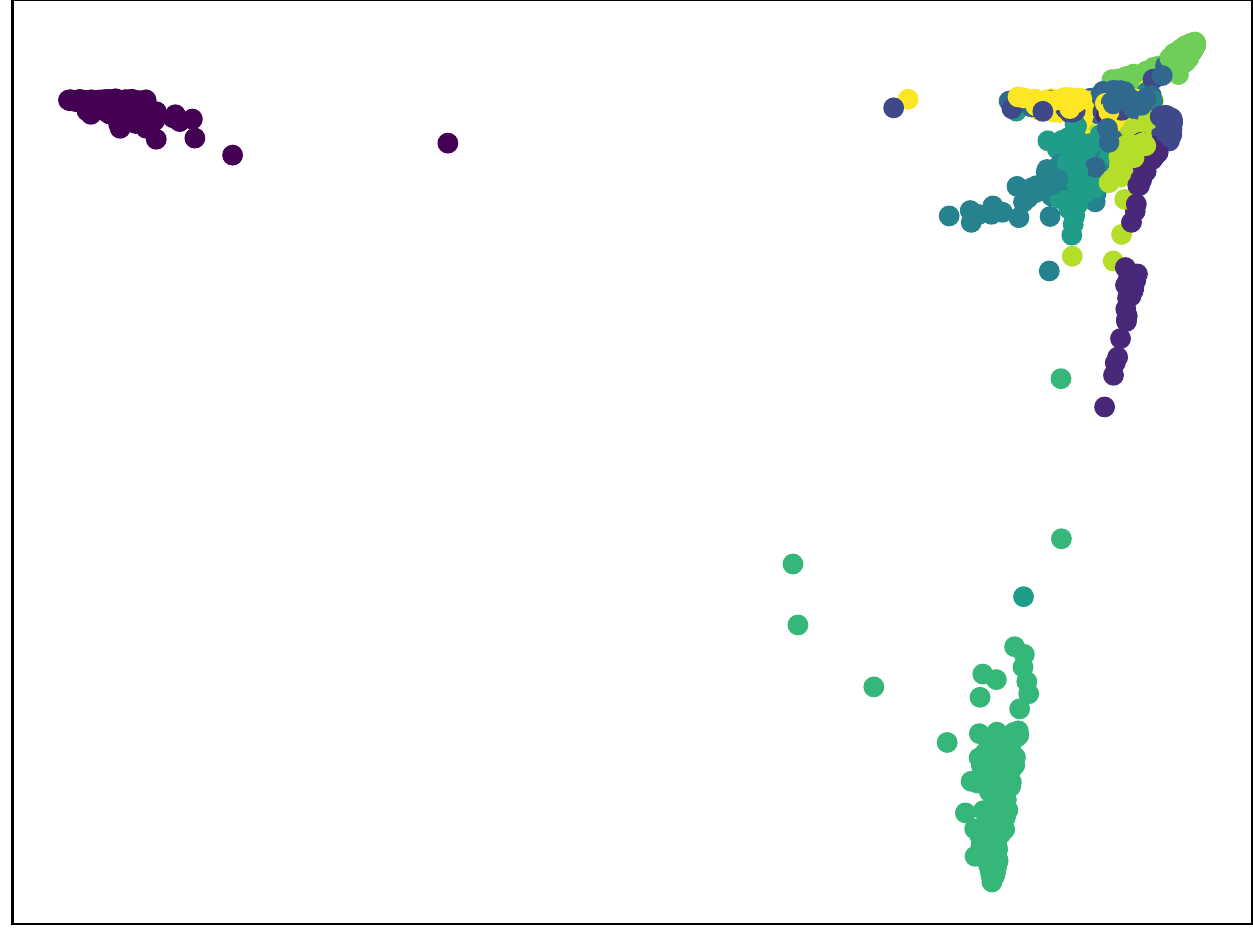}  \\
     &\includegraphics[width=0.3\textwidth]{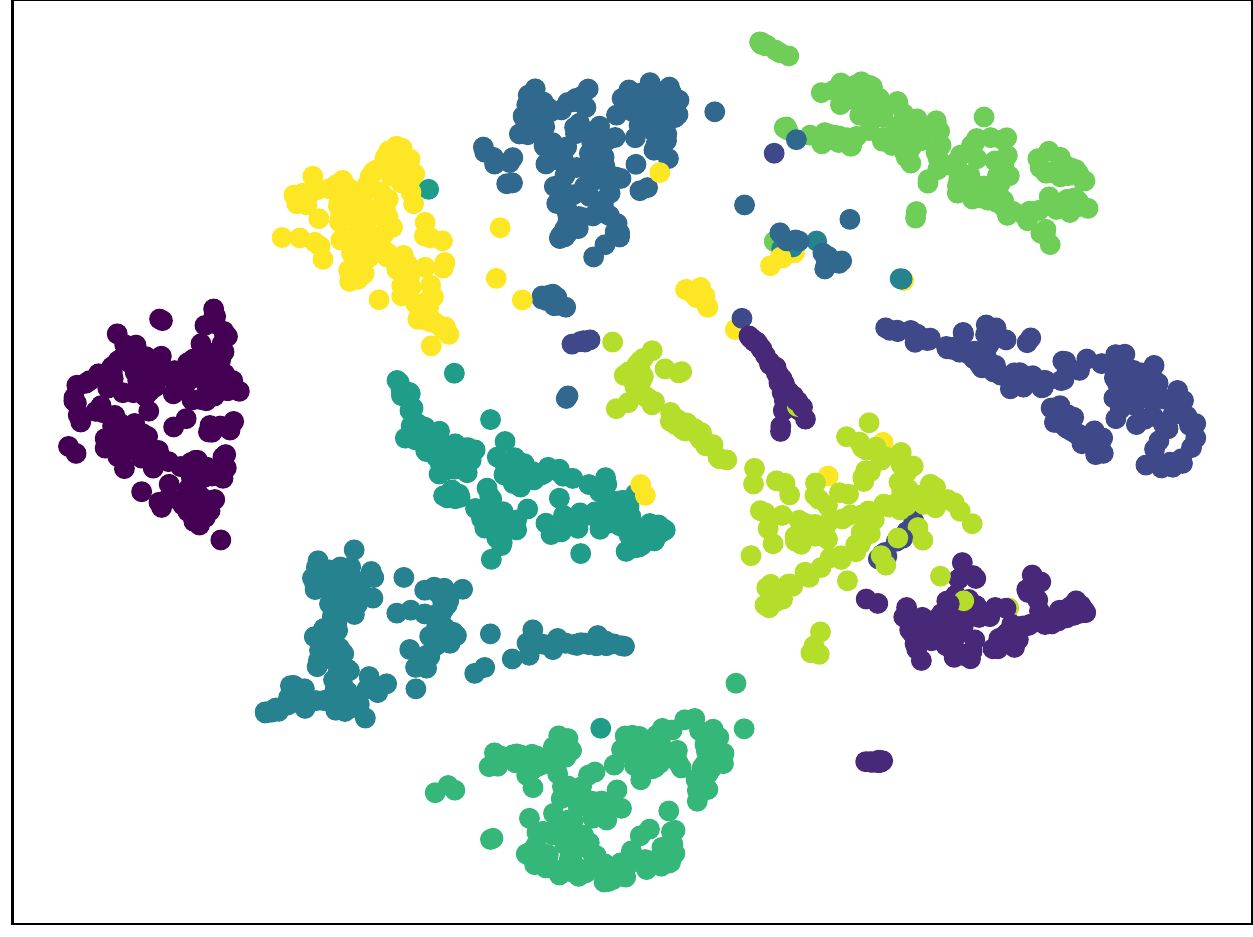} &
     \includegraphics[width=0.3\textwidth]{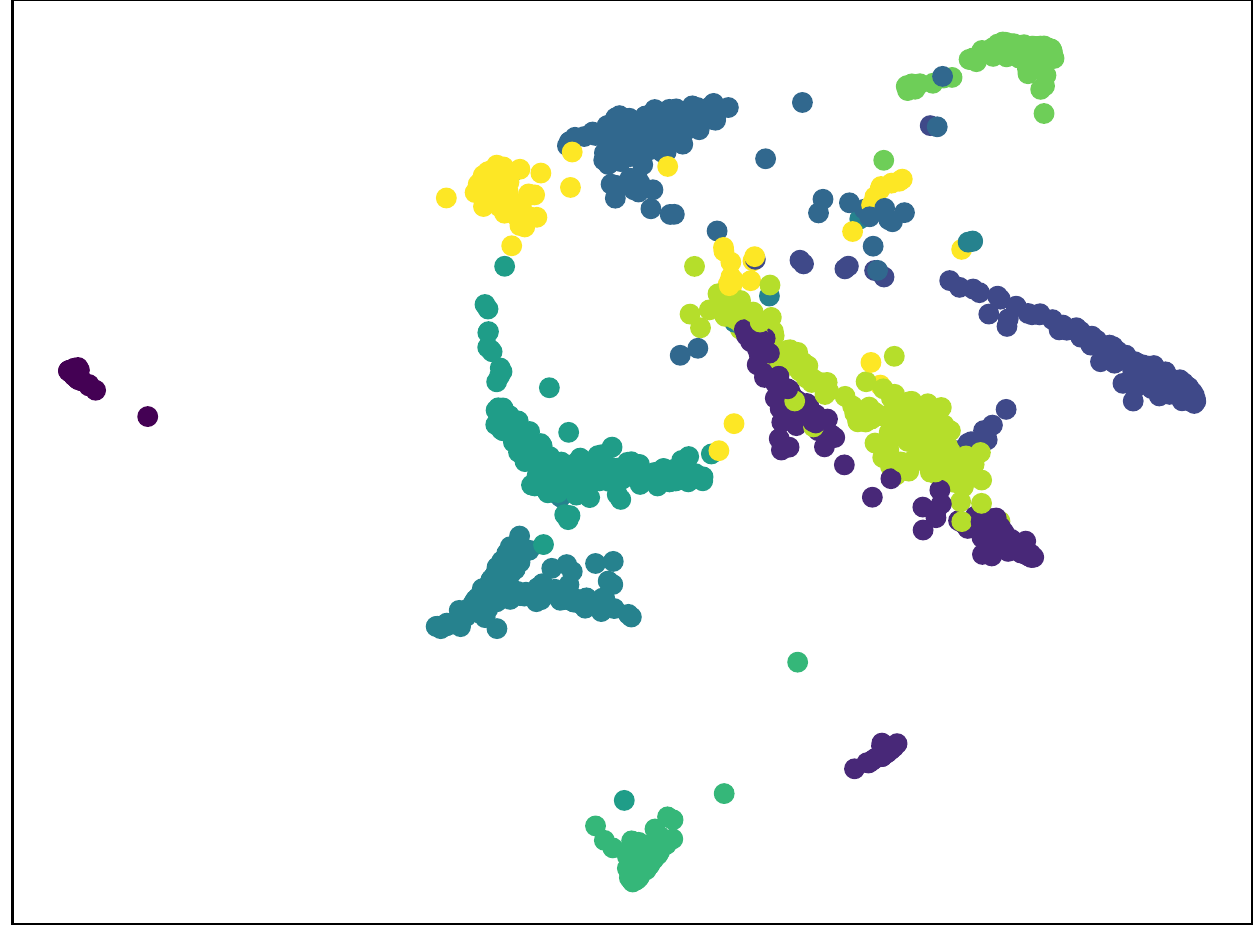}  \\
     &\includegraphics[width=0.3\textwidth]{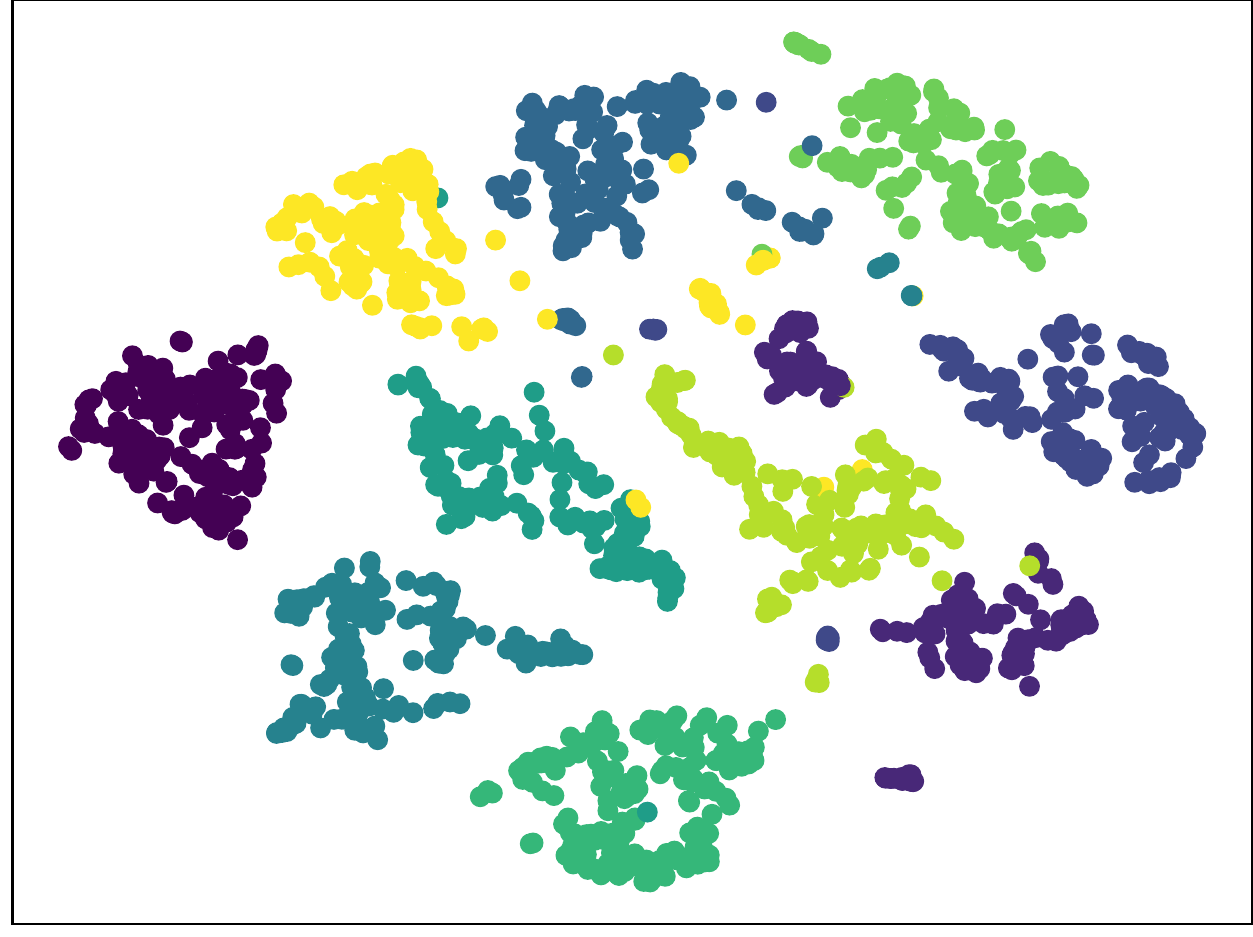} &
     \includegraphics[width=0.3\textwidth]{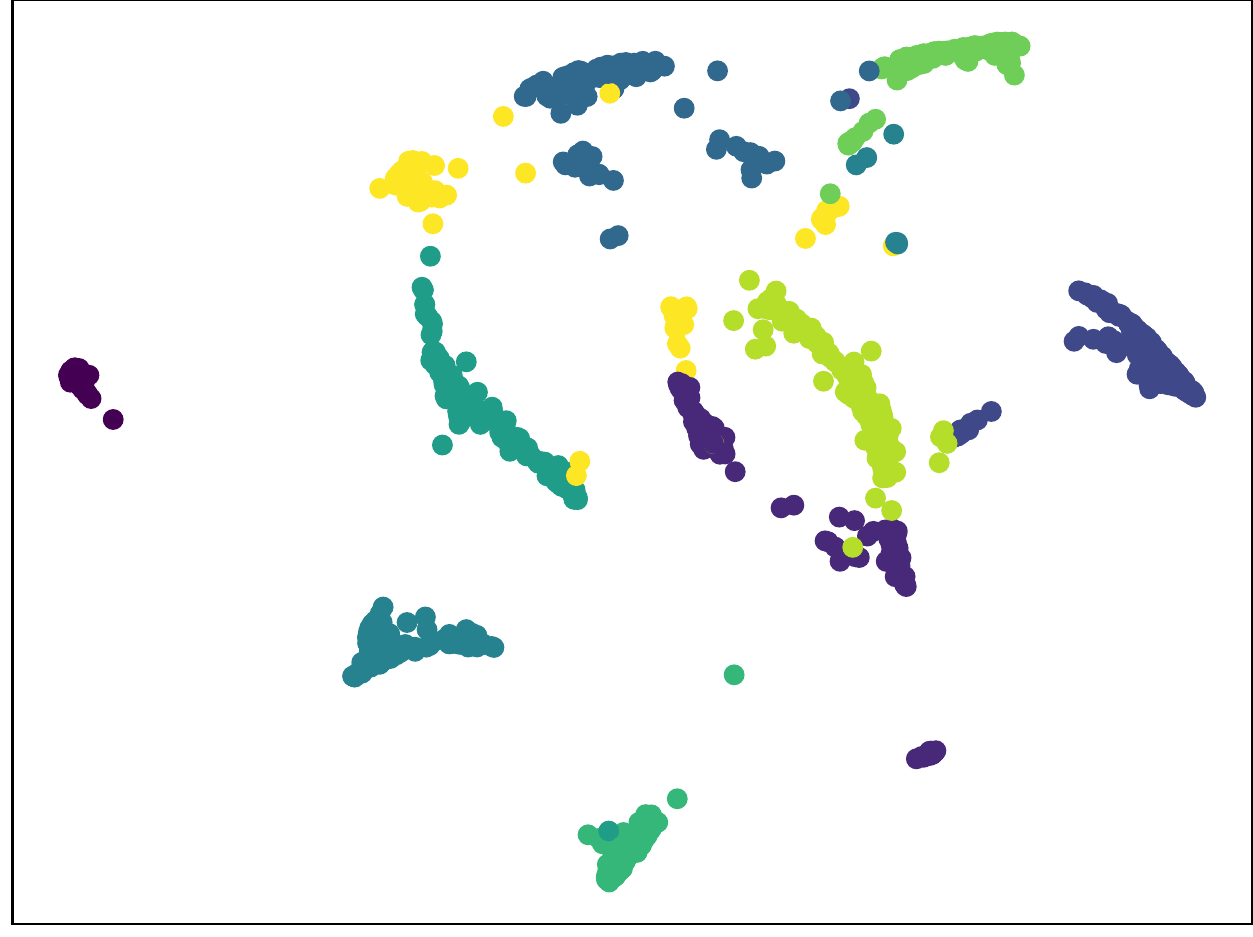}  \\
   \end{tabular}}
   \caption{A comparison of t-SNE and \textsc{Lap}tSNE in the first 50 iterations on PenDigits. (Left: t-SNE, Right: \textsc{Lap}tSNE; Row 1 is the initial stage, Row 2 to Row 4 are iteration 10, 30, 50)}
\label{Itecomp}
\end{minipage}%
\hspace{0.5cm}
\begin{minipage}{.45\textwidth}
    \centering
     \includegraphics[width=1\textwidth]{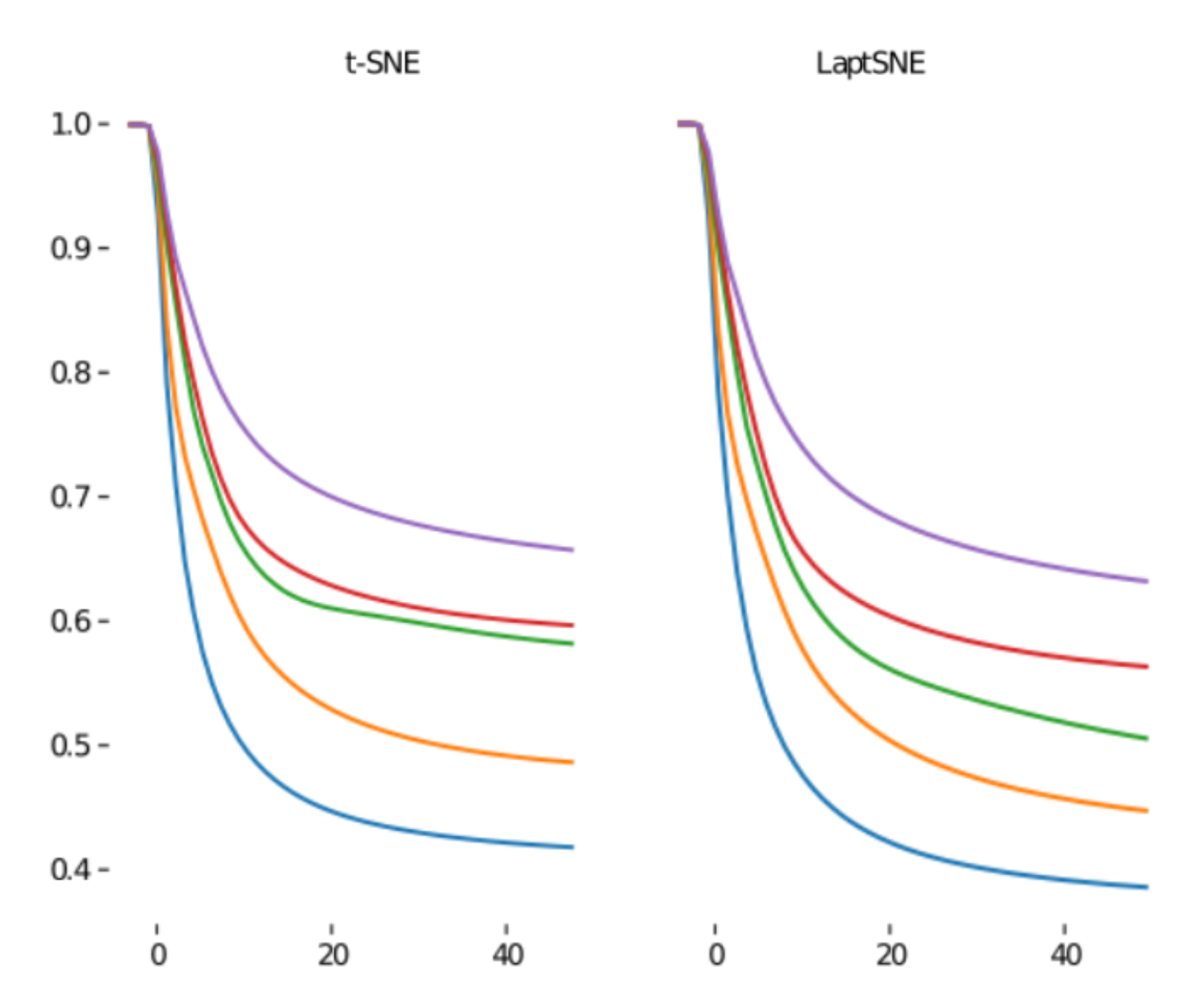}
    \caption{Part of the eigenvalues of T-Student kernel graph in the 2-dimensional embeddings of \textsc{Lap}tSNE and t-SNE (in different iterations) on PenDigits. The green curve stands for the 11-th eigenvalue.}
    \label{fig:iterationgap}
\end{minipage}
\end{figure}

We discover that \textsc{Lap}tSNE outperforms t-SNE in the early stage of iterations. As Figure \ref{Itecomp} illustrates, the lower-dimensional embeddings of \textsc{Lap}tSNE start from the same intialization as t-SNE but expand in a cluster-contractive manner, which leads to a well-clustered layout in the end.
 
Note that, shown in Figure \ref{Itecomp}, the clusters in the embeddings of \textsc{Lap}tSNE are more distinguishable than those in the embeddings of t-SNE. This is consistent with the fact that, shown in Figure \ref{fig:iterationgap}, the gap between the 11-th and 12-th eigenvalues of the lower-dimensional representation's graph Laplacian is increasingly large during the iteration of \textsc{Lap}tSNE, compared with vanilla t-SNE.

\section{Conclusions}
This work provided a new NLDR method called \textsc{Lap}tSNE and its several extensions for high-dimensional data visualization.  The proposed methods generate cluster-informative low-dimensional embedding and outperform t-SNE and UMAP visually and quantitatively on seven benchmark datasets. It should be pointed out that the proposed method can be adapted to other NLDR methods such as UMAP to boost the visualization performance.  


\begin{thebibliography}{44}


\ifx \showCODEN    \undefined \def \showCODEN     #1{\unskip}     \fi
\ifx \showDOI      \undefined \def \showDOI       #1{#1}\fi
\ifx \showISBNx    \undefined \def \showISBNx     #1{\unskip}     \fi
\ifx \showISBNxiii \undefined \def \showISBNxiii  #1{\unskip}     \fi
\ifx \showISSN     \undefined \def \showISSN      #1{\unskip}     \fi
\ifx \showLCCN     \undefined \def \showLCCN      #1{\unskip}     \fi
\ifx \shownote     \undefined \def \shownote      #1{#1}          \fi
\ifx \showarticletitle \undefined \def \showarticletitle #1{#1}   \fi
\ifx \showURL      \undefined \def \showURL       {\relax}        \fi
\providecommand\bibfield[2]{#2}
\providecommand\bibinfo[2]{#2}
\providecommand\natexlab[1]{#1}
\providecommand\showeprint[2][]{arXiv:#2}

\bibitem[Arora et~al\mbox{.}(2018)]%
        {pmlr-v75-arora18a}
\bibfield{author}{\bibinfo{person}{Sanjeev Arora}, \bibinfo{person}{Wei Hu},
  {and} \bibinfo{person}{Pravesh~K. Kothari}.} \bibinfo{year}{2018}\natexlab{}.
\newblock \showarticletitle{An Analysis of the t-SNE Algorithm for Data
  Visualization}. In \bibinfo{booktitle}{\emph{Proceedings of the 31st
  Conference On Learning Theory}} \emph{(\bibinfo{series}{Proceedings of
  Machine Learning Research}, Vol.~\bibinfo{volume}{75})},
  \bibfield{editor}{\bibinfo{person}{Sébastien Bubeck},
  \bibinfo{person}{Vianney Perchet}, {and} \bibinfo{person}{Philippe Rigollet}}
  (Eds.). \bibinfo{publisher}{PMLR}, \bibinfo{pages}{1455--1462}.
\newblock
\urldef\tempurl%
\url{https://proceedings.mlr.press/v75/arora18a.html}
\showURL{%
\tempurl}


\bibitem[Baker(1977)]%
        {baker1977numerical}
\bibfield{author}{\bibinfo{person}{Christopher~TH Baker}.}
  \bibinfo{year}{1977}\natexlab{}.
\newblock \bibinfo{booktitle}{\emph{The numerical treatment of integral
  equations}}.
\newblock \bibinfo{publisher}{Oxford University Press}.
\newblock


\bibitem[Belkin and Niyogi(2003)]%
        {belkin2003laplacian}
\bibfield{author}{\bibinfo{person}{Mikhail Belkin} {and}
  \bibinfo{person}{Partha Niyogi}.} \bibinfo{year}{2003}\natexlab{}.
\newblock \showarticletitle{Laplacian eigenmaps for dimensionality reduction
  and data representation}.
\newblock \bibinfo{journal}{\emph{Neural computation}} \bibinfo{volume}{15},
  \bibinfo{number}{6} (\bibinfo{year}{2003}), \bibinfo{pages}{1373--1396}.
\newblock


\bibitem[Breiman et~al\mbox{.}(2017)]%
        {breiman2017classification}
\bibfield{author}{\bibinfo{person}{Leo Breiman}, \bibinfo{person}{Jerome~H
  Friedman}, \bibinfo{person}{Richard~A Olshen}, {and}
  \bibinfo{person}{Charles~J Stone}.} \bibinfo{year}{2017}\natexlab{}.
\newblock \bibinfo{booktitle}{\emph{Classification and regression trees}}.
\newblock \bibinfo{publisher}{Routledge}.
\newblock


\bibitem[Buitinck et~al\mbox{.}(2013)]%
        {buitinck2013api}
\bibfield{author}{\bibinfo{person}{Lars Buitinck}, \bibinfo{person}{Gilles
  Louppe}, \bibinfo{person}{Mathieu Blondel}, \bibinfo{person}{Fabian
  Pedregosa}, \bibinfo{person}{Andreas Mueller}, \bibinfo{person}{Olivier
  Grisel}, \bibinfo{person}{Vlad Niculae}, \bibinfo{person}{Peter
  Prettenhofer}, \bibinfo{person}{Alexandre Gramfort}, \bibinfo{person}{Jaques
  Grobler}, {et~al\mbox{.}}} \bibinfo{year}{2013}\natexlab{}.
\newblock \showarticletitle{API design for machine learning software:
  experiences from the scikit-learn project}.
\newblock \bibinfo{journal}{\emph{arXiv preprint arXiv:1309.0238}}
  (\bibinfo{year}{2013}).
\newblock


\bibitem[Cai and Ma(2021)]%
        {cai2021theoretical}
\bibfield{author}{\bibinfo{person}{T~Tony Cai} {and} \bibinfo{person}{Rong
  Ma}.} \bibinfo{year}{2021}\natexlab{}.
\newblock \showarticletitle{Theoretical Foundations of t-SNE for Visualizing
  High-Dimensional Clustered Data}.
\newblock \bibinfo{journal}{\emph{arXiv preprint arXiv:2105.07536}}
  (\bibinfo{year}{2021}).
\newblock


\bibitem[Carreira-Perpin{\'a}n(2010)]%
        {carreira2010elastic}
\bibfield{author}{\bibinfo{person}{Miguel~A Carreira-Perpin{\'a}n}.}
  \bibinfo{year}{2010}\natexlab{}.
\newblock \showarticletitle{The Elastic Embedding Algorithm for Dimensionality
  Reduction.}. In \bibinfo{booktitle}{\emph{ICML}}, Vol.~\bibinfo{volume}{10}.
  Citeseer, \bibinfo{pages}{167--174}.
\newblock


\bibitem[Chatzimparmpas et~al\mbox{.}(2020)]%
        {chatzimparmpas2020t}
\bibfield{author}{\bibinfo{person}{Angelos Chatzimparmpas},
  \bibinfo{person}{Rafael~M Martins}, {and} \bibinfo{person}{Andreas Kerren}.}
  \bibinfo{year}{2020}\natexlab{}.
\newblock \showarticletitle{t-viSNE: Interactive Assessment and Interpretation
  of t-SNE Projections}.
\newblock \bibinfo{journal}{\emph{IEEE transactions on visualization and
  computer graphics}} \bibinfo{volume}{26}, \bibinfo{number}{8}
  (\bibinfo{year}{2020}), \bibinfo{pages}{2696--2714}.
\newblock


\bibitem[Chung and Graham(1997)]%
        {chung1997spectral}
\bibfield{author}{\bibinfo{person}{Fan~RK Chung} {and}
  \bibinfo{person}{Fan~Chung Graham}.} \bibinfo{year}{1997}\natexlab{}.
\newblock \bibinfo{booktitle}{\emph{Spectral graph theory}}.
\newblock Number~92. \bibinfo{publisher}{American Mathematical Soc.}
\newblock


\bibitem[Davies and Bouldin(1979)]%
        {davies1979cluster}
\bibfield{author}{\bibinfo{person}{David~L Davies} {and}
  \bibinfo{person}{Donald~W Bouldin}.} \bibinfo{year}{1979}\natexlab{}.
\newblock \showarticletitle{A cluster separation measure}.
\newblock \bibinfo{journal}{\emph{IEEE transactions on pattern analysis and
  machine intelligence}} \bibinfo{number}{2} (\bibinfo{year}{1979}),
  \bibinfo{pages}{224--227}.
\newblock


\bibitem[DeMers and Cottrell(1992)]%
        {demers1992non}
\bibfield{author}{\bibinfo{person}{David DeMers} {and}
  \bibinfo{person}{Garrison Cottrell}.} \bibinfo{year}{1992}\natexlab{}.
\newblock \showarticletitle{Non-linear dimensionality reduction}.
\newblock \bibinfo{journal}{\emph{Advances in neural information processing
  systems}}  \bibinfo{volume}{5} (\bibinfo{year}{1992}).
\newblock


\bibitem[Deng(2012)]%
        {deng2012mnist}
\bibfield{author}{\bibinfo{person}{Li Deng}.} \bibinfo{year}{2012}\natexlab{}.
\newblock \showarticletitle{The mnist database of handwritten digit images for
  machine learning research [best of the web]}.
\newblock \bibinfo{journal}{\emph{IEEE Signal Processing Magazine}}
  \bibinfo{volume}{29}, \bibinfo{number}{6} (\bibinfo{year}{2012}),
  \bibinfo{pages}{141--142}.
\newblock


\bibitem[Donoho(2017)]%
        {donoho201750}
\bibfield{author}{\bibinfo{person}{David Donoho}.}
  \bibinfo{year}{2017}\natexlab{}.
\newblock \showarticletitle{50 years of data science}.
\newblock \bibinfo{journal}{\emph{Journal of Computational and Graphical
  Statistics}} \bibinfo{volume}{26}, \bibinfo{number}{4}
  (\bibinfo{year}{2017}), \bibinfo{pages}{745--766}.
\newblock


\bibitem[Est{\'e}vez et~al\mbox{.}(2009)]%
        {estevez2009normalized}
\bibfield{author}{\bibinfo{person}{Pablo~A Est{\'e}vez},
  \bibinfo{person}{Michel Tesmer}, \bibinfo{person}{Claudio~A Perez}, {and}
  \bibinfo{person}{Jacek~M Zurada}.} \bibinfo{year}{2009}\natexlab{}.
\newblock \showarticletitle{Normalized mutual information feature selection}.
\newblock \bibinfo{journal}{\emph{IEEE Transactions on neural networks}}
  \bibinfo{volume}{20}, \bibinfo{number}{2} (\bibinfo{year}{2009}),
  \bibinfo{pages}{189--201}.
\newblock


\bibitem[Fisher(1936)]%
        {fisher1936use}
\bibfield{author}{\bibinfo{person}{Ronald~A Fisher}.}
  \bibinfo{year}{1936}\natexlab{}.
\newblock \showarticletitle{The use of multiple measurements in taxonomic
  problems}.
\newblock \bibinfo{journal}{\emph{Annals of eugenics}} \bibinfo{volume}{7},
  \bibinfo{number}{2} (\bibinfo{year}{1936}), \bibinfo{pages}{179--188}.
\newblock


\bibitem[Gisbrecht et~al\mbox{.}(2015)]%
        {gisbrecht2015parametric}
\bibfield{author}{\bibinfo{person}{Andrej Gisbrecht},
  \bibinfo{person}{Alexander Schulz}, {and} \bibinfo{person}{Barbara Hammer}.}
  \bibinfo{year}{2015}\natexlab{}.
\newblock \showarticletitle{Parametric nonlinear dimensionality reduction using
  kernel t-SNE}.
\newblock \bibinfo{journal}{\emph{Neurocomputing}}  \bibinfo{volume}{147}
  (\bibinfo{year}{2015}), \bibinfo{pages}{71--82}.
\newblock


\bibitem[Hastie and Stuetzle(1989)]%
        {hastie1989principal}
\bibfield{author}{\bibinfo{person}{Trevor Hastie} {and} \bibinfo{person}{Werner
  Stuetzle}.} \bibinfo{year}{1989}\natexlab{}.
\newblock \showarticletitle{Principal curves}.
\newblock \bibinfo{journal}{\emph{J. Amer. Statist. Assoc.}}
  \bibinfo{volume}{84}, \bibinfo{number}{406} (\bibinfo{year}{1989}),
  \bibinfo{pages}{502--516}.
\newblock


\bibitem[Hinton and Roweis(2002)]%
        {hinton2002stochastic}
\bibfield{author}{\bibinfo{person}{Geoffrey Hinton} {and}
  \bibinfo{person}{Sam~T Roweis}.} \bibinfo{year}{2002}\natexlab{}.
\newblock \showarticletitle{Stochastic neighbor embedding}. In
  \bibinfo{booktitle}{\emph{NIPS}}, Vol.~\bibinfo{volume}{15}. Citeseer,
  \bibinfo{pages}{833--840}.
\newblock


\bibitem[Hinton and Salakhutdinov(2006)]%
        {hinton2006reducing}
\bibfield{author}{\bibinfo{person}{Geoffrey~E Hinton} {and}
  \bibinfo{person}{Ruslan~R Salakhutdinov}.} \bibinfo{year}{2006}\natexlab{}.
\newblock \showarticletitle{Reducing the dimensionality of data with neural
  networks}.
\newblock \bibinfo{journal}{\emph{science}} \bibinfo{volume}{313},
  \bibinfo{number}{5786} (\bibinfo{year}{2006}), \bibinfo{pages}{504--507}.
\newblock


\bibitem[Jolliffe and Cadima(2016)]%
        {jolliffe2016principal}
\bibfield{author}{\bibinfo{person}{Ian~T Jolliffe} {and} \bibinfo{person}{Jorge
  Cadima}.} \bibinfo{year}{2016}\natexlab{}.
\newblock \showarticletitle{Principal component analysis: a review and recent
  developments}.
\newblock \bibinfo{journal}{\emph{Philosophical Transactions of the Royal
  Society A: Mathematical, Physical and Engineering Sciences}}
  \bibinfo{volume}{374}, \bibinfo{number}{2065} (\bibinfo{year}{2016}),
  \bibinfo{pages}{20150202}.
\newblock


\bibitem[Kobak and Berens(2019)]%
        {kobak2019art}
\bibfield{author}{\bibinfo{person}{Dmitry Kobak} {and} \bibinfo{person}{Philipp
  Berens}.} \bibinfo{year}{2019}\natexlab{}.
\newblock \showarticletitle{The art of using t-SNE for single-cell
  transcriptomics}.
\newblock \bibinfo{journal}{\emph{Nature communications}} \bibinfo{volume}{10},
  \bibinfo{number}{1} (\bibinfo{year}{2019}), \bibinfo{pages}{1--14}.
\newblock


\bibitem[Kohonen(1982)]%
        {kohonen1982self}
\bibfield{author}{\bibinfo{person}{Teuvo Kohonen}.}
  \bibinfo{year}{1982}\natexlab{}.
\newblock \showarticletitle{Self-organized formation of topologically correct
  feature maps}.
\newblock \bibinfo{journal}{\emph{Biological cybernetics}}
  \bibinfo{volume}{43}, \bibinfo{number}{1} (\bibinfo{year}{1982}),
  \bibinfo{pages}{59--69}.
\newblock


\bibitem[Li et~al\mbox{.}(2017)]%
        {li2017application}
\bibfield{author}{\bibinfo{person}{Wentian Li}, \bibinfo{person}{Jane~E
  Cerise}, \bibinfo{person}{Yaning Yang}, {and} \bibinfo{person}{Henry Han}.}
  \bibinfo{year}{2017}\natexlab{}.
\newblock \showarticletitle{Application of t-SNE to human genetic data}.
\newblock \bibinfo{journal}{\emph{Journal of bioinformatics and computational
  biology}} \bibinfo{volume}{15}, \bibinfo{number}{04} (\bibinfo{year}{2017}),
  \bibinfo{pages}{1750017}.
\newblock


\bibitem[Linderman et~al\mbox{.}(2019)]%
        {linderman2019fast}
\bibfield{author}{\bibinfo{person}{George~C Linderman}, \bibinfo{person}{Manas
  Rachh}, \bibinfo{person}{Jeremy~G Hoskins}, \bibinfo{person}{Stefan
  Steinerberger}, {and} \bibinfo{person}{Yuval Kluger}.}
  \bibinfo{year}{2019}\natexlab{}.
\newblock \showarticletitle{Fast interpolation-based t-SNE for improved
  visualization of single-cell RNA-seq data}.
\newblock \bibinfo{journal}{\emph{Nature methods}} \bibinfo{volume}{16},
  \bibinfo{number}{3} (\bibinfo{year}{2019}), \bibinfo{pages}{243--245}.
\newblock


\bibitem[McInnes et~al\mbox{.}(2018a)]%
        {mcinnes2018umap}
\bibfield{author}{\bibinfo{person}{Leland McInnes}, \bibinfo{person}{John
  Healy}, {and} \bibinfo{person}{James Melville}.}
  \bibinfo{year}{2018}\natexlab{a}.
\newblock \showarticletitle{Umap: Uniform manifold approximation and projection
  for dimension reduction}.
\newblock \bibinfo{journal}{\emph{arXiv preprint arXiv:1802.03426}}
  (\bibinfo{year}{2018}).
\newblock


\bibitem[McInnes et~al\mbox{.}(2018b)]%
        {mcinnes2018umap-software}
\bibfield{author}{\bibinfo{person}{Leland McInnes}, \bibinfo{person}{John
  Healy}, \bibinfo{person}{Nathaniel Saul}, {and} \bibinfo{person}{Lukas
  Grossberger}.} \bibinfo{year}{2018}\natexlab{b}.
\newblock \showarticletitle{UMAP: Uniform Manifold Approximation and
  Projection}.
\newblock \bibinfo{journal}{\emph{The Journal of Open Source Software}}
  \bibinfo{volume}{3}, \bibinfo{number}{29} (\bibinfo{year}{2018}),
  \bibinfo{pages}{861}.
\newblock


\bibitem[Nene et~al\mbox{.}(1996a)]%
        {nene1996columbia}
\bibfield{author}{\bibinfo{person}{SA Nene}, \bibinfo{person}{SK Nayar}, {and}
  \bibinfo{person}{H Murase}.} \bibinfo{year}{1996}\natexlab{a}.
\newblock \showarticletitle{Columbia university image library (coil-20)}.
\newblock \bibinfo{journal}{\emph{Technical Report CUCS-005-96}}
  (\bibinfo{year}{1996}).
\newblock


\bibitem[Nene et~al\mbox{.}(1996b)]%
        {nene1996columbia-2}
\bibfield{author}{\bibinfo{person}{Sameer~A Nene}, \bibinfo{person}{Shree~K
  Nayar}, \bibinfo{person}{Hiroshi Murase}, {et~al\mbox{.}}}
  \bibinfo{year}{1996}\natexlab{b}.
\newblock \showarticletitle{Columbia object image library (coil-100)}.
\newblock  (\bibinfo{year}{1996}).
\newblock


\bibitem[Pearson(1901)]%
        {pearson1901liii}
\bibfield{author}{\bibinfo{person}{Karl Pearson}.}
  \bibinfo{year}{1901}\natexlab{}.
\newblock \showarticletitle{LIII. On lines and planes of closest fit to systems
  of points in space}.
\newblock \bibinfo{journal}{\emph{The London, Edinburgh, and Dublin
  philosophical magazine and journal of science}} \bibinfo{volume}{2},
  \bibinfo{number}{11} (\bibinfo{year}{1901}), \bibinfo{pages}{559--572}.
\newblock


\bibitem[Pedregosa et~al\mbox{.}(2011)]%
        {scikit-learn}
\bibfield{author}{\bibinfo{person}{F. Pedregosa}, \bibinfo{person}{G.
  Varoquaux}, \bibinfo{person}{A. Gramfort}, \bibinfo{person}{V. Michel},
  \bibinfo{person}{B. Thirion}, \bibinfo{person}{O. Grisel},
  \bibinfo{person}{M. Blondel}, \bibinfo{person}{P. Prettenhofer},
  \bibinfo{person}{R. Weiss}, \bibinfo{person}{V. Dubourg}, \bibinfo{person}{J.
  Vanderplas}, \bibinfo{person}{A. Passos}, \bibinfo{person}{D. Cournapeau},
  \bibinfo{person}{M. Brucher}, \bibinfo{person}{M. Perrot}, {and}
  \bibinfo{person}{E. Duchesnay}.} \bibinfo{year}{2011}\natexlab{}.
\newblock \showarticletitle{Scikit-learn: Machine Learning in {P}ython}.
\newblock \bibinfo{journal}{\emph{Journal of Machine Learning Research}}
  \bibinfo{volume}{12} (\bibinfo{year}{2011}), \bibinfo{pages}{2825--2830}.
\newblock


\bibitem[Pezzotti et~al\mbox{.}(2016)]%
        {pezzotti2016approximated}
\bibfield{author}{\bibinfo{person}{Nicola Pezzotti},
  \bibinfo{person}{Boudewijn~PF Lelieveldt}, \bibinfo{person}{Laurens Van
  Der~Maaten}, \bibinfo{person}{Thomas H{\"o}llt}, \bibinfo{person}{Elmar
  Eisemann}, {and} \bibinfo{person}{Anna Vilanova}.}
  \bibinfo{year}{2016}\natexlab{}.
\newblock \showarticletitle{Approximated and user steerable tSNE for
  progressive visual analytics}.
\newblock \bibinfo{journal}{\emph{IEEE transactions on visualization and
  computer graphics}} \bibinfo{volume}{23}, \bibinfo{number}{7}
  (\bibinfo{year}{2016}), \bibinfo{pages}{1739--1752}.
\newblock


\bibitem[Reyes-Ortiz et~al\mbox{.}(2013)]%
        {reyes2013human}
\bibfield{author}{\bibinfo{person}{Jorge~Luis Reyes-Ortiz},
  \bibinfo{person}{Alessandro Ghio}, \bibinfo{person}{Xavier Parra},
  \bibinfo{person}{Davide Anguita}, \bibinfo{person}{Joan Cabestany}, {and}
  \bibinfo{person}{Andreu Catala}.} \bibinfo{year}{2013}\natexlab{}.
\newblock \showarticletitle{Human Activity and Motion Disorder Recognition:
  towards smarter Interactive Cognitive Environments.}. In
  \bibinfo{booktitle}{\emph{ESANN}}. Citeseer.
\newblock


\bibitem[Rousseeuw(1987)]%
        {rousseeuw1987silhouettes}
\bibfield{author}{\bibinfo{person}{Peter~J Rousseeuw}.}
  \bibinfo{year}{1987}\natexlab{}.
\newblock \showarticletitle{Silhouettes: a graphical aid to the interpretation
  and validation of cluster analysis}.
\newblock \bibinfo{journal}{\emph{Journal of computational and applied
  mathematics}}  \bibinfo{volume}{20} (\bibinfo{year}{1987}),
  \bibinfo{pages}{53--65}.
\newblock


\bibitem[Roweis and Saul(2000)]%
        {roweis2000nonlinear}
\bibfield{author}{\bibinfo{person}{Sam~T Roweis} {and}
  \bibinfo{person}{Lawrence~K Saul}.} \bibinfo{year}{2000}\natexlab{}.
\newblock \showarticletitle{Nonlinear dimensionality reduction by locally
  linear embedding}.
\newblock \bibinfo{journal}{\emph{science}} \bibinfo{volume}{290},
  \bibinfo{number}{5500} (\bibinfo{year}{2000}), \bibinfo{pages}{2323--2326}.
\newblock


\bibitem[Sammon(1969)]%
        {sammon1969nonlinear}
\bibfield{author}{\bibinfo{person}{John~W Sammon}.}
  \bibinfo{year}{1969}\natexlab{}.
\newblock \showarticletitle{A nonlinear mapping for data structure analysis}.
\newblock \bibinfo{journal}{\emph{IEEE Transactions on computers}}
  \bibinfo{volume}{100}, \bibinfo{number}{5} (\bibinfo{year}{1969}),
  \bibinfo{pages}{401--409}.
\newblock


\bibitem[Sch{\"o}lkopf et~al\mbox{.}(1998)]%
        {scholkopf1998nonlinear}
\bibfield{author}{\bibinfo{person}{Bernhard Sch{\"o}lkopf},
  \bibinfo{person}{Alexander Smola}, {and} \bibinfo{person}{Klaus-Robert
  M{\"u}ller}.} \bibinfo{year}{1998}\natexlab{}.
\newblock \showarticletitle{Nonlinear component analysis as a kernel eigenvalue
  problem}.
\newblock \bibinfo{journal}{\emph{Neural computation}} \bibinfo{volume}{10},
  \bibinfo{number}{5} (\bibinfo{year}{1998}), \bibinfo{pages}{1299--1319}.
\newblock


\bibitem[Souza(2010)]%
        {souza2010kernel}
\bibfield{author}{\bibinfo{person}{C{\'e}sar~R Souza}.}
  \bibinfo{year}{2010}\natexlab{}.
\newblock \showarticletitle{Kernel functions for machine learning
  applications}.
\newblock \bibinfo{journal}{\emph{Creative Commons
  Attribution-Noncommercial-Share Alike}}  \bibinfo{volume}{3}
  (\bibinfo{year}{2010}), \bibinfo{pages}{29}.
\newblock


\bibitem[Tenenbaum et~al\mbox{.}(2000)]%
        {tenenbaum2000global}
\bibfield{author}{\bibinfo{person}{Joshua~B Tenenbaum}, \bibinfo{person}{Vin
  De~Silva}, {and} \bibinfo{person}{John~C Langford}.}
  \bibinfo{year}{2000}\natexlab{}.
\newblock \showarticletitle{A global geometric framework for nonlinear
  dimensionality reduction}.
\newblock \bibinfo{journal}{\emph{science}} \bibinfo{volume}{290},
  \bibinfo{number}{5500} (\bibinfo{year}{2000}), \bibinfo{pages}{2319--2323}.
\newblock


\bibitem[Vaida(2005)]%
        {vaida2005parameter}
\bibfield{author}{\bibinfo{person}{Florin Vaida}.}
  \bibinfo{year}{2005}\natexlab{}.
\newblock \showarticletitle{Parameter convergence for EM and MM algorithms}.
\newblock \bibinfo{journal}{\emph{Statistica Sinica}} (\bibinfo{year}{2005}),
  \bibinfo{pages}{831--840}.
\newblock


\bibitem[Van Der~Maaten(2014)]%
        {van2014accelerating}
\bibfield{author}{\bibinfo{person}{Laurens Van Der~Maaten}.}
  \bibinfo{year}{2014}\natexlab{}.
\newblock \showarticletitle{Accelerating t-SNE using tree-based algorithms}.
\newblock \bibinfo{journal}{\emph{The Journal of Machine Learning Research}}
  \bibinfo{volume}{15}, \bibinfo{number}{1} (\bibinfo{year}{2014}),
  \bibinfo{pages}{3221--3245}.
\newblock


\bibitem[Van~der Maaten and Hinton(2008)]%
        {van2008visualizing}
\bibfield{author}{\bibinfo{person}{Laurens Van~der Maaten} {and}
  \bibinfo{person}{Geoffrey Hinton}.} \bibinfo{year}{2008}\natexlab{}.
\newblock \showarticletitle{Visualizing data using t-SNE.}
\newblock \bibinfo{journal}{\emph{Journal of machine learning research}}
  \bibinfo{volume}{9}, \bibinfo{number}{11} (\bibinfo{year}{2008}).
\newblock


\bibitem[Xiao et~al\mbox{.}(2017)]%
        {xiao2017fashion}
\bibfield{author}{\bibinfo{person}{Han Xiao}, \bibinfo{person}{Kashif Rasul},
  {and} \bibinfo{person}{Roland Vollgraf}.} \bibinfo{year}{2017}\natexlab{}.
\newblock \showarticletitle{Fashion-mnist: a novel image dataset for
  benchmarking machine learning algorithms}.
\newblock \bibinfo{journal}{\emph{arXiv preprint arXiv:1708.07747}}
  (\bibinfo{year}{2017}).
\newblock


\bibitem[Xie et~al\mbox{.}(2011)]%
        {xie2011m}
\bibfield{author}{\bibinfo{person}{Bo Xie}, \bibinfo{person}{Yang Mu},
  \bibinfo{person}{Dacheng Tao}, {and} \bibinfo{person}{Kaiqi Huang}.}
  \bibinfo{year}{2011}\natexlab{}.
\newblock \showarticletitle{m-SNE: Multiview stochastic neighbor embedding}.
\newblock \bibinfo{journal}{\emph{IEEE Transactions on Systems, Man, and
  Cybernetics, Part B (Cybernetics)}} \bibinfo{volume}{41}, \bibinfo{number}{4}
  (\bibinfo{year}{2011}), \bibinfo{pages}{1088--1096}.
\newblock


\bibitem[Yang et~al\mbox{.}(2009)]%
        {yang2009heavy}
\bibfield{author}{\bibinfo{person}{Zhirong Yang}, \bibinfo{person}{Irwin King},
  \bibinfo{person}{Zenglin Xu}, {and} \bibinfo{person}{Erkki Oja}.}
  \bibinfo{year}{2009}\natexlab{}.
\newblock \showarticletitle{Heavy-tailed symmetric stochastic neighbor
  embedding}.
\newblock \bibinfo{journal}{\emph{Advances in neural information processing
  systems}}  \bibinfo{volume}{22} (\bibinfo{year}{2009}),
  \bibinfo{pages}{2169--2177}.
\newblock


\end{thebibliography}
{\small

}

\newpage
\appendix
\section{Appendices}

\subsection{Proof for Proposition 1}\label{append_proof}

Denote $\hat{\mathcal{L}}_2(\mathbf{Y}|\mathbf{Y}_{t-1})=\hat{\mathcal{L}}_2(\mathbf{Y})$.  Using the definitions of ${\mathcal{L}}_2(\mathbf{Y})$ and $\hat{\mathcal{L}}_2(\mathbf{Y})$, we have 
\begin{equation}
{\mathcal{L}}_2(\mathbf{Y})\leq \hat{\mathcal{L}}_2(\mathbf{Y}|\mathbf{Y}_{t-1})
\end{equation}
and 
\begin{equation}
\mathcal{L}_1(\mathbf{Y})+{\mathcal{L}}_2(\mathbf{Y})\leq \mathcal{L}_1(\mathbf{Y})+\hat{\mathcal{L}}_2(\mathbf{Y}|\mathbf{Y}_{t-1}).
\end{equation}
It follows that 
\begin{equation}\label{eq proof L12}
\mathcal{L}_1(\mathbf{Y}_t)+{\mathcal{L}}_2(\mathbf{Y}_t)\leq \mathcal{L}_1(\mathbf{Y}_t)+\hat{\mathcal{L}}_2(\mathbf{Y}_{t}|\mathbf{Y}_{t-1})
\end{equation}

Supposing $\mathcal{L}_1(\mathbf{Y}_t)+\hat{\mathcal{L}}_2(\mathbf{Y}_t)\leq \mathcal{L}_1(\mathbf{Y}_{t-1})+\hat{\mathcal{L}}_2(\mathbf{Y}_{t-1})-\Delta_{t}$ and combining \eqref{eq proof L12}, we arrive at

\begin{equation}\label{eq_proof_L123}
\mathcal{L}_1(\mathbf{Y}_t)+{\mathcal{L}}_2(\mathbf{Y}_t)\leq \mathcal{L}_1(\mathbf{Y}_{t-1})+ \hat{\mathcal{L}}_2(\mathbf{Y}_{t-1})-\Delta_{t}.
\end{equation}

Invoking the fact  ${\mathcal{L}}_2(\mathbf{Y}_{t-1})=\hat{\mathcal{L}}_2(\mathbf{Y}_{t-1}|\mathbf{Y}_{t-1})=\bar{\mathcal{L}}_2(\mathbf{Y}_{t-1})$ into \eqref{eq_proof_L123},  we obtain

\begin{equation}\label{eq_proof_L1234}
\mathcal{L}_1(\mathbf{Y}_t)+{\mathcal{L}}_2(\mathbf{Y}_t)\leq \mathcal{L}_1(\mathbf{Y}_{t-1})+ {\mathcal{L}}_2(\mathbf{Y}_{t-1})-\Delta_{t}.
\end{equation}

Because $\nabla\hat{\mathcal{L}}(\mathbf{Y})$ is $L$-Lipschitz continuous, we have
\begin{equation}\label{eq_proof_lipschitz}
\begin{aligned}
\hat{\mathcal{L}}(\mathbf{Y})\leq& \hat{\mathcal{L}}(\mathbf{Y}_{t-1})+\left\langle \nabla\hat{\mathcal{L}}(\mathbf{Y}_{t-1}),\mathbf{Y}-\mathbf{Y}_{t-1}\right\rangle\\
&+\frac{L}{2}\Vert\mathbf{Y}-\mathbf{Y}_{t-1}\Vert_F^2.
\end{aligned}
\end{equation}
Invoking the update of $\mathbf{Y}$, i.e.,  $\mathbf{Y}_t=\mathbf{Y}_{t-1}-\alpha\nabla\hat{\mathcal{L}}(\mathbf{Y}_{t-1})$, into \eqref{eq_proof_lipschitz}, we get
\begin{equation}\label{eq_proof_lipschitz20}
\begin{aligned}
\hat{\mathcal{L}}(\mathbf{Y}_t)\leq \hat{\mathcal{L}}(\mathbf{Y}_{t-1})-\left(\frac{1}{\alpha}-\frac{L}{2}\right)\left\Vert\mathbf{Y}_t-\mathbf{Y}_{t-1}\right\Vert_F^2
\end{aligned}
\end{equation}
It means
\begin{equation}\label{eq_proof_lipschitz2}
\begin{aligned}
\Delta_t=\left(\frac{1}{\alpha}-\frac{L}{2}\right)\left\Vert\mathbf{Y}_t-\mathbf{Y}_{t-1}\right\Vert_F^2
\end{aligned}.
\end{equation}
Now combining \eqref{eq_proof_L1234} and \eqref{eq_proof_lipschitz2}, we arrive at
\begin{equation}\label{eq_proof_decrease}
\mathcal{L}(\mathbf{Y}_t)\leq \mathcal{L}(\mathbf{Y}_{t-1})-\left(\frac{1}{\alpha}-\frac{L}{2}\right)\left\Vert\mathbf{Y}_t-\mathbf{Y}_{t-1}\right\Vert_F^2.
\end{equation}
Sum up \eqref{eq_proof_decrease} from $t=1$ to $t=T$, we have
\begin{equation}\label{eq_proof_decrease}
\mathcal{L}(\mathbf{Y}_T)\leq \mathcal{L}(\mathbf{Y}_{0})-\left(\frac{1}{\alpha}-\frac{L}{2}\right)\sum_{t=1}^T\left\Vert\mathbf{Y}_t-\mathbf{Y}_{t-1}\right\Vert_F^2.
\end{equation}
This finished the proof.

\subsection{Gradient Computation for Gaussian Kernel}

For Gaussian kernel with $\sigma_{i}$ for $\mathbf{x}_i$,
\begin{equation}
    K_{ij} = \exp\left(- \frac{\| \mathbf{y_i} - \mathbf{y_j} \|^2 }{2 \sigma_i^2}\right).
\end{equation}

As the adjacent matrix $\mathbf{Q}$ is symmetric, 
\begin{equation}
 q_{ji} = q_{ij} = \frac{K_{ij} + K_{ji}}{2}   
\end{equation}

\begin{equation}
 \begin{aligned}
    & \frac{\partial \mathrm{Tr}(\mathbf{V^{\top}LV})}{\partial \mathbf{Y}} \\
    & = \sum_m \sum_n (U^0_{mn} + U^1_{mn} + U^2_{mn}) \odot  \frac{\partial q_{mn}}{\partial \mathbf{Y}} \\
    & = \sum_m \sum_n (U^0_{mn} + U^1_{mn} + U^2_{mn}) \sum_i \sum_j  \frac{\partial q_{mn}}{\partial K_{ij}}\frac{\partial K_{ij}}{\partial \mathbf{Y}}\\
    & = \sum_i \sum_j (\sum_m \sum_n (U^0_{mn} + U^1_{mn} + U^2_{mn}) \frac{\partial q_{mn}}{\partial K_{ij}})\odot \frac{\partial K_{ij}}{\partial \mathbf{Y}} \\
    & = \sum_i \sum_j [\frac{1}{2}(U^0_{ij} + U^1_{ij} + U^2_{ij}) + \frac{1}{2}(U^0_{ji} + U^1_{ji} + U^2_{ji})]\odot \frac{\partial K_{ij}}{\partial \mathbf{Y}} \\
    & = \sum_i \sum_j (U^0_{ij} + U^1_{ij} + U^2_{ij})\odot \frac{\partial K_{ij}}{\partial \mathbf{Y}} \\
\end{aligned}   
\end{equation}

In case of Gaussian kernel matrix, we can calculate the gradient on element $\mathbf{y_a}$ of $\mathbf{Y}$ as below.
\begin{equation}   
\begin{aligned}
    & \frac{\partial \mathrm{Tr}(\mathbf{V^{\top}LV})}{\partial \mathbf{y_a}} \\
    &= \sum _i \sum _j \frac{\partial \mathrm{Tr}(\mathbf{V^{\top}LV})}{\partial K_{ij}} \frac{\partial K_{ij}}{\partial \mathbf{y_a}} \\
     &= \begin{cases}
      \frac{1}{-\sigma_a^2} \sum _j \frac{\partial \mathrm{Tr}(\mathbf{V^{\top}LV})}{\partial K_{aj}} K_{aj}) (\mathbf{y_a} - \mathbf{y_j}) & i = a\\
      \sum _i \frac{1}{-\sigma_i^2} \frac{\partial \mathrm{Tr}(\mathbf{V^{\top}LV})}{\partial K_{ia}} K_{ia}) (\mathbf{y_a} - \mathbf{y_i}) & j = a
    \end{cases}\\
    & \left ( \text{let} \quad U_{ij} = \frac{U^0_{ij} + U^1_{ij} + U^2_{ij}}{\sigma_{i}^2} \odot K_{ij} \right ) \\
    & = \begin{cases}
    \sum _j U_{aj} \mathbf{y_j} - \mathbf{y_a} \sum _j U_{aj} & i = a\\
    \sum _i U_{ia} \mathbf{y_i} - \mathbf{y_a} \sum _i U_{ia} & j = a\\
    \end{cases}\\
   & \text{so,} \quad \frac{\partial \mathrm{Tr}(\mathbf{V^{\top}LV})}{\partial \mathbf{Y}} \\
   & = \begin{cases}
      \mathbf{UY} - \mathbf{Y}\odot \mathbf{C}\\
      \mathbf{U}^{\top} \mathbf{Y} - \mathbf{Y} \odot \mathbf{C}'
      \end{cases}\\
    & \text{where,} \quad \mathbf{C} = \mathbf{U}\mathbf{1}_{N}\mathbf{1}_{d}^{\top} \quad \mathbf{C}' = \mathbf{1}_{d}\mathbf{1}_{N}^{\top}\mathbf{U}
 \end{aligned}
\end{equation}
To sum up, the calculation of gradient $\frac{\partial \mathrm{Tr}(\mathbf{V^{\top}LV})}{\partial \mathbf{Y}}$ is 

\begin{equation}
\begin{aligned}
\frac{\partial \mathrm{Tr}(\mathbf{V^{\top}LV})}{\partial \mathbf{Y}} = (\mathbf{U} + \mathbf{U}^{\top})\mathbf{Y} - \mathbf{Y} \odot (\mathbf{C} + \mathbf{C}').
\end{aligned}
\end{equation}

\subsection{More Numerical Results}

In section \ref{sec_exp}, NMI, SC, DBI scores of \textsc{Lap}tSNE outperform others in most cases. As for $k$-means on low-dimensional embedding, we can choose $k$ as the estimated number of potential cluster $\hat{k}$ to examine \textsc{Lap}tSNE, rather than the exact number of the ground-truth labels. The results in terms of NMI, SC, and DBI on the seven datasets are reported in Table \ref{table: additional kmeans}. Our \textsc{Lap}tSNE and \textsc{Lap}tSNE-Mini still outperform the baselines in most cases.

\begin{table}[h]
\footnotesize
\centering
\caption{Comparison of clustering performances (NMI, SC and DBI) between \textsc{Lap}tSNE, \textsc{Lap}tSNE-Mini, t-SNE, UMAP, Eigenmaps and PCA on all datasets. \textit{Note: $\uparrow$ means the higher is better, whereas $\downarrow$ indicates the lower is better}. The best value in each case is highlighted in bold.}
\label{table: additional kmeans}
\begin{tabular}{m{1em}c|cccccc}
\toprule
   {\begin{sideways}\textbf{~~}\end{sideways}} & \textbf{score} & \textbf{\textsc{Lap}tSNE}  & \textbf{\textsc{Lap}tSNE-Mini}  & \textbf{t-SNE}  & \textbf{UMAP} & \textbf{Eigenmaps} &\textbf{PCA}
   \\ 
\cline{2-8}
\rotatebox{90}{PenDigits}& 
\makecell{NMI $\uparrow$\\SC $\uparrow$\\ DBI $\downarrow$} &
\makecell{ \textbf{0.9284} \\ \textbf{0.7605} \\ \textbf{0.3806}}   & 
\makecell{0.8901 \\ 0.6920 \\ 0.4175}   & 
\makecell{ 0.7679 \\ 0.4887 \\ 0.7228}   & 
\makecell{ 0.9083 \\0.6115 \\ 0.5407}   & 
\makecell{0.7804 \\ 0.6832 \\ 0.4784}  &
\makecell{0.5164 \\  0.3918 \\ 0.7933} 
\\
\cline{2-8}
\rotatebox{90}{COIL20}& 
\makecell{NMI $\uparrow$\\SC $\uparrow$\\ DBI $\downarrow$} &
\makecell{ \textbf{0.8787} \\ \textbf{0.8023} \\ \textbf{0.2863}}   & 
\makecell{0.8785 \\ 0.8021 \\ 0.2866}   & 
\makecell{0.8218 \\ 0.5024 \\ 0.6741}   & 
\makecell{0.8458 \\ 0.5700 \\ 0.5935}   & 
\makecell{0.5477 \\ 0.5720 \\ 0.6499}  &
\makecell{0.5527 \\  0.5760 \\ 0.6394} 
\\
\cline{2-8}
\rotatebox{90}{COIL100}& 
\makecell{NMI $\uparrow$\\SC $\uparrow$\\ DBI $\downarrow$} &
\makecell{0.5119 \\ 0.3762 \\ 0.7909}   & 
\makecell{0.5035 \\ 0.3721 \\ 0.7923}   & 
\makecell{0.4741 \\ 0.3572 \\ 0.7846}   & 
\makecell{\textbf{0.5362} \\ 0.3659 \\ 0.8021}   & 
\makecell{0.4532 \\ \textbf{0.4384} \\ 0.7899}  &
\makecell{0.4440 \\ 0.4095 \\ \textbf{0.7371} }
\\
\cline{2-8}
\rotatebox{90}{Wavform}& 
\makecell{NMI $\uparrow$\\SC $\uparrow$\\ DBI $\downarrow$} &
\makecell{ 0.3555 \\ \textbf{0.9185} \\ \textbf{0.1100}}   & 
\makecell{0.3529 \\ 0.7367 \\ 0.3634}   & 
\makecell{ 0.3592 \\ 0.3535 \\ 0.8586}   & 
\makecell{ \textbf{0.3633} \\0.3480 \\ 0.8483}   & 
\makecell{0.3101 \\ 0.3464 \\ 0.8447}  &
\makecell{0.3589 \\ 0.4633 \\ 0.6763} 
\\
\cline{2-8}
\rotatebox{90}{MNIST}& 
\makecell{NMI $\uparrow$\\SC $\uparrow$\\ DBI $\downarrow$} &
\makecell{\textbf{0.7278} \\ \textbf{0.6147} \\ \textbf{0.4885}}  & 
\makecell{0.7017\\ 0.6111 \\ 0.5264}   & 
\makecell{0.6129 \\ 0.3908 \\ 0.8074}   & 
\makecell{0.7004 \\ 0.4882 \\ 0.7182}   & 
\makecell{0.3137 \\ 0.3435 \\ 0.8247}  &
\makecell{0.3492\\ 0.3435 \\ 0.8247} 
\\
\cline{2-8}

\rotatebox{90}{{\small F-MNIST}}   & 
\makecell{NMI $\uparrow$\\SC $\uparrow$\\ DBI $\downarrow$} &
\makecell{\textbf{0.5910} \\ \textbf{0.6114} \\ \textbf{0.5043}}   & 
\makecell{0.5818 \\ 0.5125 \\ 0.6680}   & 
\makecell{0.5221 \\ 0.4178 \\ 0.7068}   & 
\makecell{0.5881 \\0.5093 \\ 0.7055}   & 
\makecell{0.4565 \\ 0.4734 \\ 0.7244}  &
\makecell{0.4261 \\ 0.3815 \\ 0.8280} 
\\
\cline{2-8}
\rotatebox{90}{{\small HAR}}   & 
\makecell{NMI $\uparrow$\\SC $\uparrow$\\ DBI $\downarrow$} &
\makecell{0.6296 \\ 0.4586 \\ 0.7764}   & 
\makecell{0.5689 \\ 0.3797 \\ 0.8363}   & 
\makecell{0.4602 \\ 0.3623 \\ 0.8446}   & 
\makecell{\textbf{0.6179} \\ 0.4151\\ 0.8311}   & 
\makecell{0.5835 \\ \textbf{0.7413} \\ \textbf{0.5168}}  &
\makecell{0.4245 \\ 0.3906 \\ 0.8153} 
\\
\bottomrule
\end{tabular}
\end{table}

\subsection{More Qualitative Comparison}

Figure \ref{com-othmed} shows the qualitative comparison of \textsc{Lap}tSNE with other methods. \textsc{Lap}tSNE has comparable performance with UMAP on visual results, and it is significantly better than other methods.

\begin{figure}[t]\label{com-othmed}
\centering
\includegraphics[width=0.8\textwidth]{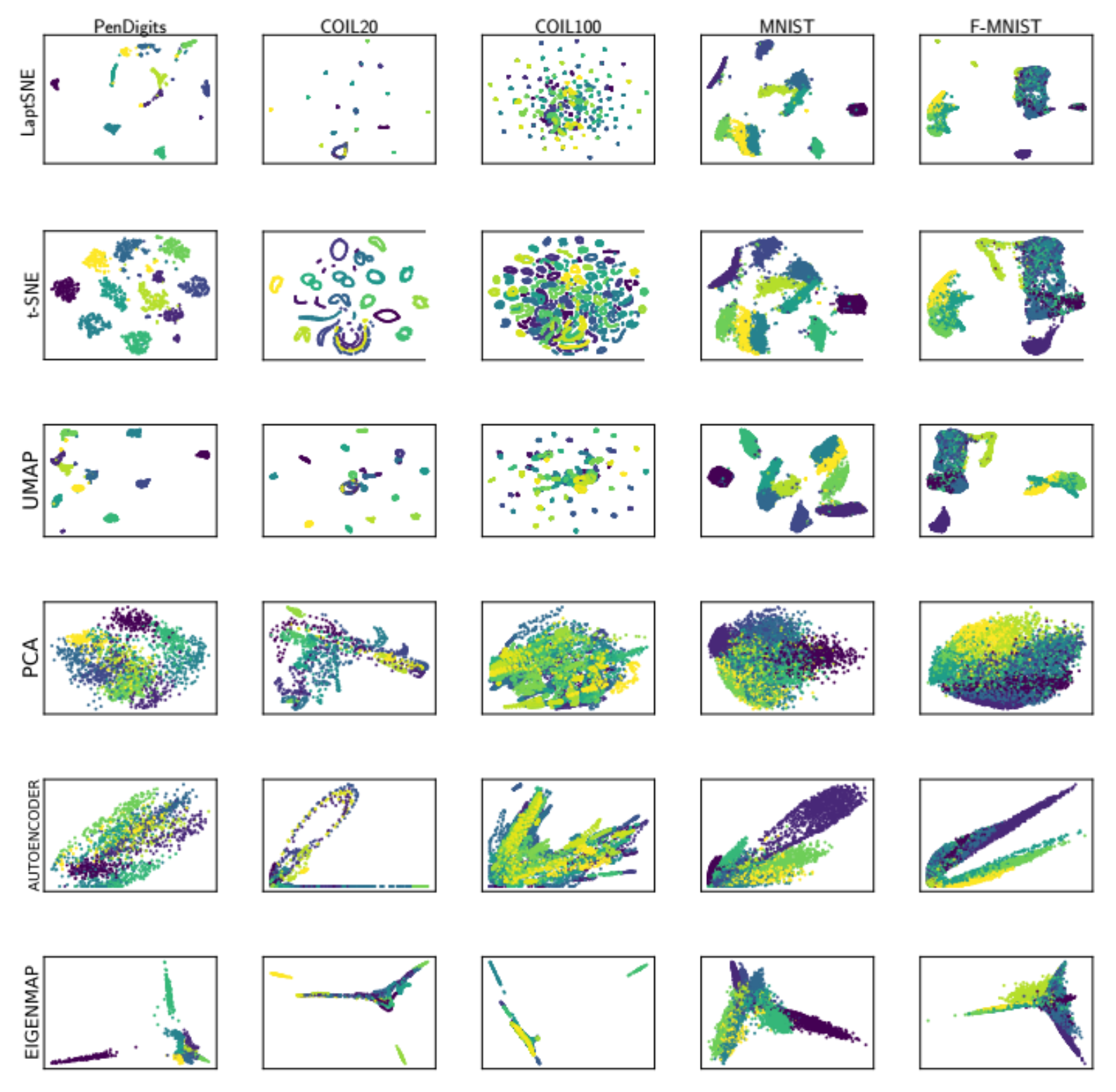}
  \caption{A comparison of \textsc{Lap}tSNE and other dimensionality reduction methods for all datasets.}
\label{com-othmed}
\end{figure}

\end{document}